\documentclass{article}

    \PassOptionsToPackage{numbers, compress}{natbib}

\usepackage[final]{neurips_data_2023}

\usepackage{etex}
\usepackage[utf8]{inputenc} 
\usepackage[T1]{fontenc}    
\usepackage{hyperref}       
\usepackage{url}            
\usepackage{booktabs}       
\usepackage{amsfonts}       
\usepackage{nicefrac}       
\usepackage{microtype}      
\usepackage{xcolor}         

\usepackage[commandnameprefix=always]{changes}
\usepackage{wrapfig,lipsum,booktabs}
\usepackage{hyperref} 
\usepackage{cancel}
\usepackage{fontawesome5}
\usepackage{times}
\usepackage{latexsym}
\usepackage{graphicx}
\usepackage{adjustbox}
\usepackage{multirow, multicol}
\usepackage{caption}
\usepackage{multicol}
\usepackage{makecell}
\usepackage{amsmath}
\usepackage{tabularray}
\usepackage{arydshln}
\usepackage{enumitem}
\usepackage{listings}
\usepackage{float}
\usepackage{color}
\usepackage{outlines}
\usepackage{tcolorbox}
\usepackage{xcolor}
\usepackage{xspace}
\usepackage{threeparttable}
\usepackage[noabbrev,capitalize,nameinlink]{cleveref}
\usepackage{bbding}

\usepackage{titletoc}
\usepackage{wasysym}
\usepackage{longtable} 
\usepackage{tabularx}
\usepackage{booktabs}
\usepackage{threeparttable}

\usepackage{bbding}
\usepackage{enumitem}
\usepackage{pifont}
\newcommand{\cmark}{\ding{51}}
\newcommand{\xmark}{\ding{55}}

\newcommand{\eg}{\textit{e.g.}} 
\newcommand{\ie}{{\textit{i.e.}}} 
\newcommand{\etal}{{\textit{et al.}}} 
\newcommand{\MAE}{{$\text{M}^3\text{AE}$\xspace}}

\crefname{section}{Sec.}{Sec.}
\Crefname{Section}{Sec.}{Sec.}

\newcommand{\equal}[1]{{\hypersetup{linkcolor=black}\thanks{#1}}}

\title{EHRXQA: A Multi-Modal Question Answering Dataset for Electronic Health Records with Chest X-ray Images}

\author{
    Seongsu Bae$^{1}$\equal{These authors contributed equally}\;,
    Daeun Kyung$^{1}$\footnotemark[1]\;,
    Jaehee Ryu$^{1}$, 
    Eunbyeol Cho$^{1}$,
    Gyubok Lee$^{1}$, \\
    \textbf{Sunjun Kweon$^{1}$},
    \textbf{Jeongwoo Oh$^{1}$}, 
    \textbf{Lei Ji$^{2}$},  
    \textbf{Eric I-Chao Chang$^{3}$}, 
    \textbf{Tackeun Kim$^{4}$}, 
    \textbf{Edward Choi$^{1,}$\thanks{Corresponding author}} \\
    KAIST$^{1}$ \enskip
    Microsoft Research Asia$^{2}$ \enskip
    Centre of Perceptual and Interactive Intelligence$^{3}$ \\
    Seoul National University Bundang Hospital$^{4}$\\
    \texttt{\{seongsu,kyungdaeun,edwardchoi\}@kaist.ac.kr}$^{1}$
}

\begin{document}

\maketitle

\vspace{-1mm}
\begin{abstract}
\vspace{-1mm}
Electronic Health Records (EHRs), which contain patients' medical histories in various multi-modal formats, often overlook the potential for joint reasoning across imaging and table modalities underexplored in current EHR Question Answering (QA) systems. In this paper, we introduce \textbf{EHRXQA}, a novel multi-modal question answering dataset combining structured EHRs and chest X-ray images. To develop our dataset, we first construct two uni-modal resources: 1) The MIMIC-CXR-VQA dataset, our newly created medical visual question answering (VQA) benchmark, specifically designed to augment the imaging modality in EHR QA, and 2) EHRSQL (MIMIC-IV), a refashioned version of a previously established table-based EHR QA dataset. By integrating these two uni-modal resources, we successfully construct a multi-modal EHR QA dataset that necessitates both uni-modal and cross-modal reasoning. To address the unique challenges of multi-modal questions within EHRs, we propose a NeuralSQL-based strategy equipped with an external VQA API. This pioneering endeavor enhances engagement with multi-modal EHR sources and we believe that our dataset can catalyze advances in real-world medical scenarios such as clinical decision-making and research. 
EHRXQA is available at \url{https://github.com/baeseongsu/ehrxqa}.

\end{abstract}

\vspace{-2mm}
\section{Introduction}
\label{sec:intro}
\vspace{-2mm}
Electronic Health Records (EHRs) are large-scale databases that store the entire medical history of patients, including but not limited to structured medical records (\textit{e.g.}, diagnosis, procedure, medication), medical images (\textit{e.g.}, chest X-ray, MRI, CT), and clinical text (\textit{e.g.}, discharge summary, nursing note).
This wealth of patient information reveals tremendous clinical knowledge about individual patients and cohorts, marking them as an indispensable resource for healthcare professionals (\textit{e.g.}, physicians, nurses, administrators) in routine clinical practice.

\begin{figure}[t]
    \includegraphics[width=1.0\columnwidth]{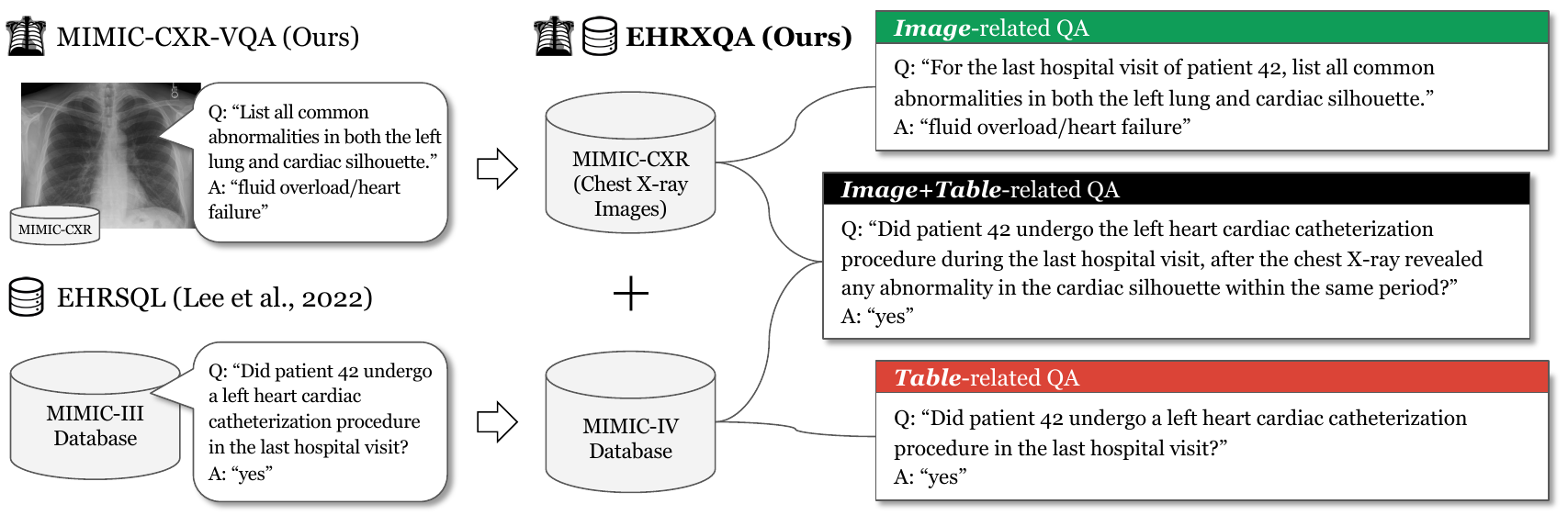}
    \caption{
    Our EHRXQA dataset is constructed from three uni-modal resources: MIMIC-IV for the \textit{table} modality, MIMIC-CXR for the \textit{image} modality, and Chest ImaGenome as a high-quality annotated version of MIMIC-CXR (not shown in the figure).
    Our dataset features questions for individual EHR modalities and those requiring multi-modal reasoning.
    It encompasses three types of QA scope: \textit{Image-}, \textit{Table-}, and \textit{Image+Table-}related QA.
    }
    \label{fig:overview_dataset}
    \vspace{-6mm}
\end{figure}

Recent years have seen an upsurge in research~\cite{lee2022ehrsql,lehman2022learning, pampari2018emrqa, raghavan2021emrkbqa, wang2020text} into question answering (QA) systems for EHRs.
These systems are designed to effectively retrieve information from EHRs, each specializing in a different information modality within the records.
For instance, table-based EHR QA systems can easily retrieve specific information from structured databases and answer questions like ``Did patient 42 undergo a left heart cardiac catheterization procedure in the last hospital visit?'' (see EHRSQL part in \cref{fig:overview_dataset}) by executing an SQL query on the relational database.
On the other hand, image-based EHR QA (\textit{i.e.}, medical visual question answering) models are designed to handle questions related to individual medical images.
For instance, given a question such as ``List all common abnormalities in both the left lung and cardiac silhouette.'' (see MIMIC-CXR-VQA part in \cref{fig:overview_dataset}) along with a patient's chest radiograph, these models generate a response, thereby serving as an effective aid for radiologists.
However, despite their undeniable utility, a main challenge in the current landscape of EHR QA systems lies in their focus on a single information modality, overlooking EHRs' inherently multi-modal nature.
To fully utilize EHRs' potential, it is crucial to develop QA systems capable of seamlessly navigating across these multiple modalities such as ``Did patient 42 undergo the left heart cardiac catheterization procedure during the last hospital visit, after the chest X-ray revealed any abnormality in the cardiac silhouette within the same period?'' (see EHRXQA part in \cref{fig:overview_dataset}).
This capability significantly enhances our ability to build a comprehensive model of a patient's status, thereby improving the quality of the clinical decision-making process.

The progression from uni-modal to multi-modal EHR QA is a promising and self-evident step in the healthcare domain.
Currently, however, only one multi-modal EHR QA dataset~\cite{bardhan2022drugehrqa} integrates structured EHRs with clinical text.
On the other hand, the integration of table modalities with imaging modalities, such as chest X-rays (CXR), remains unexplored~\cite{lin2021medical}.
Our research aims to bridge this gap. 
This has the potential to unlock significant clinical benefits, enhance cross-modal analysis, and catalyze advances in medical research.

To sum up, our contributions are threefold:
\begin{enumerate}[leftmargin=5mm,itemsep=0pt,topsep=-2pt]
    \item[$\bullet$]
        To address the lack of publicly accessible image-based EHR QA datasets that can be combined with structured EHRs, we present MIMIC-CXR-VQA (\cref{sec:mimiccxrvqa}). This is a complex, diverse, and large-scale visual question answering dataset in the medical domain. We not only use its questions as a basis for multi-modal EHR questions, but also exploit this dataset to benchmark existing medical VQA approaches.
    \item[$\bullet$] 
        We present EHRXQA (\cref{sec:ehrxqa}), the first multi-modal EHR QA dataset for table and image modality.
        By leveraging uni-modal resources (\textit{i.e.}, data sources \& question templates), we integrate patients' structured databases with their aligned chest X-ray images, thereby creating a comprehensible set of QA pairs covering \textit{Image}-related, \textit{Table}-related, \textit{Image+Table}-related questions.
    \item[$\bullet$] 
        We propose a NeuralSQL-based approach (\cref{sec:neuralsql}) that integrates Large Language Models (LLMs) with an external VQA application programming interface (API) to handle multi-modal questions over a structured database with images. 
        Despite facing unique challenges of reasoning based on single or multiple images, or even a combination of images and tables, our approach effectively extracts relevant information from multi-modal EHRs in response to natural language queries.
\end{enumerate}

\vspace{-1mm}
\section{Related Work}

\vspace{-2mm}
\paragraph{Image-based EHR Question Answering}
Image-based EHR QA~\cite{hu2023interpretable, huang2023pvqa, huang2022ovqa, kovaleva2020towards} is a distinct subset of medical visual question answering (VQA)~\cite{abacha2019vqa, ben2021overview, hasan2018overview, he2021towards, lau2018dataset, liu2021slake}, given that it focuses on answering questions related to a specific patient's single medical image, primarily within the radiography domain. 
Despite intriguing research directions in existing datasets such as patient-centric QA~\cite{huang2023pvqa} or dialogue~\cite{kovaleva2020towards}, there remains a noticeable gap in efforts to view patient images as an integral part of the EHR database or to synchronize them effectively with the structured tabular data in EHRs.
Presently, MIMIC-CXR~\cite{johnson2019mimiccxr} is the only publicly available imaging resource that links to patient IDs (\textit{i.e.}, subject IDs) in the MIMIC-IV database~\cite{johnson2023mimiciv}, offering a comprehensive perspective on EHRs.
Although there exist two medical VQA datasets~\cite{hu2023interpretable, kovaleva2020towards} based on MIMIC-CXR, neither is publicly available. Moreover, their question templates are less complex (\textit{i.e.}, they lack complex set operations or logical operations) and are largely encompassed by our question template scope.\footnote{In comparison: Hu \etal~\cite{hu2023interpretable} provides around 15 templates across 6 types, Kovaleva \etal~\cite{kovaleva2020towards} offers 1 template across 1 type, while our dataset presents 48 templates across 7 types.}

\vspace{-2mm}
\paragraph{Table-based EHR Question Answering}
Table-based EHR QA~\cite{bae2021question,dobbins2023leafai,lee2022ehrsql,raghavan2021emrkbqa,soni2023quehry,tarbell2023towards,wang2020text} focuses on extracting structured information from a hospital's relational database.
The task is typically approached through semantic parsing~\cite{berant2013semantic}, where natural language utterances are translated into either a query language~\cite{li2023can, yu2018spider, zhong2017seq2sql} or domain-specific logical forms~\cite{raghavan2021emrkbqa,soni2023quehry}.
Wang \etal~\cite{wang2020text} introduced MIMICSQL dataset for the text-to-SQL generation task on MIMIC-III, employing slot-filling for pre-defined templates and using crowd-sourced paraphrasing.
Pampari \etal~\cite{pampari2018emrqa} constructed emrKBQA dataset, a large-scale text-to-logical form dataset tailored for patient-specific QA on MIMIC-III, drawing from the logical forms identified in emrQA~\cite{pampari2018emrqa}.
Recently, Lee \etal~\cite{lee2022ehrsql} introduced a novel text-to-SQL dataset, EHRSQL, associated with both MIMIC-III and eICU~\cite{pollard2018eicu}.
This dataset presents unique challenges, including time-sensitive questions and unanswerable queries.

\vspace{-2mm}
\paragraph{Question Answering over Multi-Modal Knowledge Sources}
Recent research~\cite{chang2022webqa,chen2022murag,chen2023symphony,christmann2022conversational,lu2022learn,singh2021mimoqa,talmor2021multimodalqa,urban2023towards,zhao2022multihiertt} has delved into generating responses to queries using multi-modal knowledge sources.
However, the major challenge when dealing with multi-modal databases, such as EHRs~\cite{bardhan2022drugehrqa}, is integrating rich unstructured data (\textit{e.g.}, image, text) into a structured database (\textit{e.g.}, table) and effectively leveraging this information within the QA system.
Urban \etal~\cite{urban2023towards} introduced MMDBs, a new category of database systems, which allow seamless querying of text and tables using SQL.
Similarly, Chen \etal~\cite{chen2023symphony} proposed Symphony, a QA system for multi-modal data lakes, particularly designed to handle text and tables by using a unified representation for multi-modal datasets.
Drawing inspiration from recent studies like Binder~\cite{cheng2022binding}, a training-free neural-symbolic framework that uses GPT-3 Codex~\cite{chen2021evaluating} to map task inputs to programs, our research broadens the SQL syntax to create a QA system specifically intended for image processing within the database.

\vspace{-1mm}
\section{Preliminary: Ingredients for Multi-Modal EHR QA}
\label{sec:preliminary}

\vspace{-1mm}
\subsection{Uni-Modal Data Resources}
\vspace{-1mm}
To construct a comprehensive EHR database that integrates both \textit{table} and \textit{image} modalities, we need uni-modal resources that meet our criteria: (\romannumeral 1) publicly accessible; (\romannumeral 2) presence of common patients across datasets; (\romannumeral 3) contain high-quality image annotations.
After careful consideration, we strategically select three datasets: MIMIC-IV~\cite{johnson2023mimiciv} for \textit{table} modality, MIMIC-CXR~\cite{johnson2019mimiccxr} for \textit{image} modality, and Chest ImaGenome~\cite{wu2chestimagenome} as a high-quality annotated version of MIMIC-CXR. 
Note that all datasets share a significant number of patient IDs (19,264), while incompatible patient IDs exist due to the varying data collection periods.
We briefly introduce each of the source datasets.\footnote{All three datasets are publicly accessible through the PhysioNet platform ({\url{https://physionet.org/}}), with users required to request and obtain credentialed access under its established procedure.}

\begin{itemize}[itemsep=1pt, topsep=-2pt, leftmargin=*]
    \item 
        \textbf{MIMIC-IV (v2.2)}~\cite{johnson2023mimiciv} is a large, freely accessible relational database of deidentified health-related data (\eg, diagnoses, procedures, and treatments) associated with 50,920 patients who stayed in critical care units of Beth Israel Deaconess Medical Center (BIDMC) between 2008-2019. 
    \item 
        \textbf{MIMIC-CXR (v2.0.0)}~\cite{johnson2019mimiccxr} is a large-scale publicly available dataset of 377,110 chest radiographs associated with 227,827 imaging studies sourced from the BIDMC between 2011-2016. 
        MIMIC-CXR can be linked to MIMIC-IV using lookup tables that connect patient identifiers.
    \item 
        \textbf{Chest ImaGenome (v1.0.0)}~\cite{wu2chestimagenome}, organized with scene graphs for 242,072 frontal images sourced from MIMIC-CXR, illustrates the relationships between anatomical locations and their corresponding attributes within each image.
        This dataset comprises two primary subsets: the \textit{silver}\footnote{For the \textit{silver} dataset, given the high inter-annotator agreement score (0.984 for 500 reports)~\cite{wu2chestimagenome}, the reliability is strongly suggested. This score substantiates the decision to use the \textit{silver} dataset for building our MIMIC-CXR-VQA and EHRXQA, providing confidence in the accuracy and quality of the derived information.} dataset with automatically generated scene graphs for each chest X-ray image, and the \textit{gold} dataset containing a subset that has been manually validated and corrected by clinicians, serving as a reliable held-out set for research derived from 500 unique patients. 
\end{itemize}

\vspace{-1mm}
\subsection{Uni-Modal EHR QA datasets}
\label{sec:preliminary_unimodal_qa}
\vspace{-1mm}
We aim to build a multi-modal EHR QA dataset featuring questions for each modality individually, as well as those that require cross-modal reasoning. 
To achieve this, we utilize uni-modal QA datasets based on MIMIC nature.
For \textit{table} modality, we take the existing questions templates from EHRSQL~\cite{lee2022ehrsql}, and adapt them to MIMIC-IV.
For \textit{image} modality, to address the lack of diverse question templates and the absence of accessible VQA datasets based on MIMIC-CXR, we craft our templates and further construct a medical VQA dataset called \textbf{MIMIC-CXR-VQA}~(\cref{sec:mimiccxrvqa}).

\vspace{-1mm}
\subsubsection{Table-based EHR QA: EHRSQL}
\label{sec:ehrsql}
\vspace{-1mm}
EHRSQL~\cite{lee2022ehrsql} is a text-to-SQL dataset curated for structured EHRs, assembled from the responses of various hospital staff.
EHRSQL provides (Question, SQL) samples for two publicly accessible EHR datasets, namely MIMIC-III~\cite{johnson2016mimiciii} and eICU~\cite{pollard2018eicu}, and samples consist of both \textit{answerable} and \textit{unanswerable} questions.
Since our research scope primarily focuses on building a multi-modal QA dataset, we have selected only the \textit{answerable} question templates from EHRSQL for MIMIC-III.
These templates were converted to align with our MIMIC-IV setting, while maintaining their comprehensive template schema, including multiple value slots (\eg, operation and condition value slots) and time filter slots.
For more details about the conversion process of question templates from MIMIC-III to MIMIC-IV, please refer to \cref{supp_prelim_tabdataset}.

\vspace{-1mm}
\subsubsection{Image-based EHR QA: MIMIC-CXR-VQA}
\label{sec:mimiccxrvqa}
\vspace{-1mm}

\begin{wrapfigure}{r}{0.40\textwidth}
\vspace{-5mm}
\caption{
Upper: Scene graphs of multiple CXR studies derived from the Chest ImaGenome.
Lower: Our processed CXR features, obtained from these scene graphs.
Due to spatial constraints, only a subset of the original Chest ImaGenome labels is displayed.
}
\label{fig:chestimagenome_overview}
\vspace{-2mm}
\centering
\includegraphics[width=0.40\textwidth]{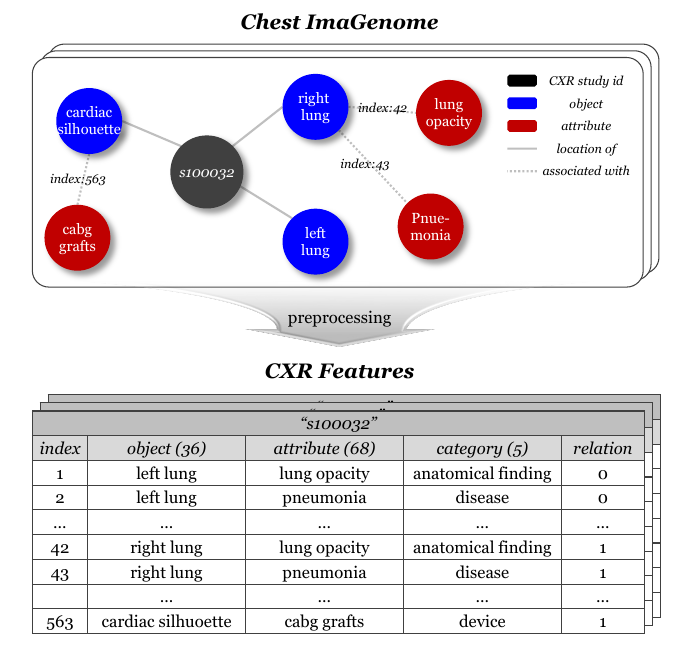}
\vspace{-10mm}
\end{wrapfigure}

\paragraph{Data Preprocessing}\hspace{-3mm}
\label{sec:mimiccxrvqa_data_preprocessing}
We use MIMIC-CXR~\cite{johnson2019mimiccxr} as our image source and Chest ImaGenome~\cite{wu2chestimagenome} for label information.
In MIMIC-CXR, each patient can have multiple studies arranged in chronological order, and each study can contain multiple CXR images. 
From each study, we select one representative frontal view (\ie, AP, PA) image.
We then assign labels to these images derived from the Chest ImaGenome \textit{silver}/\textit{gold} datasets.
As a result, each CXR image features 563 distinct \textit{relations} among 36 \textit{objects}, each linked to several attributes from a pool of 68 \textit{attributes} (across 5 \textit{categories}\footnote{The 5 categories include `anatomical finding', `disease', `device', `tubes/lines', `technical assessment'.}).
As illustrated in Figure~\ref{fig:chestimagenome_overview}, each \textit{relation} indicates the presence (1) or absence (0) of an \textit{attribute} (\textit{e.g.}, lung cancer) within a \textit{category} (\textit{e.g.}, disease), linked to an \textit{object} (\textit{e.g.}, left lung).
For data splitting, we use the machine-generated \textit{silver} label dataset for training and validation, with a 95:5 split, while the human-labeled \textit{gold} dataset serves as the testing dataset.
For more details of data preprocessing, please refer to \cref{supp_prelim_imgdataset}.

\vspace{-2mm}
\paragraph{Question Template Construction}\hspace{-3mm}
\label{sec:mimiccxrvqa_template_construction}
We started by analyzing existing medical VQA datasets~\cite{abacha2019vqa, ben2021overview, hasan2018overview, he2021towards, lau2018dataset, liu2021slake} and templatized their questions to match our preprocessed data schema (\textit{i.e.}, object, attribute, category), thus handcrafting our initial seed templates.
We drew inspiration from general VQA datasets~\cite{antol2015vqa,gokhale2020vqa,hudson2019gqa,johnson2017clevr}, enhancing these seed templates using logical and set operations to create a more diverse and complex set of question templates.
We further incorporated clinically relevant factors~\cite{lau2018dataset} into our templates, such as the patient's gender, CXR view position, and size-related features (\textit{i.e.}, width ratio between two anatomical locations).
As a result, we defined a total of 48 templates, all of which were evaluated by a medical expert for clinical importance.
For more details about template construction including a list of our templates, please refer to \cref{supp_prelim_imgdataset}.

\vspace{-2mm}
\paragraph{VQA dataset generation}\hspace{-3mm} 
We generated our VQA dataset by sampling (image \textit{I}, question \textit{Q}, answer \textit{A}) triples. 
For example, consider the template ``Is there \$\{attribute\} in the \$\{object\}?''.
We filled this template using sampled arguments (\textit{e.g.}, \$\{object\}=`left lung', \$\{attribute\}=`lung cancer'), which led to the creation of the question $Q$: ``Is there lung cancer in the left lung?''.
Next, we sampled an image $I$ and executed a predefined program\footnote{For each template, we define a program to produce an answer $A$ using the given question $Q$ and relationship information from the preprocessed data (see \cref{sec:mimiccxrvqa_data_preprocessing})  of the image $I$.} to generate an answer $A$.
To enrich linguistic diversity while preserving focus on the medical domain~\cite{nori2023capabilities}, we devised a paraphrasing strategy (an average of 16.5 paraphrases for each template) using carefully designed prompts based on GPT-4~\cite{openai2023gpt4}. 
Finally, we present \textbf{MIMIC-CXR-VQA}, a dataset composed of 377,391 unique ($I, Q, A$) triples across seven content types\footnote{
Questions are divided into 7 categories based on the content of the question: `presence', `anatomy', `attribute', `abnormality', `size', `plane', `gender'.
}. 
For a deeper dive into the statistics of MIMIC-CXR-VQA and its comparisons to other medical VQA datasets, please refer to \cref{supp_prelim_imgdataset}.

\vspace{-2mm}
\section{EHRXQA: A Multi-Modal EHR Question Answering Dataset}
\label{sec:ehrxqa}

\vspace{-1mm}
\subsection{Dataset Construction}
\vspace{-1mm}
In this section, we outline the construction process for the EHRXQA dataset. 
We begin by integrating CXR images from MIMIC-CXR and tables from MIMIC-IV into our EHRXQA database (see \cref{sec:ehrxqa_database_construction}). Next, we detail the creation of question templates (see \cref{sec:ehrxqa_question_template_construction}), and the incorporation of the corresponding SQL/NeuralSQL annotations (see \cref{sec:ehrxqa_annotation}). Finally, we discuss our systematic data generation process (see \cref{sec:ehrxqa:qa_dataset_generation}) employed to build our EHRXQA dataset.

\vspace{-1mm}
\subsubsection{Database Construction}
\label{sec:ehrxqa_database_construction}
\vspace{-1mm}

\textbf{CXR Integration into MIMIC-IV}\;\;
To cross-reference CXR images with structured EHRs (\textit{e.g.}, to find CXR images of patients who have been prescribed a specific drug), an integrated database system is crucial.
To achieve this, we developed an image reference table named \texttt{TB\_CXR}. 
This table comprises six columns: \texttt{subject\_id}, \texttt{hadm\_id}, \texttt{study\_id}, \texttt{image\_id}, \texttt{studydatetime}, and \texttt{viewposition}, connecting patient-related identifiers with CXR images of MIMIC-CXR. 
Through this table, patient CXR images can be retrieved alongside other table data (\textit{e.g.}, diagnosis, procedure, and prescriptions) from MIMIC-IV using the \texttt{subject\_id} or \texttt{hadm\_id}. 
For more details on the database construction process, please refer to \cref{supp_EHRXQA_dataconstruct}.

\textbf{Timeframe Adjustment}\;\;
We condensed the event times in each patient's records, which originally spanned from 2100 to 2200 due to the de-identification process in MIMIC-IV~\cite{johnson2023mimiciv}, to a more realistic timeframe (2100-2105).
This adjustment was performed while preserving the integrity of CXR images and individual medical event timelines. 
To enable relative time expressions like `last year', we set `2105-12-31 23:59:00' as the \textit{current time} and excluded any records beyond this point. 
We consider patients without hospital discharge times, due to this exclusion, as currently admitted.

\textbf{Building Silver/Gold Databases}\;\;
The Chest ImaGenome~\cite{wu2chestimagenome} dataset includes two types of cohorts based on image information: \textit{silver} (\ie, machine-generated) and \textit{gold} (\ie, human-labeled). We selected subsets of patients from each cohort to create two distinct databases: the \textit{silver} database, comprising 800 patients, and the \textit{gold} database, comprising 400 patients. These databases are utilized for different purposes: the \textit{silver} database is used for training and validating the QA dataset, while the \textit{gold} database is used for testing the QA dataset.

\begin{table}[t!]
\caption{
    Sample questions in EHRXQA, categorized by \textbf{modality-based} (\textit{Image}, \textit{Table}, \textit{Image+Table}) and \textbf{patient-based} scope (\textit{none}, \textit{single}, \textit{group}), illustrating our dataset's diversity and complexity.
}
\label{tab:sample-table}
\centering
\renewcommand{\arraystretch}{1.0}
\begin{adjustbox}{width=\columnwidth,center}
\begin{tabular}{cccl}
    \toprule
    \textbf{modality-based} & 
    \multicolumn{2}{c}{\textbf{patient-based}} & 
    \multicolumn{1}{c}{\textbf{Sample question}} 
    \\ \midrule
    \multirow{7}{*}{\textit{Image}} & \multirow{6}{*}{\textit{single}} & 1-\textit{image} & 
    \begin{tabular}[c]{@{}l@{}}
    Given the last study of patient 15439, which anatomical finding is associated with the right lower lung zone,\\pneumothorax or vascular redistribution?\end{tabular}
    \\ \cmidrule(lr){3-4} 
    & & 2-\textit{image} & 
    \begin{tabular}[c]{@{}l@{}}
    Enumerate all diseases that are newly detected based on the last study of patient 19290 in 2103 compared to\\the previous study. \end{tabular}\\ \cmidrule(lr){3-4} 
    & & N-\textit{image} & 
    \begin{tabular}[c]{@{}l@{}}How many times has the chest X-ray of patient 18489 shown linear/patchy atelectasis in the left lung on the\\current hospital visit? \end{tabular}
    \\ \cmidrule(lr){2-4} 
    & \multicolumn{2}{c}{\textit{group}} &   
    Count the number of patients whose chest X-ray studies this year showed any abnormalities in the mediastinum.
    \\ \cmidrule(lr){1-4} 
    \multirow{4}{*}{\textit{Table}} & 
        \multicolumn{2}{c}{\textit{none}}
        & What's the cost of a drug named lopinavir-ritonavir? 
        \\ \cmidrule(lr){2-4} & 
        \multicolumn{2}{c}{\textit{single}} & 
        Did patient 16164 receive any magnesium lab tests last year?       
        \\ \cmidrule(lr){2-4} & 
        \multicolumn{2}{c}{\textit{group}} & 
        What was the top three diagnosis that had the highest two year mortality rate?
        \\ \cmidrule(lr){1-4}
    \multirow{2.5}{*}{\textit{Image+Table}} & 
        \multicolumn{2}{c}{\textit{single}} & 
        \begin{tabular}[c]{@{}l@{}}
        Did a chest X-ray study for patient 15110 reveal any anatomical findings within 2 month after the prescription\\of hydralazine since 2102?
        \end{tabular}
        \\ \cmidrule(lr){2-4} & 
        \multicolumn{2}{c}{\textit{group}} & 
        Provide the ids of patients in the 20s whose chest X-ray showed low lung volumes in the right lung this month.
        \\ \bottomrule 
\end{tabular}
\end{adjustbox}
\vspace{-5mm}
\end{table}

\subsubsection{Question Template Construction}
\label{sec:ehrxqa_question_template_construction}
We define the scope of our question templates using two key criteria: \textbf{modality-based} and \textbf{patient-based} scopes. 
The \textbf{modality-based} scope classifies templates into three categories, \textit{Image}-related, \textit{Table}-related, and \textit{Image+Table}-related, depending on the type of data modality they require.
The \textbf{patient-based} scope classifies templates according to whether they relate to a \textit{single} patient, a \textit{group} of patients, or \textit{none} (\textit{i.e.}, do not relate to specific patients). 
To accommodate these scopes with diverse and comprehensive question templates, we employ existing uni-modal question resources discussed in \cref{sec:preliminary_unimodal_qa}: MIMIC-CXR-VQA for \textit{image} modality and EHRSQL for \textit{table} modality. 
Examples of our \textit{modality}- and \textit{patient}-based question templates, which illustrate the diversity and complexity of EHRXQA dataset, can be found in \cref{tab:sample-table}.

Recognizing the critical role of time expressions in real-world questions in the hospital workplace~\cite{lee2022ehrsql}, we further refined our question templates.
We adopted the time filter concept from EHRSQL and applied it to all question templates.
This enhancement allows our question templates to better meet the specific needs in clinical practice.
Note that these time filters can be categorized into three types: 1) $\text{[time\_filter\_global]}$ restricts the time range of interest, such as `last year' or `in 2022'; 
2) $\text{[time\_filter\_within]}$, incorporating the keyword `within', pinpoints events happening within specific temporal boundaries, such as `within the same hospital visit' or `within the same day';
3) $\text{[time\_filter\_exact]}$ refers to a precise temporal point, such as the `last CXR study' or a specific date and time like `2105-12-26 15:00:00'.

Our template construction process included 1) clinical needs across both image and table modalities via consulting a medical expert, 2) grounding our templates in these needs for both CXR images and EHR tables, and 3) ensuring clinical relevance. Note that the entire process of designing templates was validated by a board-certified medical expert from the department of neurosurgery to ensure clinical utility. For a full list or an in-depth discussion on template construction strategy, please refer to \cref{supp_EHRXQA_templateconstruct}. The following details how we tailored question templates for each modality.

\vspace{-1mm}
\paragraph{\textit{Image}-related}
Questions related to \textit{image} modality can be defined as inquiries requiring pixel-level information from CXR images retrieved from EHR, which can aid in analyzing visual diagnoses for individual or cohort patient conditions in real-world medical scenarios.
To cater to these queries, we used the 48 MIMIC-CXR-VQA templates (\textit{e.g.}, ``List all diseases.'') and integrated with expressions to specify our target images (\textit{e.g.}, ``The last study of patient 42''). 
This integration (\textit{e.g.},``Given the last study of patient 42, list all diseases.'') enables retrieval of CXR images from the EHR and subsequent analysis based on natural language requests.
We further enhanced the templates focusing on a \textit{single} patient to include queries that compare two consecutive CXR studies (\textit{e.g.}, ``Given the last study of patient 42, are there any newly detected diseases compared to the previous study?'') or multiple studies (\textit{e.g.}, ``Has patient 42 had any chest X-ray study indicating any anatomical findings in 2023?'') from the same patient.
This process resulted in 168 templates for the \textit{image} modality.

\vspace{-1mm}
\paragraph{\textit{Table}-related}
The \textit{table} modality, a significant part of EHRs, covers questions primarily requiring structured information from EHR tables.
These questions relate to patient demographics, diagnoses, procedures, medications, and other clinical details typically recorded in structured EHR formats.
EHRSQL, which offers a wealth of questions seeking information from EHR tables, proves to be an invaluable resource in this context. 
Considering the substantial overlap between the MIMIC-III and MIMIC-IV schemas, we leveraged the question templates from EHRSQL's MIMIC-III templates, adapting them appropriately to fit the MIMIC-IV schema with minimal modifications.
This process resulted in 174 templates for the \textit{table} modality.

\vspace{-1mm}
\paragraph{\textit{Image+Table}-related}
In the \textit{image+table} modality, all templates are designed to require multi-modal information from both CXR images and structured data from EHRs.
We leveraged both MIMIC-CXR-VQA and EHRSQL templates to build multi-modal question templates. 
Since we recognize the essential role of temporal analysis in multi-modal medical events, we designed templates to capture three primary scenarios:
1) Co-occurring table and CXR events. 
(\textit{e.g.}, ``On the same visit, did patient 42 receive nitroglycerin and have a CXR showing any abnormality in the cardiac silhouette?'');
2) A CXR event following a table event. 
(\textit{e.g.}, ``After being prescribed nitroglycerin, did patient 42 have a CXR during the same visit revealing any abnormality in the cardiac silhouette?'')
3) A table event following a CXR event. 
(\textit{e.g.}, ``Was patient 42 prescribed nitroglycerin during the same visit after a CXR showed cardiac silhouette abnormalities?'').
These templates allow for comprehensive analysis of combined events, the cause-and-effect relationships in CXR diagnosis, and relevant follow-up measures related to the CXR diagnosis. 
To eliminate confusion arising from overlapping information between the CXR and diagnoses/procedures tables, we ensure that questions explicitly specify when a `CXR study' is necessary.
This led to 75 templates for the \textit{image+table} modality, enabling simulations across diverse scenarios.

\begin{figure}[t]
    \includegraphics[width=1.0\columnwidth]{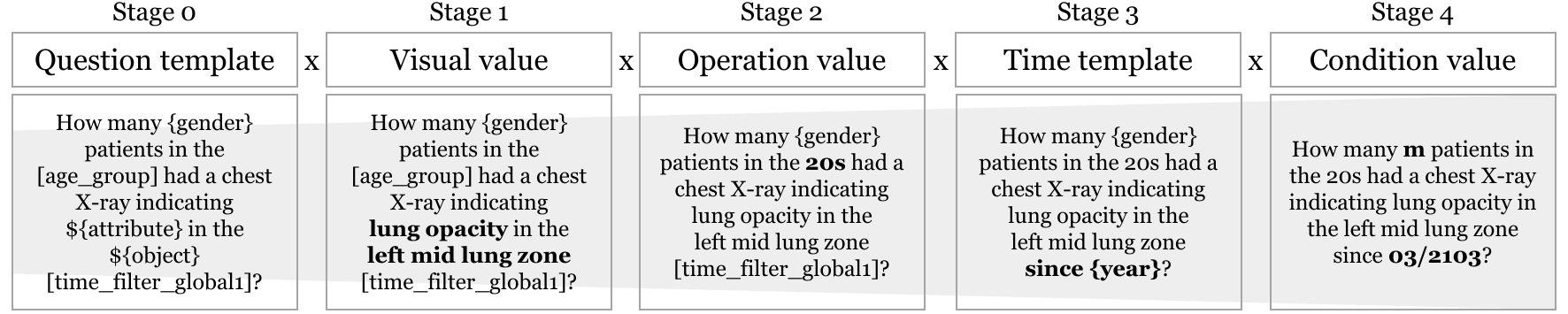}
    \vspace{-5mm}
    \caption{QA data generation process}
    \label{fig:qa_generation}
    \vspace{-4mm}
\end{figure}

\subsubsection{SQL/NeuralSQL Annotation}
\label{sec:ehrxqa_annotation}
Standard SQL queries are effective for retrieving structured data from EHRs~\cite{wang2020text,lee2022ehrsql}, such as demographic information or lab results stored in tables. 
However, they are not designed to handle unstructured data, such as CXR images, which also contain valuable patient information.
This limitation prevents us from using SQL to retrieve answers for complex, multi-modal questions that span both structured and unstructured data.
To overcome this limitation, we adopt NeuralSQL, which is inspired by the Binder approach~\cite{cheng2022binding}.
NeuralSQL acts as an executable representation, extending SQL's capabilities to process unstructured image data.
NeuralSQL utilizes a pretrained neural model to extract features from medical images, turning them into a structured format suitable for SQL queries. 
For more details about our NeuralSQL-based strategy, please refer to \cref{sec:neuralsql}. 

For \textit{Table}-related question templates, we utilize the SQL annotations provided by EHRSQL and modify them to be compatible with the MIMIC-IV schema.
For question templates related to \textit{Image} or \textit{Image+Table}, we annotate them using NeuralSQL representation.
The entire SQL/NeuralSQL annotation process was manually undertaken by four graduate students over a span of two months, involving iterative revisions. 
During this process, the students transformed question templates into their corresponding SQL or NeuralSQL formats.

\subsubsection{Data Generation}
\label{sec:ehrxqa:qa_dataset_generation}
The question generation process, illustrated in \cref{fig:qa_generation}, begins with choosing a template at Stage 0, followed by a four-step systematic process (Stages 1-4) that specifies semantics of the template.
These steps involve the sampling of \textit{visual value} (Stage 1), \textit{operation value} (Stage 2), \textit{time template} (Stage 3), and \textit{condition value} (Stage 4).
In Stage 1, we augment the question with \textit{visual values} by filling in object, attribute, and category slots (described in \cref{sec:mimiccxrvqa_template_construction}), tailored specifically for CXR images.
Stage 2 involves sampling \textit{operation values} (\textit{e.g.}, 20s) from a predefined set of options such as [age\_group] = (20s, 30s, 40s, 50s, 60 or above), which are independent of the database schema or records.
Stage 3 incorporates \textit{time templates}, translated into natural language expressions to establish a temporal context within the questions.
Lastly, Stage 4 incorporates \textit{condition value} sampling, filling placeholders such as \{gender\} and \{year\} to provide context-specific conditions to the question.

The corresponding SQL/NeuralSQL query also contains these slots, filled with the same values during the question creation process, thereby completing the (Question, SQL/NeuralSQL) pair.
These (Question, SQL/NeuralSQL) pairs are only added to the data pool if the sampled SQL/NeuralSQL query yields a valid answer when executed.
To enhance linguistic diversity, we use GPT-4 to paraphrase each question. 
These paraphrases are then manually reviewed by our team to ensure quality.
Further details can be found in the \cref{supp_EHRXQA_generation}.

\begin{figure}[t]
    \centerline{
    \includegraphics[width=0.9\columnwidth]{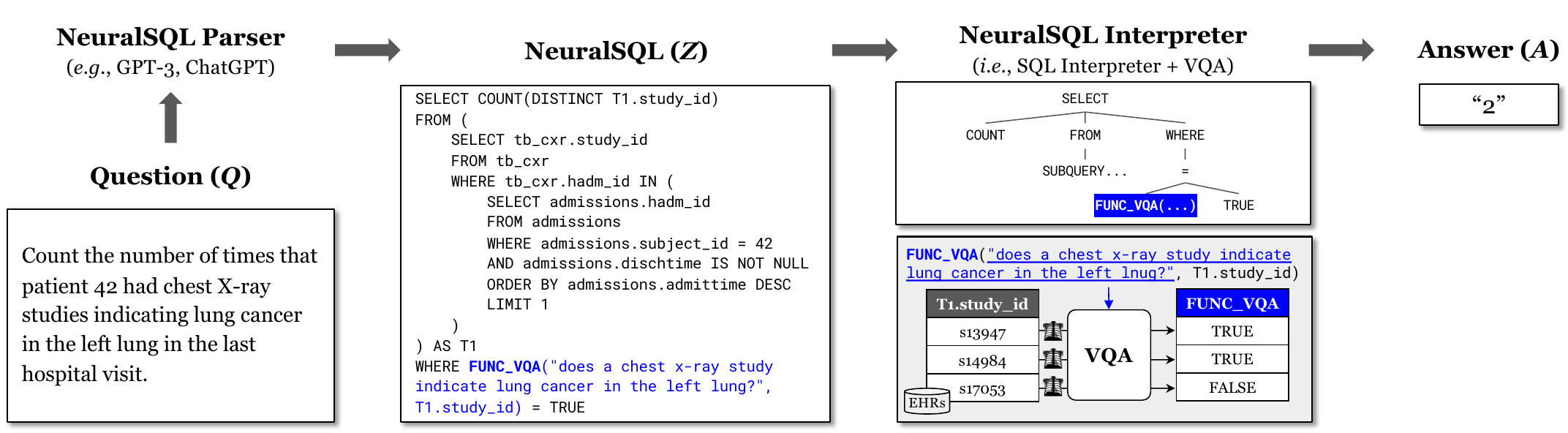}
    }
    \caption{Overview of our NeuralSQL-based Approach.}
    \label{img:nsql_overview}
    \vspace{-5mm}
\end{figure}

\begin{wraptable}[7]{r}{0.33\textwidth}
\vspace{-5mm}
        \captionsetup{font={small}}
        \caption{
        Overall statistics of EHRXQA including the number of samples for each modality.
        }
    \resizebox{1.0\linewidth}{!}{
        \begin{tabular}{lccc}
        \toprule
                                        & train     & valid     & test      \\ \midrule
    \textit{image}-related QA           & 12,860    & 1,838     & 1,668     \\
    \textit{table}-related QA           & 12,961    & 1,852     & 1,716     \\
    \textit{image+table}-related QA     & 10,353    & 1,480     & 1,424     \\
    total \# of samples                 & 36,174    & 5,170     & 4,808     \\
    \bottomrule
    \end{tabular}
}
\label{tab:xqastatistics}
\end{wraptable}

\begin{table}[t]
\caption{
A comparison of EHRXQA with other EHR QA datasets based on the MIMIC database.
}
\label{tab:ehrqa_comparison}
\centering
\resizebox{\textwidth}{!}{%
\begin{tabular}{lcccccccc}
\toprule
    & Data source 
    & Image 
    & Table 
    & Text
    & Patient scope 
    & \# of tables / DB 
    & \# of questions  
    & compositional
    \\ \midrule
Mimic-VQA~\cite{hu2023interpretable} 
    & MIMIC-CXR
    & {\color{teal}\checkmark}
    & -
    & -
    & single
    & -
    & 297,723
    & -
    \\
MIMIC-CXR-VQA (ours)
    & MIMIC-CXR, Chest ImaGenome
    & {\color{teal}\checkmark}
    & -
    & -
    & single              
    & -                 
    & 377,726
    & {\color{black}\checkmark}
    \\ \midrule
MIMICSQL~\cite{wang2020text}
    & MIMIC-III             
    & -
    & {\color{red}\checkmark}    
    & -             
    & none, single, group 
    & 5                 
    & 10,000
    & {\color{black}\checkmark}
    \\
EHRSQL~\cite{lee2022ehrsql}
    & MIMIC-III, eICU
    & -
    & {\color{red}\checkmark}
    & -
    & none, single, group 
    & 13.5
    & 24,411
    & {\color{black}\checkmark}
    \\ \midrule
DrugEHRQA~\cite{bardhan2022drugehrqa}
    & MIMIC-III
    & -
    & {\color{red}\checkmark}
    & {\color{black}\checkmark}
    & single
    & 3
    & 70,381
    & -
    \\ 
EHRXQA (ours)
    & MIMIC-IV, MIMIC-CXR, Chest ImaGenome
    & {\color{teal}\checkmark}
    & {\color{red}\checkmark}
    & -
    & none, single, group 
    & 18
    & 46,152
    & {\color{black}\checkmark}
    \\ \bottomrule
\end{tabular}%
}
\vspace{-5mm}
\end{table}

\subsection{Data Statistics and Comparisons with other EHR QA datasets}
\vspace{-2mm}
EHRXQA consists of a total of 46,152 samples including 16,366 \textit{image}-related samples, 16,529 \textit{table}-related samples, and 13,257 samples involving both \textit{images} and \textit{tables}. 
Overall Statistics are summarized in \cref{tab:xqastatistics}.
For a comprehensive breakdown of the dataset's distribution across various modalities and patient scopes, please refer to \cref{supp_data_analysis}.

\cref{tab:ehrqa_comparison} provides a comparison of EHRXQA with other EHR QA datasets based on the MIMIC database.
Compared to other image-based EHR QA datasets (rows 1-2), EHRXQA incorporates information from EHR tables. 
This allows for more complex queries about images, such as comparing the clinical condition of two specific images. 
This feature extends beyond the existing VQA scope and aims to maximize EHR data utilization. 
Compared with table-based EHR QA datasets (rows 3-5), EHRXQA shows the most complex data structure, featuring up to 18 tables per database and a comprehensive range of patients. 
These features broaden the spectrum of potential questions that can be posed. 
To the best of our knowledge, EHRXQA is the first attempt to merge image and tabular modalities in medical QA.

\vspace{-2mm}
\section{NeuralSQL with Visual Question Answering}
\label{sec:neuralsql}
\vspace{-3mm}
EHRXQA presents three unique challenges for EHR QA systems that handle both image and table modalities:
1) retrieving and analyzing a single image from the database solely based on natural language expressions;
2) handling multiple images, which include comparative queries across multiple studies;
3) and reasoning across multi-modal data over tables and images.
To overcome these challenges, we introduce a NeuralSQL-based approach, inspired by the Binder~\cite{cheng2022binding} framework. 
Our approach integrates a large language model (LLM)-based parser with an external VQA API module, effectively handling both structured information and images.
As depicted in~\cref{img:nsql_overview}, the NeuralSQL-based approach consists of two stages: 
\begin{enumerate}
[leftmargin=4mm,itemsep=-2pt,topsep=-2pt]
    \item \textbf{NeuralSQL Parsing}:
    Given a database $D$ and question $Q$, the parser model translates the question $Q$ to an executable NeuralSQL query $Z$. 
    Note that for all \textit{Image}-related and \textit{Image+Table}-related questions, we annotated the corresponding NeuralSQL query, as discussed in \cref{sec:ehrxqa_annotation}.
    This query features a specific VQA API call function (\texttt{FUNC\_VQA}), which handles image-related queries by calling an external VQA model. 
    This API function requires two arguments: (1) a subquestion, $q_{I}$, which seeks information related to the image, and (2) the relevant image identifier, $c_I$, linking to the \texttt{study\_id} column in \texttt{TB\_CXR}.
    \item \textbf{NeuralSQL Execution}:
    This execution stage involves parsing the NeuralSQL query into an abstract syntax tree (AST), guided by the extended grammar. 
    During this process, the interpreter executes the parsed tree in sequence, including any API calls.
    Upon encountering a VQA API call, the interpreter employs an internal image loader for the corresponding image(s) $I$ based on $c_{I}$. 
    These image(s) are then fed into the VQA model, which infers the information based on the provided question $q_{I}$ and image(s) ${I}$. 
    The output of the API call is preserved as a column data object, making it compatible with the standard SQL grammar. 
    This allows the NeuralSQL interpreter to execute the program seamlessly and derive the final answer $A$.
\end{enumerate}

\vspace{-2mm}
\section{Experiments}
\vspace{-2mm}
In this section, we evaluate medical visual question answering methods on our MIMIC-CXR-VQA dataset (\cref{exp:mimiccxrvqa}). Subsequently, we use the best-performing model as an external VQA API for benchmarking our EHRXQA dataset (\cref{exp:ehrxqa}).

\vspace{-2mm}
\subsection{MIMIC-CXR-VQA}
\label{exp:mimiccxrvqa}

\begin{wraptable}{r}{80mm}
\setlength{\tabcolsep}{3pt}
\vspace{-4mm}
\centering
\captionsetup{font={small}}
\caption{Performance of five baselines on MIMIC-CXR-VQA. To ensure a fair comparison, we pre-trained VLP models (indicated by $\ast$) using the same corpus.}
\resizebox{\linewidth}{!}{%
    \begin{tabular}{@{}cccccc@{}}
    \toprule
    \multirow{2}{*}{Model} & \multicolumn{2}{c}{Valid set} & \multicolumn{3}{c}{Test set} \\
    \cmidrule(l){2-6}
                           & Acc & F1 (micro) & Acc & F1 (micro) & AUC$_{\text{rel}}$ \\
    \midrule
    Prior (Most)~\cite{antol2015vqa}   & 26.8 & 0.27 & 25.4 & 0.25 & - \\
    Prior (Question)~\cite{antol2015vqa} & 34.3 & 0.34 & 32.4 & 0.32 & - \\
    PubMedCLIP~\cite{eslami2023pubmedclip} & $55.1 \pm 1.7$ & $0.56 \pm 0.02$ & $54.9 \pm 1.3$ & $0.54 \pm 0.02$ & $0.82 \pm 0.09$ \\
    PubMedCLIP$^{\ast}$ & $56.6 \pm 1.9$ & $0.58 \pm 0.02$ & $56.5 \pm 2.1$ & $0.56 \pm 0.02$ & $0.83 \pm 0.09$ \\
    MedViLL$^{\ast}$~\cite{moon2022multi} & $64.7 \pm 0.2$ & $0.69 \pm 0.00$ & $63.6 \pm 0.1$ & $0.67 \pm 0.00$ & $0.98 \pm 0.08$ \\
    M$^{3}$AE~\cite{chen2022multim3ae} & $68.9 \pm 0.2$ & $0.73 \pm 0.00$ & $68.9 \pm 0.3$ & $0.72 \pm 0.00$ & $1.02 \pm 0.08$ \\
    \textbf{M$^{3}$AE$^{\ast}$} & $70.2 \pm 0.1$ & $0.74 \pm 0.00$ & $69.2 \pm 0.4$ & $0.73 \pm 0.00$ & $1.05 \pm 0.09$ \\
    \bottomrule
    \vspace{-9mm}
    \end{tabular}%
}
\label{tab:baseline}
\end{wraptable}

\vspace{-2mm}
\paragraph{Task \& Evaluation}
We define the VQA task as a multi-label classification with 110 distinct answer labels.
This includes 36 objects, 68 attributes, and 4 extras (\textit{i.e.}, `M', `F', `AP', `PA'), as well as `yes' and `no' responses.

In MIMIC-CXR-VQA, \textit{verify} questions (\textit{i.e.}, ``Is there \$\{attribute\} in \$\{object\}?'') test a model's basic perception, while other questions demand a logical combination of corresponding perception abilities. 
Therefore, both perception and logical combination are necessary to solve our QA dataset. 
However, unlike logical operations with clear answers, even radiologists cannot achieve perfect perception accuracy in CXRs~\cite{brady2012discrepancy,brady2017error}. 
Thus, it is very likely that the upper bound QA performance of MIMIC-CXR-VQA is lower than 100\%. 
We thus aim to estimate the highest achievable perception accuracy for single-image \textit{verify} questions as a reference score. 
To simplify the problem, we design a reference model as a classification model that can answer our basic \textit{verify} questions. 
We propose the performance of this model as a reference score for perception performance and introduce a new metric by comparing this reference score with the performance of the VQA model. 
For each object-attribute pair $(o, a)$, $m_{rel}(o, a) = \frac{m_{VQA}(o, a)}{m_{ref}(o, a)}$
where $o$ and $a$ denote a specific object and attribute. 
$m_{VQA}$ and $m_{ref}$ denote the metric scores of the VQA model and the reference model, and $m_{rel}$ is our proposed relative metric. 
We use Area Under the Receiver Operating Characteristic (AUROC) as our measure $m$ (denote as AUROC$_{rel}$). 
We provide a comprehensive evaluation of the model, not only our relative score, but also standard metrics like accuracy and F1 score.
For further details on the reference model, please refer to \cref{supp_exp_vqa}.

\vspace{-2mm}
\paragraph{VQA Baselines} 
We evaluate five VQA baselines: two prior models~\cite{antol2015vqa}, PubMedCLIP~\cite{eslami2023pubmedclip}, MedViLL~\cite{moon2022multi}, and \MAE~\cite{chen2022multim3ae}. 
Prior (Most) or Prior (Question) returns the most probable answer estimated from the entire training set or the corresponding question. 
PubMedCLIP, MedViLL, and $\text{M}^3\text{AE}$ are vision-language pre-training (VLP) models, each leveraging unique pre-training objectives and architectures.
To ensure a fair comparison, we pre-trained all models on the same MIMIC-CXR (image, report) pre-training corpus, with those models denoted by an asterisk ($\ast$).
For more details, please refer to \cref{supp_exp_vqa}.

\vspace{-2mm}
\paragraph{Results and Findings} \label{sec:vqa_eval}
\cref{tab:baseline} presents the baseline results on MIMIC-CXR-VQA dataset. The model Prior (Question), which depends solely on language, yields an accuracy of around 30\%. This result attests to the reduced language bias in our dataset, emphasizing the importance of multi-modal reasoning. 
Among the models evaluated, $\text{M}^3\text{AE}$ achieves the best performance, likely due to its more fine-grained pre-training objectives compared to PubMedCLIP and MedViLL. 

\vspace{-2mm}
\subsection{EHRXQA}
\label{exp:ehrxqa}

\vspace{-2mm}
\paragraph{Task}
We use semantic parsing to bridge natural language and machine-executable language. 
The \textit{Image}-related and \textit{Image+Table}-related QA scopes are formulated as a Text-to-NeuralSQL task, facilitating complex queries across images and tables. 
The \textit{Table}-related QA scope, focusing solely on tabular data, is tackled as a Text-to-SQL task.

\vspace{-2mm}
\paragraph{Evaluation}
We employ three metrics to assess the effectiveness of the parsing and execution stages described in \cref{sec:neuralsql}, as well as the overall performance of the QA system: 
1) \textit{Logical Form Accuracy} ($Acc_{LF}$) evaluates the performance of the parsing stage ($Q\rightarrow Z$). 
It computes the accuracy by performing an exact match comparison between the logical form of the predicted program $\hat{Z}$ and that of the ground truth program $Z$;
2) \textit{Ground-truth Execution Accuracy} ($Acc_{EX|gt}$) assesses the accuracy of the execution stage ($Z\rightarrow A$) by comparing the result of the ground truth program $Z$ with the ground truth answer $A$.
For \textit{Table}-related QA in EHRXQA, this metric yields 100\% accuracy.
For \textit{Image}-related QA and \textit{Image+Table}-related QA, this equates to measuring the VQA performance;
3) \textit{Prediction Execution Accuracy} ($Acc_{EX|pred}$) evaluates the accuracy of execution with the predicted program $\hat{Z}$, providing an assessment of the overall system performance, including both parsing and execution stages.

\vspace{-2mm}
\paragraph{Baselines}
We build a strong QA baseline by combining ChatGPT~\cite{openai2022chatgpt} and $\text{M}^3\text{AE}$~\cite{chen2022multim3ae}, which are outperforming models in the fields of semantic parsing (\eg, Text-to-Query) and medical VQA (\eg, MIMIC-CXR-VQA), respectively.
For ChatGPT, we conduct in-context learning~\cite{brown2020language} (ICL) through two different prompt strategies: 1) Fixed: using fixed \texttt{N}-shot (Question, Query) pairs; 2) BM25 (train)~\cite{robertson2009probabilistic}: retrieving \texttt{N} relevant (Question, Query) pairs from the training QA dataset for a given question. 
These retrieved pairs are then used as few-shot examples.
Here, we use \texttt{N} as 10.
For $\text{M}^3\text{AE}$, we first train it on our MIMIC-CXR-VQA and then deploy it as our external VQA API, integrated within NeuralSQL.
For more detailed implementations, please refer to \cref{supp_exp_xqa}. 

\begin{table}[t]
\caption{Comparison of ChatGPT (\texttt{gpt-3.5-turbo-0613}) with $\text{M}^3\text{AE}$ model on EHRXQA dataset using two different prompting strategies for \textit{Image}-, \textit{Table}-, and \textit{Image+Table}-related QA.
}
\label{tab:ehrxqa_exp}
\centering
\resizebox{\textwidth}{!}{%
\begin{tabular}{ccccccccccc}
\toprule
  \multirow{3}{*}{Model} & 
  \multirow{3}{*}{Prompt} & 
  \multicolumn{3}{c}{\textit{Image}-related} &
  \multicolumn{3}{c}{\textit{Table}-related} &
  \multicolumn{3}{c}{\textit{Image+Table}-related} \\  \cmidrule(lr){3-5} \cmidrule(lr){6-8} \cmidrule(lr){9-11} 
     & &
  $Acc_{LF}$ &
  $Acc_{EX|gt}$ &
  $Acc_{EX|pred}$ &
  $Acc_{LF}$ &
  $Acc_{EX|gt}$ &
  $Acc_{EX|pred}$ &
  $Acc_{LF}$ &
  $Acc_{EX|gt}$ &
  $Acc_{EX|pred}$ \\ \midrule
\multirow{2.5}*{\begin{tabular}[c]{@{}c@{}}
ChatGPT \\ + $\text{M}^3\text{AE}$ \end{tabular}} & 
   Fixed &
    1.1 & 49.4 & 17.4 &
    4.9 & 100.0 & 30.0 &
    4.8 & 68.8 & 35.7
  \\ \cmidrule(lr){2-11} 
 &
 BM25 (train) &
    87.3 & 49.4 & 48.2 &
    73.0 & 100.0 & 92.9 &
    72.5 & 68.8 & 65.9
  \\ \bottomrule
\vspace{-10mm}
\end{tabular}%
}
\end{table}

\vspace{-2mm}
\paragraph{Results and Findings}
\cref{tab:ehrxqa_exp} shows the performance of EHRXQA, with three metrics for each \textit{modality}-based scope. 
The first row of the table shows the performance when using a fixed prompt for all questions, while the second row shows the performance when given a different prompt for each question using BM25. 
As shown in the \cref{tab:ehrxqa_exp}, giving relevant few-shot examples using BM25 significantly boosts performance. 
In the case of \textit{Table}-related questions, our model achieves 92.9\% $Acc_{EX|pred}$ score with 73.0\% $Acc_{LF}$ score.
However, when it comes to the remaining questions that rely on image information, our model demonstrates a relatively low performance, even though it maintains a high $Acc_{LF}$ score.
Specifically, for \textit{Image}-related questions, the $Acc_{LF}$ is 87.3\% as compared to the $Acc_{EX|pred}$ of 48.2\%. 
For \textit{Image+Table}-related questions, the model achieves an $Acc_{LF}$ of 72.5\%, while the $Acc_{EX|pred}$ is 65.9\%.

Notably, the model's performance at the execution stage ($Acc_{EX|pred}$) is affected by the number of images that the model (\textit{i.e.}, VQA model) needs to process. 
For example, in the context of \textit{Image}-related QA, we observed that the $Acc_{EX|pred}$ drops to 39.6\% when the model has to process multiple images (\textit{i.e.}, (Image, single, N-image) scope described in \cref{tab:sample-table}) within a \textit{single} patient QA scope. 
The situation worsens in a \textit{group} QA scope where the model faces the challenge of accurately predicting a large number of image results, leading to an $Acc_{EX|pred}$ of 1.7\%. 
This observed trend contributes to the relatively reduced performance for \textit{Image}-related (48.2\%) and \textit{Image+Table}-related questions (65.9\%), even when considering the model's peak overall test set performance (69.2\%) as detailed in \cref{tab:baseline}. 

This trend also explains the model showing superior performance ($Acc_{EX|pred}$) on \textit{Image+Table}-related questions ($65.9\%$) than on \textit{Image}-related questions ($48.2\%$).
Given the complex conditions present in \textit{Image+Table}-related questions, the scope of images becomes more specified. 
This leads to a lower number of images to process in comparison to \textit{Image}-related scenarios, resulting in a relatively higher performance for these multi-modal queries.
Overall, the huge gap between $Acc_{LF}$ and $Acc_{EX|pred}$ suggests visual perception could be a bigger roadblock to AI models being deployed in clinical practice than logical reasoning, and future research should put as much emphasis on perception as complex logical reasoning.

\vspace{-3mm}
\section{Discussion}
\label{sec:discussion}
\vspace{-3mm}

\textbf{Limitations}\;\;
Though we have carefully designed the dataset, several limitations exist: 1) Since our dataset is based on the MIMIC database, it potentially limits its generalizability. 2) Due to the constrained label scope of Chest Imagenome, our dataset lacks the capability to address more detailed visual questions, such as identifying specific tumor sizes from chest X-rays. 3) Unlike EHRSQL, our model does not include unanswerable questions, an aspect that, if addressed, could enhance our model's comprehensiveness and applicability. Future work should aim to address these constraints. \\[0pt]
\textbf{Future Direction}\;\;
Our study signifies a substantial step forward in multi-modal EHR QA systems, but notable potential for refinement remains. Key future directions include: 1) Enlarging the scope of our dataset by enhancing the multi-modal dialogue system~\cite{li2022mmcoqa}; 2) Incorporating mechanisms to address unanswerable questions or ambiguous images, which is crucial for real-world applications~\cite{lee2022ehrsql}; and 3) Broadening our modality by evolving our dataset to support tri-modal question answering~\cite{hannan2020manymodalqa, talmor2021multimodalqa}. These forward-looking endeavors will leverage our dataset as a valuable resource, laying the groundwork for more comprehensive and practical healthcare solutions.

\newpage
\begin{ack}
We are grateful to Jiho Kim, Jiyoung Lee, Youngjune Lee, JongHak Moon, Hyunseung Chung, and Seungho Kim for their fruitful comments and inspiration.
We would like to thank three anonymous reviewers for their time and insightful comments.
This work was (partially) supported by Microsoft Research Asia, Institute of Information \& Communications Technology Planning \& Evaluation (IITP) grant (No.2019-0-00075, RS-2022-00155958), National Research Foundation of Korea (NRF) grant (NRF-2020H1D3A2A03100945), and the Korea Health Industry Development Institute (KHIDI) grant (No.HR21C0198), funded by the Korea government (MSIT, MOHW).
\end{ack}


\medskip

{
\small
\bibliographystyle{plain}
\bibliography{neurips_data_2023}
}

\newpage
\appendix

\newpage
\newcommand*{\rom}[1]{\expandafter\romannumeral #1}
\setcounter{figure}{0} 
\setcounter{table}{0} 
\renewcommand{\thetable}{\Alph{section}\arabic{table}}
\renewcommand{\thefigure}{\Alph{section}\arabic{figure}}

\newpage
\startcontents[supplementary]
\printcontents[supplementary]{l}{1}{\section*{Supplementary Contents}}

\newpage
\section{Datasheet for Datasets}
\paragraph{A.1\;\;\;\;Motivation}
\begin{itemize}
    \item \textbf{For what purpose was the dataset created?}
    
    We created EHRXQA to provide a valuable resource for advancing machine learning applications in multi-modal question answering systems on structured electronic health records (EHRs) and chest X-ray images. 
    As an affiliated dataset, we created MIMIC-CXR-VQA to provide a benchmark for medical visual question answering systems.
    
    \item \textbf{Who created the dataset (e.g., which team, research group) and on behalf of which entity (e.g., company, institution, organization)?}
    
    The authors of this paper.

    \item \textbf{Who funded the creation of the dataset? If there is an associated grant, please provide the name of the grantor and the grant name and number.}

    This work was (partially) supported by Microsoft Research Asia, Institute of Information \& Communications Technology Planning \& Evaluation (IITP) grant (No.2019-0-00075, RS-2022-00155958), National Research Foundation of Korea (NRF) grant (NRF-2020H1D3A2A03100945), and the Korea Health Industry Development Institute (KHIDI) grant (No.HR21C0198), funded by the Korea government (MSIT, MOHW).
\end{itemize}

\paragraph{A.2\;\;\;\;Composition}
\begin{itemize}
    \item \textbf{What do the instances that comprise the dataset represent (e.g., documents, photos, people, countries)?}
    
    EHRXQA contains natural questions and corresponding SQL/NeuralSQL queries (text). 
    MIMIC-CXR-VQA contains the image ID of the MIMIC-CXR dataset and their related natural questions.

    \item \textbf{How many instances are there in total (of each type, if appropriate)?}

    In EHRXQA, there are about 46.2K instances (16,366 image-related samples, 16,529 table-related samples, and 13,257 image+table-related samples).
    In MIMIC-CXR-VQA, there are about 377.4K instances.
    
    \item \textbf{Does the dataset contain all possible instances or is it a sample (not necessarily random) of instances from a larger set?}

    We will provide all instances in our GitHub repository for EHRXQA\footnote{\href{https://github.com/baeseongsu/ehrxqa}{\texttt{https://github.com/baeseongsu/ehrxqa}}} and MIMIC-CXR-VQA\footnote{\href{https://github.com/baeseongsu/mimic-cxr-vqa}{\texttt{https://github.com/baeseongsu/mimic-cxr-vqa}}}.

    \item \textbf{What data does each instance consist of?}

    EHRXQA contains (Question, SQL/NeuralSQL, Answer) pair for each instance.
    MIMIC-CXR-VQA contains (Question, CXR image ID, Answer) pair for each instance.

    \item \textbf{Is there a label or target associated with each instance?}

    The answer (label) is provided for each question.

    \item \textbf{Is any information missing from individual instances? If so, please provide a description, explaining why this information is missing (e.g., because it was unavailable). This does not include intentionally removed information, but might include, e.g., redacted text.}
    
    No.

    \item \textbf{Are relationships between individual instances made explicit (e.g., users' movie ratings, social network links)?}

    No.

    \item \textbf{Are there recommended data splits (e.g., training, development/validation, testing)?}

    See \cref{supp_prelim_imgdataset}, and \cref{supp_ehrxqa_split}.

    \item \textbf{Are there any errors, sources of noise, or redundancies in the dataset?}

    Questions are created by filling the slots in the templates with pre-defined values and records from the database. Thus, some questions can be grammatically incorrect but not critical (\eg, verb tense).

    \item \textbf{Is the dataset self-contained, or does it link to or otherwise rely on external resources (e.g., websites, tweets, other datasets)?}

    EHRXQA depends on three open-source databases: MIMIC-IV\footnote{\url{https://physionet.org/content/mimiciv/2.2/}}, MIMIC-CXR\footnote{\url{https://physionet.org/content/mimic-cxr/2.0.0/}}, and Chest ImaGenome\footnote{\url{https://physionet.org/content/chest-imagenome/1.0.0/}}, which are accessible via PhysioNet\footnote{\url{https://physionet.org/}}. 
    MIMIC-CXR-VQA depends on two open-source databases: MIMIC-CXR and Chest ImaGenome.

    \item \textbf{Does the dataset contain data that might be considered confidential (e.g., data that is protected by legal privilege or by doctor-patient confidentiality, data that includes the content of individuals' non-public communications)?}

    No.

    \item \textbf{Does the dataset contain data that, if viewed directly, might be offensive, insulting, threatening, or might otherwise cause anxiety?}

    No.

    \item \textbf{Does the dataset relate to people?}

    Yes.

    \item \textbf{Does the dataset identify any subpopulations (e.g., by age, gender)?}
    
    No.

    \item \textbf{Does the dataset contain data that might be considered sensitive in any way (e.g., data that reveals race or ethnic origins, sexual orientations, religious beliefs, political opinions or union memberships, or locations; financial or health data; biometric or genetic data; forms of government identification, such as social security numbers; criminal history)?}

    No. The source datasets are already de-identified.
    
\end{itemize}

\paragraph{A.3\;\;\;\;Collection process}
\begin{itemize}
    \item \textbf{How was the data associated with each instance acquired?}
    To collect diverse questions, we constructed question templates and their associated query (SQL, NerualSQL) by analyzing existing resources (for image, medical VQA datasets, and for table, table-based EHR QA datasets). Then, we sampled QA samples from source databases (MIMIC-IV, MIMIC-CXR, Chest ImaGenome) for each question template.

    \item \textbf{What mechanisms or procedures were used to collect the data (e.g., hardware apparatuses or sensors, manual human curation, software programs, software APIs)?}
    
    We mainly used Excel, Google Sheets, and Python to collect, process and label the data.
    In addition, we used OpenAI's ChatGPT (GPT-4) to generate paraphrases for each question template.

    \item \textbf{If the dataset is a sample from a larger set, what was the sampling strategy (e.g., deterministic, probabilistic with specific sampling probabilities)?}

    When it needs random sampling such as data splitting, patient sampling, CXR image sampling for each (question, answer) pair, we fixed the random seed and randomly selected a fixed number of samples from the larger set.
    
    \item \textbf{Who was involved in the data collection process (e.g., students, crowd workers, contractors) and how were they compensated (e.g., how much were crowd workers paid)?}

    The data collection and construction process, which included SQL/NeuralSQL labeling, was performed exclusively by the authors of the study. No crowd workers were involved due to the sensitive nature of the data and the specialized knowledge required for query labeling. 

    \item \textbf{Over what timeframe was the data collected?}

    The EHRXQA and MIMIC-CXR-VQA datasets were constructed in 2023. They were built using data from the MIMIC-CXR and MIMIC-IV databases. The MIMIC-IV data was collected between 2008 and 2019, and the MIMIC-CXR data was collected between 2011 and 2016.

    \item \textbf{Were any ethical review processes conducted (e.g., by an institutional review board)?}

    N/A.

    \item \textbf{Does the dataset relate to people?}

    Yes.

    \item \textbf{Did you collect the data from the individuals in question directly, or obtain it via third parties or other sources (e.g., websites)?}

    N/A.

    \item \textbf{Were the individuals in question notified about the data collection?}

    N/A.

    \item \textbf{Did the individuals in question consent to the collection and use of their data?}

    N/A.

    \item \textbf{If consent was obtained, were the consenting individuals provided with a mechanism to revoke their consent in the future or for certain uses?}

    N/A.

    \item \textbf{Has an analysis of the potential impact of the dataset and its use on data subjects (e.g., a data protection impact analysis) been conducted?}

    The dataset does not have individual-specific information.
\end{itemize}

\paragraph{A.4\;\;\;\;Preprocessing/cleaning/labeling}
\begin{itemize}
    \item \textbf{Was any preprocessing/cleaning/labeling of the data done (e.g., discretization or bucketing, tokenization, part-of-speech tagging, SIFT feature extraction, removal of instances, processing of missing values)?}

    N/A.

    \item \textbf{Was the ``raw'' data saved in addition to the preprocess/cleaned/labeled data (e.g., to support unanticipated future uses)?}

    N/A.

    \item \textbf{Is the software that was used to preprocess/clean/label the data available?}

    Preprocessing, cleaning, and labeling are done via Excel, Google Sheets, and Python.   
\end{itemize}

\paragraph{A.5\;\;\;\;Uses}
\begin{itemize}
    \item \textbf{Has the dataset been used for any tasks already?}

    No.

    \item \textbf{Is there a repository that links to any or all papers or systems that use the dataset?}

    No.

    \item \textbf{What (other) tasks could the dataset be used for?}

    Our dataset is designed to promote research in question answering systems related to structured electronic health records (EHRs), chest X-ray (CXR) images, or a combination of both.

    \item \textbf{Is there anything about the composition of the dataset or the way it was collected and preprocessed/cleaned/labeled that might impact future uses?}

    N/A.

    \item \textbf{Are there tasks for which the dataset should not be used?}

    N/A.
\end{itemize}

\paragraph{A.6\;\;\;\;Distribution}
\begin{itemize}
    \item \textbf{Will the dataset be distributed to third parties outside of the entity (e.g., company, institution, organization) on behalf of which the dataset was created?}

    No.

    \item \textbf{How will the dataset be distributed?}

    The datasets will be released at \href{https://github.com/baeseongsu/ehrxqa}{\texttt{https://github.com/baeseongsu/ehrxqa}} and \href{https://github.com/baeseongsu/mimic-cxr-vqa}{\texttt{https://github.com/baeseongsu/mimic-cxr-vqa}} upon publication.

    \item \textbf{Will the dataset be distributed under a copyright or other intellectual property (IP) license, and/or under applicable terms of use (ToU)?}

    The dataset is released under MIT License.

    \item \textbf{Have any third parties imposed IP-based or other restrictions on the data associated with the instances?}

    No.

    \item \textbf{Do any export controls or other regulatory restrictions apply to the dataset or to individual instances?}

    No.
\end{itemize}

\paragraph{A.7\;\;\;\;Maintenance}
\begin{itemize}
    \item \textbf{Who will be supporting/hosting/maintaining the dataset?}

    The authors of this paper.

    \item \textbf{How can the owner/curator/manager of the dataset be contacted(e.g., email address)?}
    
    Contact the first authors (\texttt{seongsu@kaist.ac.kr} \& \texttt{kyungdaeun@kaist.ac.kr}).

    \item \textbf{Is there an erratum?}

    No.

    \item \textbf{Will the dataset be updated (e.g., to correct labeling erros, add new instances, delete instances)?}

    If any corrections are required, our plan is to upload an updated version of the dataset with comprehensive explanations for the changes. Furthermore, as we broaden our QA scope, we will consistently update the dataset with new QA templates/instances.

    \item \textbf{If the dataset relates to people, are there applicable limits on the retention of the data associated with the instances (e.g., were the individuals in question told that their data would be retained for a fixed period of time and then deleted)?}

    N/A

    \item \textbf{Will older versions of the dataset continue to be supported/hosted/maintained?}

    Primarily, we plan to maintain only the most recent version of the dataset. 
    However, under certain circumstances, such as significant updates to our dataset or the need for validation of previous research work using older versions, we will exceptionally preserve previous versions of the dataset for up to one year.

    \item \textbf{If others want to extend/augment/build on/contribute to the dataset, is there a mechanism for them to do so?}

    Contact the authors in this paper.
\end{itemize}

\newpage
\section{Preliminary} 
\subsection{Uni-modal data resources}\label{supp_prelim_datasource}
For our research, we utilize the dataset under the PhysioNet license, ensuring compliance with the required credentials and permissions. The following are the data resources we utilize:
\begin{itemize}[itemsep=1pt, topsep=-2pt, leftmargin=7mm]
    \item MIMIC-IV: Available at \href{https://physionet.org/content/mimiciv/2.2/}{https://physionet.org/content/mimiciv/2.2/}
    \item MIMIC-CXR: Available at \href{https://physionet.org/content/mimic-cxr-jpg/2.0.0/}{https://physionet.org/content/mimic-cxr-jpg/2.0.0/}
    \item Chest ImaGenome: Available at \href{https://physionet.org/content/chest-imagenome/1.0.0/}{https://physionet.org/content/chest-imagenome/1.0.0/}
\end{itemize}

\subsection{Uni-modal EHR QA datasets}\label{supp_prelim_dataset}
\subsubsection{Table-based EHR QA}\label{supp_prelim_tabdataset}
\paragraph{\CheckedBox\;\; Template construction for MIMIC-IV}
Given the structural similarities between the MIMIC-IV and MIMIC-III databases, we successfully adapted the original MIMIC-III question templates from the EHRSQL dataset for use with MIMIC-IV.
Our methodology involved a comprehensive analysis of the question templates associated with both the MIMIC-III and MIMIC-IV database schemas to identify similarities and discrepancies.
Through this comparative study, we were able to pinpoint the discrepancies between the question templates designed for MIMIC-III and those compatible with MIMIC-IV.
While a substantial portion (\ie, 165 templates) of the question templates from MIMIC-III could be seamlessly adapted to MIMIC-IV, we identified several question templates that were unique to each database, as presented in \cref{tasupp_tabs:mimic_template_comparison}.
For instance, MIMIC-IV provides information about microbiology test names, a feature absent in MIMIC-III.
Taking these differences into account, we have assembled a collection of 174 question templates specifically designed for MIMIC-IV.

\begin{longtblr}[
  caption={We present a comparative analysis of template suitability between MIMIC-III and MIMIC-IV. This includes question templates from EHRSQL, indicating their applicability to MIMIC-III and/or MIMIC-IV. Checkmarks ({\color{teal}\cmark}) and crosses ({\color{red}\xmark}) are used to denote compatibility and incompatibility, respectively.},
  label={tasupp_tabs:mimic_template_comparison}
]{%
  colspec={Q[c] X[12,l,m] Q[1.5,c,m] Q[1.5,c,m]}, 
  colsep=3.0pt,
  rowhead=1, 
  hlines,
  rows={font=\scriptsize},
  rowsep=0.5pt,
  rowfoot=1 
}
\textbf{No.} & \textbf{Question Template} & \textbf{MIMIC-III} & \textbf{MIMIC-IV} \\
21 & How many [unit\_count] have passed since the [time\_filter\_exact1] time patient \{patient\_id\} was transferred to ward {ward\_id} on the current hospital visit? 
& {\color{teal}\cmark} & {\color{red}\xmark} \\
22 & How many [unit\_count] have passed since the [time\_filter\_exact1] time patient \{patient\_id\} received a procedure on the current hospital visit? 
& {\color{red}\xmark} & {\color{red}\xmark} \\
23 & How many [unit\_count] have passed since the [time\_filter\_exact1] time patient \{patient\_id\} received a {procedure\_name} procedure on the current hospital visit? 
& {\color{red}\xmark} & {\color{red}\xmark} \\
29 & What was the [time\_filter\_exact1] ward of patient \{patient\_id\} [time\_filter\_global1]? & {\color{teal}\cmark} & {\color{red}\xmark} \\
46 & What was the name of the allergy that patient \{patient\_id\} had [time\_filter\_global1]?
& {\color{red}\xmark} & {\color{red}\xmark} \\
47 & What was the name of the substance that patient \{patient\_id\} was allergic to [time\_filter\_global1]? 
& {\color{red}\xmark} & {\color{red}\xmark} \\
49 & What was the organism name found in the [time\_filter\_exact1] \{test\_name\} test of patient \{patient\_id\} [time\_filter\_global1]? 
& {\color{red}\xmark} & {\color{teal}\cmark} \\
51 & What was the name of the microbiology test that patient \{patient\_id\} [time\_filter\_exact1] received [time\_filter\_global1]? 
& {\color{red}\xmark} & {\color{teal}\cmark} \\
78 & When was patient {patient\_id}'s [time\_filter\_exact1] \{test\_name\} test [time\_filter\_global1]? 
& {\color{red}\xmark} & {\color{teal}\cmark} \\
85 & Has\_verb patient {patient\_id} received a \{procedure\_name\} procedure in other than the current hospital [time\_filter\_global1]? 
& {\color{red}\xmark} & {\color{red}\xmark} \\
98 & Has\_verb patient \{patient\_id\} had any allergy [time\_filter\_global1]? 
& {\color{red}\xmark} & {\color{red}\xmark} \\
101 & Has\_verb patient \{patient\_id\} had any \{test\_name\} test result [time\_filter\_global1]?
& {\color{red}\xmark} & {\color{teal}\cmark} \\
103 & Has\_verb there been any organism found in the [time\_filter\_exact1] \{test\_name\} test of patient \{patient\_id\} [time\_filter\_global1]? 
& {\color{red}\xmark} & {\color{teal}\cmark} \\
137 & Count the number of patients who stayed in ward \{ward\_id\} [time\_filter\_global1]. 
& {\color{teal}\cmark} & {\color{red}\xmark} \\
153 & Count the number of patients who received a \{test\_name\} test [time\_filter\_global1]. 
& {\color{red}\xmark} & {\color{teal}\cmark} \\
176 & What are\_verb the top [n\_rank] frequent microbiology tests [time\_filter\_global1]? 
& {\color{red}\xmark} & {\color{teal}\cmark} \\
178 & What are\_verb the top [n\_rank] frequent microbiology tests that patients had [time\_filter\_within] after having been diagnosed with \{diagnosis\_name\} [time\_filter\_global1]? 
& {\color{red}\xmark} & {\color{teal}\cmark} \\
180 & What are\_verb the top [n\_rank] frequent microbiology tests that patients had [time\_filter\_within] after having received a \{procedure\_name\} procedure [time\_filter\_global1]? 
& {\color{red}\xmark} & {\color{teal}\cmark} \\
\end{longtblr}

\paragraph{\CheckedBox\;\; SQL annotations}
Following a similar approach to the template construction, we utilize the EHRSQL SQL templates and adapt them for MIMIC-IV. 
Due to the structural similarities between MIMIC-III and MIMIC-IV, we can seamlessly map the schema from MIMIC-III to MIMIC-IV without altering the SQL query logic significantly. 
This enables the efficient conversion of SQL queries from MIMIC-III to MIMIC-IV, thus reducing the time and effort required to adapt the queries for the new database.
While most of the schema information (\ie, table and column names) remains the same, there are minor modifications introduced in MIMIC-IV compared to MIMIC-III, as presented in ~\cref{supp_tabs:mimic_schema_mapping}.
Therefore, we annotate the corresponding SQL queries for 174 question templates.

\begin{table}[H]
\centering
\caption{Column-wise schema mapping from MIMIC-III to MIMIC-IV}
\label{supp_tabs:mimic_schema_mapping}
\renewcommand{\arraystretch}{1.2} 
\resizebox{0.6\textwidth}{!}{%
    \begin{tabular}{cc}
    \hline
    \textbf{MIMIC-III schema} & \textbf{MIMIC-IV schema} \\
    \hline
    d\_icd\_diagnoses.icd9\_code & d\_icd\_diagnoses.icd\_code \\
    d\_icd\_diagnoses.short\_title & d\_icd\_diagnoses.long\_title \\
    d\_icd\_procedures.icd9\_code & d\_icd\_procedures.icd\_code \\
    d\_icd\_procedures.short\_title & d\_icd\_procedures.long\_title \\
    diagnoses\_icd.icd9\_code & diagnoses\_icd.icd\_code \\
    procedures\_icd.icd9\_code & procedures\_icd.icd\_code \\
    prescriptions.startdate & prescriptions.starttime \\
    prescriptions.enddate & prescriptions.stoptime \\
    chartevents.icustay\_id & chartevents.stay\_id \\
    inputevents\_cv.icustay\_id & inputevents.stay\_id \\
    inputevents\_cv.charttime & inputevents.starttime \\
    outputevents.icustay\_id & outputevents.stay\_id \\
    icustays.icustay\_id & icustays.stay\_id \\
    transfers.icustay\_id & transfers.transfer\_id \\
    \hline
    \end{tabular}
}
\end{table}

\paragraph{\CheckedBox\;\; Other implementation details}
We use the fundamental template grammar from EHRSQL, including key components like the operation value slot, condition value slot, and time filter slot. 
Each slot has an associated natural language expression and a corresponding SQL pattern.
For a complete list of time templates, operation values, and condition values, please refer to the comprehensive listing detailed in the original EHRSQL paper.

\newpage
\subsubsection{Image-based EHR QA: MIMIC-CXR-VQA}\label{supp_prelim_imgdataset}
\paragraph{\CheckedBox\;\;Preprocessing}
Our dataset is preprocessed following these steps:
\begin{enumerate}[itemsep=1pt, topsep=-2pt, leftmargin=7mm]
    \item \textbf{Image selection}
    \begin{enumerate}
        \item 
        We filter out images captured from frontal view positions (\ie, PA, AP).
        \item 
        Per study, we select one representative CXR image based on the earliest study datetime.
        \item
        When multiple images in a study share the same study datetime, we choose the image whose dicom ID is alphabetically first.
    \end{enumerate}
    \item \textbf{Outlier removal}
    \begin{enumerate}
        \item
        We cap the maximum number of consecutive studies per patient at 20.
        \item
        We eliminate images missing bounding boxes for any anatomical locations.
        \item
        We discard images with widths exceeding three standard deviations from the mean for each anatomical location.
        \item
        To maintain uniformity between the gold and silver datasets' object and attribute pools, we eliminate six attributes (\ie, ``\textit{aortic graft/repair}'', ``\textit{artifact}'', ``\textit{bronchiectasis}'', ``\textit{diaphragmatic eventration (benign)}'', ``\textit{pigtail catheter}'', ``\textit{skin fold}'') and one object (\ie, ``\textit{left arm}'') that are exclusively present in the silver dataset. 
        \item
        We also exclude the object ``\textit{right arm}'' due to its association with the object ``\textit{left arm}'', which is not present in the gold dataset. 
    \end{enumerate}
    \item \textbf{Label refinement}\;\;
        \begin{enumerate}
            \item
            To enhance the precision of label assignments, we employ three types of CXR ontology used in Chest ImaGenome: (\rom{1}) parent-to-child (object) relationships; (\rom{2}) parent-to-child (attribute) relationships; (\rom{3}) possible relationships (object-attribute).
            \item
            For parent-to-child (object) relationships, we propagate the presence of attributes from a child object to its parent object.
            For instance, if the left middle lung zone (a child object) is labeled with pneumonia (an attribute), then the left lung (its parent object) must also be labeled with pneumonia.
            \item
            For parent-to-child (attribute) relationships, we propagate the presence of a child attribute to its parent attribute.
            For example, if lung cancer (a child attribute) is present in the left lung (an object), then lung opacity (a parent attribute) must also be present in the same object.
            \item
            For possible relationships (object-attribute), we exclude any relationships between an object and its associated attributes that are not allowed by the ontology.
            For instance, according to this ontology, the object `left lung' cannot be associated with the attribute `clavicle fracture'.
        \end{enumerate}
    \item \textbf{Dataset split}
        \begin{enumerate}
            \item 
            From the \textit{silver} dataset in Chest ImaGenome, which is machine-generated, we divide it into a 95:5 ratio, with 95\% serving as the training image pool and 5\% as the validation image pool. 
            These image pools consist of 164,398 training images and 8,653 validation images.
            \item 
            The \textit{gold} dataset, which is human-labeled, is used as the test image pool, consisting of 500 test images.
            \item
            During the splitting process, we also balance the distribution between abnormal images (\ie, studies with at least one attribute present) and normal images (\ie, studies without any attributes).
            \item
            Note that each image in image pools represents 563 relationships between 36 anatomical locations (\ie, objects) and their associated attributes (a total of 68), indicating the presence or absence of an attribute for an object.
        \end{enumerate}
\end{enumerate}

\newpage
\paragraph{\CheckedBox\;\;Question template construction - \textit{argument}}
In our template, we designate five primary arguments, represented as \$\{...\}: \$\{object\}, \$\{attribute\}, \$\{category\}, \$\{viewpos\}, and \$\{gender\}. 
When an argument is required to appear more than once in a question template, each time with a unique value, we append an index to it, like \$\{object\_1\} or \$\{object\_2\}. 
Each of these arguments can be replaced by a specific value, as will be displayed in the following ~\cref{tab:my_label}.

\begin{table}[h]
\centering
\caption{Mapping of Arguments to their Potential Values}
    \begin{tabularx}{1.0\textwidth}{cX}
    \toprule
    \textbf{Argument} & \textbf{Values} \\
    \midrule
    \$\{object\} & abdomen, aortic arch, cardiac silhouette, carina, cavoatrial junction, left apical zone, left breast, left chest wall, left clavicle, left costophrenic angle, left hemidiaphragm, left hilar structures, left lower lung zone, left lung, left mid lung zone, left shoulder, left upper lung zone, mediastinum, neck, right apical zone, right atrium, right breast, right chest wall, right clavicle, right costophrenic angle, right hemidiaphragm, right hilar structures, right lower lung zone, right lung, right mid lung zone, right shoulder, right upper lung zone, spine, svc, trachea, upper mediastinum \\
    \midrule
    \$\{attribute\} & airspace opacity, alveolar hemorrhage, aspiration, atelectasis, bone lesion, breast/nipple shadows, cabg grafts, calcified nodule, cardiac pacer and wires, chest port, chest tube, clavicle fracture, consolidation, copd/emphysema, costophrenic angle blunting, cyst/bullae, elevated hemidiaphragm, endotracheal tube, enlarged cardiac silhouette, enlarged hilum, enteric tube, fluid overload/heart failure, goiter, granulomatous disease, hernia, hydropneumothorax, hyperaeration, ij line, increased reticular markings/ild pattern, infiltration, interstitial lung disease, intra-aortic balloon pump, linear/patchy atelectasis, lobar/segmental collapse, low lung volumes, lung cancer, lung lesion, lung opacity, mass/nodule (not otherwise specified), mediastinal displacement, mediastinal drain, mediastinal widening, multiple masses/nodules, pericardial effusion, picc, pleural effusion, pleural/parenchymal scarring, pneumomediastinum, pneumonia, pneumothorax, prosthetic valve, pulmonary edema/hazy opacity, rib fracture, rotated, scoliosis, shoulder osteoarthritis, spinal degenerative changes, spinal fracture, sub-diaphragmatic air, subclavian line, subcutaneous air, superior mediastinal mass/enlargement, swan-ganz catheter, tortuous aorta, tracheostomy tube, vascular calcification, vascular congestion, vascular redistribution \\
    \midrule
    \$\{category\} & anatomicalfinding, device, disease, technicalassessment, tubesandlines \\
    \midrule
    \$\{viewpos\} & AP, PA \\
    \midrule
    \$\{gender\} & male, female \\
    \bottomrule
    \end{tabularx}
\label{tab:my_label}
\end{table}
\paragraph{\CheckedBox\;\;Question template construction - \textit{template component}}
We define each question template in our structure with three major components:
\begin{itemize}[itemsep=1pt, topsep=-2pt, leftmargin=7mm]
\item
    \textbf{Filter condition}: This defines the question's domain or subject area. For instance, in the template ``\textit{Is there \$\{attribute\} in the \$\{object\}?}'', ``\textit{\$\{object\}}'' serves as the filter condition, focusing the question on a particular anatomical location. Filter conditions can include multiple arguments to create more complex queries using unions, intersections, or differences.
    \item
    \textbf{Target pattern}: This denotes the particular detail within the filter condition's scope that the question seeks to explore. In the example template, ``\textit{Is there \$\{attribute\} in the \$\{object\}?}'', ``\textit{\$\{attribute\}}'' forms the target pattern. When this is combined with the semantic type, it yields a question with a completed intent.
    \item
    \textbf{Semantic type}: This labels the question based on the nature of the expected response. There are three primary semantic types: `\textit{verify}' for yes/no questions, `\textit{query}' for answers in the form of a list or set, and `\textit{choose}' for questions that involve selection from provided options.
\end{itemize}

\newpage
\paragraph{\CheckedBox\;\;Question template construction - \textit{content type}}
Following the construction of our templates, we classify our 48 question templates into seven distinct \textbf{content types} (See \cref{tab:vqacategory}): \textit{anatomy}, \textit{attribute}, \textit{presence}, \textit{abnormality}, \textit{plane}, \textit{gender}, and \textit{size}. 
Although these categories are often referred to as ``question type'' in other medical VQA datasets, we choose to refer to them as ``content type'' to provide a more precise characterization. 
Each content type is described in detail as follows:
\begin{itemize}[itemsep=1pt, topsep=-2pt, leftmargin=7mm]
    \item \textit{anatomy}: 
    We include all question templates related to asking about anatomical locations in the target pattern, but exclude verification questions from this content type.
    \item \textit{attribute}: 
    We include all question templates related to asking about attributes or categories in the target pattern, but exclude verification questions from this content type.
    \item \textit{presence}: 
    We include all verify questions that ask about the presence of attributes or categories given the entire image or specific anatomical locations.
    \item \textit{abnormality}:
    We include all question templates that are related to abnormality (defining the concept of ``\textit{abnormality}'' as a superset of four categories) in their questions.
    \item \textit{plane}: 
    We include the determination of the radiography's view position, following the VQA-RAD's QA scope.
    \item \textit{gender}: 
    We include the identification of gender from the images, following the VQA-RAD's QA scope.
    \item \textit{size}:
    We include two clinically significant measurements: cardiothoracic ratio (\ie, CTR) and mediastinal-thoracic ratio (\ie, MTR).
    The CTR measures the maximal horizontal cardiac diameter against the maximal horizontal thoracic diameter (inner edge of ribs/edge of pleura). 
    Conversely, the MTR calculates the ratio of the maximum mediastinal width to the maximum thoracic width. 
    We derive these ratios using three measurements: the cardiac silhouette's width, the upper mediastinum's width, and the thorax width. 
    The thorax width is defined by the largest x-axis value of the left lung and the smallest x-axis value of the right lung, considering the original reverse orientation of an X-ray. 
    We have established normal measurement thresholds, aligning with the conventional parameters in radiology (CTR: 1/2, MTR: 1/3).
\end{itemize}

\begin{table}[H]
\centering
\caption{Content type of VQA question templates on a chest X-ray image. 
}
\resizebox{0.9\textwidth}{!}{%
    \begin{tabular}{ll}
    \toprule
     \multicolumn{1}{c}{Content type} & \multicolumn{1}{c}{Sample question} \\ 
     \midrule
     anatomy & What are all anatomical locations where both infiltration and interstitial lung diseases can be found?  \\
     attribute & List all detected anatomical findings. \\
     presence & Does the cardiac silhouette show any evidence of diseases or devices? \\
     abnormality & Are there signs of abnormalities in both the left lung and the right lung? \\ 
     plane & Is this X-ray image in the AP or PA view? \\
     gender & Please specify the patient's gender. \\
     size & Is the cardiac silhouette's width larger than half of the total thorax width? \\ 
     \bottomrule
    \end{tabular}%
}
\label{tab:vqacategory}
\end{table}

\paragraph{\CheckedBox\;\;VQA dataset generation - \textit{dataset balancing}} 
To build an unbiased VQA dataset, we designed the balancing rules based on the following considerations:
\begin{itemize}[itemsep=1pt, topsep=-2pt, leftmargin=7mm]
    \item Balancing Answers: To avoid language biases within the VQA dataset, we maximized the answer entropy during the sampling. This approach ensures diverse and well-distributed answers to each question, promoting comprehensive image understanding.
    
    \item Balancing Questions per Image: We considered the number of questions per image for sampling a variety of images. We limited each question template to one use per CXR image. Therefore, an image can have a minimum of 0 questions and a maximum of 48 (\ie, the total number of our templates). We globally defined an image counter to increase the probability of less frequently sampled images being selected, thereby promoting greater diversity in our image set.
    
    \item Balancing Sampled Questions per Template: Lastly, we ensured a balanced number of sampled questions per template to maintain uniformity. It ensures that no particular template is over or under-sampled, leading to a fair and diverse question dataset.
\end{itemize}

\newpage
\paragraph{\CheckedBox\;\;Dataset collection - \textit{paraphrasing}}
To generate paraphrases, we leveraged the OpenAI UI, applying \cref{prompt:vqa_data_prompt_para} to create 30 paraphrases per template using the GPT-4 model (version May 24, 2023) as a base. 
Following this, human reviewers pruned any paraphrases that strayed from the initial template's meaning. 
For enhanced diversity in the training, validation, and test sets, we performed k-means clustering ($k=4$) on the paraphrases of each template, based on their edit distance. This process grouped alike paraphrases, which we then distributed in a 3-to-1 ratio for the train and validation/test sets respectively. 
To conclude, we randomly implemented these paraphrases into the datasets, thereby ensuring a broad spectrum of linguistic variations.
On an average, we had 16.5 paraphrases representing each template.
\begin{figure}[H]
\centering
\scalebox{0.80}{
\input{supp_prompts/vqa_data_prompt_para}
}
\caption{Prompt Template for Paraphrasing Question Templates for MIMIC-CXR-VQA. Elements enclosed within double braces \{\{\}\} are substituted with values specific to each template.}
\label{prompt:vqa_data_prompt_para}
\end{figure}

\newpage
\paragraph{\CheckedBox\;\;Dataset statistics of MIMIC-CXR-VQA}
\cref{tab:dataset_stats_overall}, \cref{tab:dataset_stats_content}, and \cref{tab:dataset_stats_semantic} present comprehensive statistics of the MIMIC-CXR-VQA dataset, detailing its overall, content type, and semantic type distributions respectively.
\begin{table}[H]
    \centering
    \caption{Overall statistics of MIMIC-CXR-VQA.}
    \label{tab:dataset_stats_overall}
    \scalebox{0.7}{
        \begin{tabular}{lrrr}
        \toprule
        \textbf{} & \textbf{Training} & \textbf{Validation} & \textbf{Test} \\
        \midrule
        Images    & 133,687 & 8,610  & 500    \\
        Questions & 132,387  & 31,148 & 7,565  \\
        Answers   & 6,628   & 2,508  & 700    \\
        Samples   & 290,031 & 73,567 & 13,793 \\
        \bottomrule
        \end{tabular}
    }
\end{table}

\begin{table}[H]
    \centering
    \caption{Statistics of MIMIC-CXR-VQA by content type.}
    \label{tab:dataset_stats_content}
    \scalebox{0.6}{
        \begin{tabular}{lrrr}
        \toprule
        \textbf{Content Type} & \textbf{Training} & \textbf{Validation} & \textbf{Test} \\
        \midrule
            presence    & 109,455   (37.7\%)    & 26,153 (35.5\%)       & 4,566 (33.1\%) \\
            anatomy     & 37,952    (13.1\%)    & 10,210 (13.9\%)       & 1,963 (14.2\%) \\
            attribute   & 49,948    (17.2\%)    & 13,111 (17.8\%)       & 2,578 (18.7\%) \\
            abnormality & 60,692    (20.9\%)    & 16,109 (21.9\%)       & 3,199 (23.2\%) \\
            size        & 16,000 \enskip(5.5\%) & 4,000 \enskip(5.4\%)  & 705 \enskip(5.1\%) \\
            plane       & 7,992 \enskip(2.8\%)  & 1,992 \enskip(2.7\%)  & 386 \enskip(2.8\%) \\
            gender      & 7,992 \enskip(2.8\%)  & 1,992 \enskip(2.7\%)  & 396 \enskip(2.9\%) \\
        \bottomrule
        \end{tabular}
    }
    \vspace{-5pt}
\end{table}

\begin{table}[H]
    \centering
    \caption{Statistics of MIMIC-CXR-VQA by semantic type.}
    \label{tab:dataset_stats_semantic}
    \scalebox{0.7}{
        \begin{tabular}{lrrr}
        \toprule
        \textbf{Semantic Type} & \textbf{Training} & \textbf{Validation} & \textbf{Test} \\
        \midrule
        verify  & 162,689 (56.1\%)       & 39,336 (53.5\%) & 6,945 (50.4\%) \\
        choose  & 28,560 \enskip(9.8\%)  & 7,806  (10.6\%) & 1,523 (11.0\%) \\
        query   & 98,782 (34.1\%)        & 26,425 (35.9\%) & 5,325 (38.6\%) \\
        \bottomrule
        \end{tabular}
    }
\end{table}

\paragraph{\CheckedBox\;\;Comparison with other medical VQA datasets}
\cref{tab:vqastatistics} provides a comparison of MIMIC-CXR-VQA to other medical VQA datasets. 
Compared to other VQA datasets, MIMIC-CXR-VQA presents broader templates and covers a wider range of question types. 
While PathVQA, SLAKE, and P-VQA also have diverse question templates, they primarily focus on pathological questions and the utilization of medical knowledge graphs, which differs from our focus. 
We emphasize questions that can be answered by solely looking at the patient's X-ray image. 
When considering the number of templates, other datasets often categorize templates with minor linguistic differences as distinct, even if they express the same content and semantics (\eg, ``Is the \@POS scan normal?'' and ``Is the \@POS normal?''). 
We choose to unify these variations, denoting the count in parentheses in the ``\# Templates'' column of \cref{tab:vqastatistics}, thereby recognizing them as identical templates. 
This approach highlights that our dataset includes a significantly larger number of unique templates.\footnote{Note that the asterisk ($\ast$) in the ``\# Templates'' column indicates that the templates were either manually created by physicians, derived from natural questions, or undocumented, making it challenging to represent the number of templates accurately.} 
Furthermore, our dataset's templates are more complex than others, incorporating compositional templates developed through set/logical operations, which is not commonly observed in other datasets.

\begin{table}[H]
\caption{Statistics for MIMIC-CXR-VQA and comparisons with existing datasets.}
\label{tab:vqastatistics}
\resizebox{\textwidth}{!}{%
\begin{tabular}{@{}ccccccccc@{}}
\toprule
Dataset &
  \# Images &
  \# QA pairs &
  Source of images &
  QA  Creation &
  \# Question types &
  \# Templates &
  \multicolumn{1}{c}{Compositional} &
  Publicly accessible \\ \midrule
VQA-RAD &
  315 &
  3,515 &
  MedPix database &
  natural &
  11 &
  $\ast$ &
  &
  \checkmark
   \\
VQA-Med-2019 &
  4,200 &
  15,292 &
  MedPix database &
  synthetic &
  4 &
  $\ast$ &
   &
  \checkmark
   \\
PathVQA &
  4,998 &
  32,799 &
  Pathology textbook &
  synthetic &
  7 &
  $\ast$ &
  &
  \checkmark
   \\
VQA-Med-2020 &
  5,000 &
  5,000 &
  MedPix database &
  synthetic &
  1 &
  18 (8) &
   &
  \checkmark \\
SLAKE &
  642 &
  14,000 &
  \begin{tabular}[c]{@{}c@{}}Medical Decathlon\\ NIH Chest X-ray\\ CHAOS\end{tabular} &
  natural &
  10 &
  $\ast$ &
  &
  \checkmark \\
VQA-Med-2021 &
  5,000 &
  5,000 &
  MedPix database &
  synthetic &
  1 &
  6 (4) &
   &
  \checkmark \\
\hline
RadVisDial &
  91,060 &
  455,300 &
  MIMIC-CXR &
  synthetic, natural &
  1 &
  1 (1) &
   &
   \\
OVQA &
  2,001 &
  19,020 &
  EMRs &
  synthetic &
  6 &
  72 (19) &
   &
   \checkmark
   \\
Mimic-VQ &
  134,400 &
  297,723 &
  MIMIC-CXR &
  synthetic &
  6 &
  15 (13) &
   &
   \\
P-VQA &
  2,169 &
  24,800 &
  hospitals &
  synthetic &
  13 &
  $\ast$ &
   &
   \checkmark
   \\ \midrule
\textbf{MIMIC-CXR-VQA} &
  142,797 &
  377,391 &
  MIMIC-CXR &
  synthetic, paraphrased & 
  7 &
  794 (48) &
  \checkmark &
  \checkmark \\
\bottomrule
\end{tabular}%
}
\end{table}

\newpage
\paragraph{\CheckedBox\;\;Full list of VQA question template in MIMIC-CXR-VQA}\;\;
\begin{longtblr}
[
  caption = {Full list of 48 VQA question templates in MIMIC-CXR-VQA},
  label = {tab:question_template_full},
]
{
  colspec = {X[c,m]X[1.2,c,m]X[8,l,m]},
  colsep = 0.3pt,
  rowhead = 1,
  hlines,
  rows={font=\scriptsize},
  rowsep=0.3pt,
}
\textbf{Index} & \textbf{Content Type} & {\SetCell[c=1]{c}\textbf{Question Template}} \\
1 & presence & Are there any \$\{category\} in the \$\{object\}? \\
2 & presence & Is there \$\{attribute\} in the \$\{object\}? \\
3 & abnormality & Is the \$\{object\} abnormal? \\
4 & presence & Are there any \$\{category\_1\} or \$\{category\_2\} in the \$\{object\}? \\
5 & presence & Are there both \$\{attribute\_1\} and \$\{attribute\_2\} in the \$\{object\}? \\
6 & presence & Is there either \$\{attribute\_1\} or \$\{attribute\_2\} in the \$\{object\}? \\
7 & attribute & List all \$\{category\} in the \$\{object\}. \\
8 & abnormality & List all abnormalities in the \$\{object\}. \\
9 & attribute & List all \$\{category\_1\} and \$\{category\_2\} in the \$\{object\}. \\
10 & attribute & Which \$\{category\} is related to the \$\{object\}, \$\{attribute\_1\} or \$\{attribute\_2\}? \\
11 & abnormality & Are there any abnormalities in either the \$\{object\_1\} or the \$\{object\_2\}? \\
12 & abnormality & Are there any abnormalities in both the \$\{object\_1\} and the \$\{object\_2\}? \\
13 & attribute & List all \$\{category\} in either the \$\{object\_1\} or the \$\{object\_2\}. \\
14 & attribute & List all common \$\{category\} in both the \$\{object\_1\} and the \$\{object\_2\}. \\
15 & attribute & List all \$\{category\} only in the \$\{object\_1\} but not in the \$\{object\_2\}. \\
16 & abnormality & List all abnormalities in either the \$\{object\_1\} or the \$\{object\_2\}. \\
17 & abnormality & List all common abnormalities in both the \$\{object\_1\} and the \$\{object\_2\}. \\
18 & abnormality & List all abnormalities only in the \$\{object\_1\} but not in the \$\{object\_2\}. \\
19 & presence & Are there any \$\{category\}? \\
20 & abnormality & Are there any abnormalities? \\
21 & presence & Are there any \$\{category\_1\} or \$\{category\_2\}? \\
22 & presence & Is there \$\{attribute\}? \\
23 & presence & Are there both \$\{attribute\_1\} and \$\{attribute\_2\}? \\
24 & presence & Is there either \$\{attribute\_1\} or \$\{attribute\_2\}? \\
25 & attribute & List all \$\{category\}. \\
26 & attribute & List all \$\{category\_1\} and \$\{category\_2\}. \\
27 & abnormality & List all abnormalities. \\
28 & attribute & Which \$\{category\} is related, \$\{attribute\_1\} or \$\{attribute\_2\}? \\
29 & presence & Are both the \$\{object\_1\} and the \$\{object\_2\} related to \$\{attribute\}? \\
30 & presence & Is either the \$\{object\_1\} or the \$\{object\_2\} related to \$\{attribute\}? \\
31 & anatomy & List all anatomical locations related to \$\{attribute\}. \\
32 & anatomy & Which anatomical location is related to \$\{attribute\}, the \$\{object\_1\} or the \$\{object\_2\}? \\
33 & abnormality & Which anatomical location is abnormal, the \$\{object\_1\} or the \$\{object\_2\}? \\
34 & anatomy & List all anatomical locations related to either \$\{attribute\_1\} or \$\{attribute\_2\}. \\
35 & anatomy & List all common anatomical locations related to both \$\{attribute\_1\} and \$\{attribute\_2\}. \\
36 & anatomy & List all anatomical locations related to \$\{attribute\_1\} but not \$\{attribute\_2\}. \\
37 & presence & Are there any \$\{category\} related to the \$\{object\_1\} and the \$\{object\_2\}? \\
38 & presence & Are there any \$\{category\} related to the \$\{object\_1\} or the \$\{object\_2\}? \\
39 & anatomy & List all anatomical locations related to any \$\{category\}. \\
40 & anatomy & List all anatomical locations related to any \$\{category\_1\} or \$\{category\_2\}. \\
41 & plane & Is this an \$\{viewpos\} view? \\
42 & plane & Which view is in this image, AP or PA? \\
43 & plane & What is the view of this image? \\
44 & gender & Is this patient \$\{gender\}? \\
45 & gender & What is the gender of this patient, male or female? \\
46 & gender & What is the gender of this patient? \\
47 & size & Is the width of the cardiac silhouette wider than 1/2 of the thorax width? \\
48 & size & Is the width of the upper mediastinum wider than 1/3 of the thorax width?
\end{longtblr}

\newpage
\section{EHRXQA} \label{supp_EHRXQA}
\subsection{Database construction} \label{supp_EHRXQA_dataconstruct}
\subsubsection{Database pre-processing}
\begin{itemize}[itemsep=1pt, topsep=-2pt, leftmargin=7mm]
    \item We create a ``\textit{dod}'' column to the PATIENTS table and assign it the date of birth calculated as follows: \texttt{dob = anchor\_year - anchor\_age}. The month and day of dob are randomly sampled.
    \item We create an ``\textit{age}'' column in the ADMISSIONS table. To calculate the age at the time of admission for each subject, we subtract their anchor year from their admission year and then add their anchor age. This can be represented as: \texttt{age = (admission\_year - anchor\_year) + anchor\_age}.
    \item We includes patients aged between 11 and 89.
    \item We manually time-shift each patient's first study time to a random time point between 2100 and 2105, while preserving the same intervals between all records.
    \item We limit the number of current patients to approximately 10\% of the total patient population.
    \item We sample 800 patients for the silver database and 400 for the gold database.
    \item  If a particular type of value has multiple associated units of measurement, we retain only the value with the most common unit and discard the others from the database.
    \item All records are converted to lowercase.
\end{itemize}

\subsubsection{Overview of EHR database}
\begin{figure}[H]
    \centering
    \includegraphics[width=1.0\columnwidth]{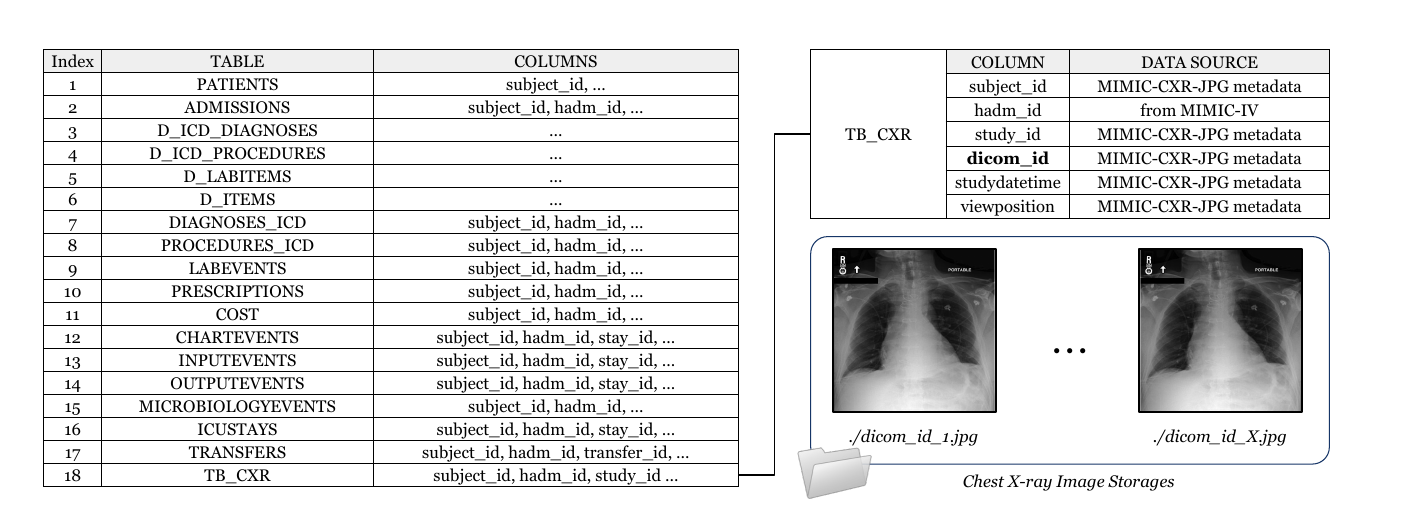}
    \caption{Overview of our EHR database schema, which comprises 18 tables: 17 from MIMIC-IV and one from MIMIC-CXR. The TB\_CXR table includes DICOM image identifiers, enabling the loading of corresponding images directly from CXR image storage.}
    \label{fig:overview_db_schema}
\end{figure}

The entire database schema is illustrated in \cref{fig:overview_db_schema}.

\newpage
\subsection{Question template construction} \label{supp_EHRXQA_templateconstruct}
\subsubsection{Detail of template construction strategy}
\paragraph{\CheckedBox\;\;\textit{Image}-related question}
In the EHRXQA dataset, we present a unique scenario that diverges from the traditional Visual Question Answering (VQA) framework. 
In this scenario, instead of providing just an image and a question, our question templates also involve retrieving relevant images from a database to answer the question. 
However, specifying a particular study by directly stating its unique study ID can be complex and inconvenient for users, especially when questions refer to one or two images. To address this issue, we propose using practical, conversational expressions such as ``\textit{last study of patient {patient\_id} in 2023}'' or ``\textit{compared to the previous study}''.
This approach allows users to intuitively identify chest X-ray (CXR) studies using a variety of practical natural language expressions, as demonstrated in ~\cref{tab:study_indication}.

Regarding templates for 2-image question templates, our aim is to replicate clinical scenarios where we compare two separate or consecutive patient studies conducted during a hospital stay. In these situations, determining the severity of a particular disease can be subjective across different physicians, which can lead to higher costs for labeling.
To address this issue, we have established four comparison labels: ``\textit{still present}'', ``\textit{still absent}'', ``\textit{newly detected}'' and ``\textit{resolved}''. 
These labels are automatically assigned based on changes in the presence of an attribute in an object.
For example, if an initial study shows no signs of pneumonia in the left lung, but a subsequent study reveals the presence of pneumonia, the comparison label would be assigned as ``\textit{newly detected}''.

\begin{table}[h]
\caption{Example expressions for indicating CXR Studies in EHRXQA Dataset}
\label{tab:study_indication}
\centering
\renewcommand{\arraystretch}{1.5} 
\resizebox{0.7\textwidth}{!}{%
    \begin{tabularx}{\textwidth}{@{}cXX@{}}
        \toprule
        \# of images     
        & reference target study  
        & reference to compared study \\ 
        \midrule
        \multirow{3.5}{*}{1-image} 
        & study id \texttt{\{study\_id\}}
        & - \\ \cline{2-3}
        & \texttt{[time\_filter\_exact1]} study of patient \texttt{\{patient\_id\}} \texttt{[time\_filter\_global1]} 
        & \multirow{3}{*}{-} \\ 
        \midrule
        \multirow{6.5}{*}{2-image} 
        & study id \texttt{\{study\_id1\}} 
        & study id \texttt{\{study\_id2\}} \\ \cline{2-3}
        & study id \texttt{\{study\_id1\}}
        & previous study \\ \cline{2-3}
        & \texttt{[time\_filter\_exact1]} study of patient \texttt{\{patient\_id\}} \texttt{[time\_filter\_global1]}
        & \texttt{[time\_filter\_exact2]} study of patient \texttt{\{patient\_id\}} \texttt{[time\_filter\_global2]} \\ \cline{2-3}
        & \texttt{[time\_filter\_exact1]} study of patient \texttt{\{patient\_id\}} \texttt{[time\_filter\_global1]} & \multirow{3}{*}{previous study} \\ 
        \bottomrule
    \end{tabularx}
}
\end{table}

\paragraph{\CheckedBox\;\;\textit{Image+Table}-related question} 
In the \textit{Image+Table} modality, our focus is on three primary table events found in structured electronic health records (EHRs) along with CXR events: diagnosis, prescriptions, and procedures. It's important to note that CXR events are considered at the same admission level as these three medical events, indicating their comparable temporal hierarchy.
By incorporating both structured medical events and CXR events, we explore three primary temporal scenarios: co-occurring events, where the events happen simultaneously; CXR events that occur after table events; and table events that follow the CXR events.
Furthermore, when constructing our question templates in the \textit{Image+Table} modality, we take into account demographic information such as age and gender.

\subsubsection{Full List of Question Template in EHRXQA}
We provide a comprehensive collection of question templates for your reference: 168 image-related templates can be found in \cref{tab:img_template_full}, 174 table-related templates are detailed in \cref{tab:tab_template_full}, and a further 75 templates are presented in \cref{tab:mm_template_full}.

\begin{longtblr}
[
  caption = {Full list of 168 \textit{Image}-related question templates in EHRXQA},
  label = {tab:img_template_full},
]
{
  colspec = {X[1.2,c,m] X[c,m] X[c,m]X[10,l,m]},
  colsep = 0.2pt,
  rowhead = 1,
  hlines,
  rows={font=\tiny},
  rowsep=0.3pt,
}
\textbf{Patient scope} & {\SetCell[c=2]{c} \textbf{Modality scope}} &  & {\SetCell[c=1]{c}\textbf{Question Template}} \\ 
single & Image & 1-image & Given the [time\_filter\_exact1] study of patient \{patient\_id\} [time\_filter\_global1], are there any \$\{category\} in the \$\{object\}? \\
single & Image & 1-image & Given the [time\_filter\_exact1] study of patient \{patient\_id\} [time\_filter\_global1], is there \$\{attribute\} in the \$\{object\}? \\
single & Image & 1-image & Given the [time\_filter\_exact1] study of patient \{patient\_id\} [time\_filter\_global1], is the \$\{object\} abnormal? \\
single & Image & 1-image & Given the [time\_filter\_exact1] study of patient \{patient\_id\} [time\_filter\_global1], are there any \$\{category\_1\} or \$\{category\_2\} in the \$\{object\}? \\
single & Image & 1-image & Given the [time\_filter\_exact1] study of patient \{patient\_id\} [time\_filter\_global1], are there both \$\{attribute\_1\} and \$\{attribute\_2\} in the \$\{object\}? \\
single & Image & 1-image & Given the [time\_filter\_exact1] study of patient \{patient\_id\} [time\_filter\_global1], is there either \$\{attribute\_1\} or \$\{attribute\_2\} in the \$\{object\}? \\
single & Image & 1-image & Given the [time\_filter\_exact1] study of patient \{patient\_id\} [time\_filter\_global1], list all \$\{category\} in the \$\{object\}. \\
single & Image & 1-image & Given the [time\_filter\_exact1] study of patient \{patient\_id\} [time\_filter\_global1], list all abnormality in the \$\{object\}. \\
single & Image & 1-image & Given the [time\_filter\_exact1] study of patient \{patient\_id\} [time\_filter\_global1], list all \$\{category\_1\} and \$\{category\_2\} in the \$\{object\}. \\
single & Image & 1-image & Given the [time\_filter\_exact1] study of patient \{patient\_id\} [time\_filter\_global1], which \$\{category\} is related to the \$\{object\}, \$\{attribute\_1\} or \$\{attribute\_2\}? \\
single & Image & 1-image & Given the [time\_filter\_exact1] study of patient \{patient\_id\} [time\_filter\_global1], are there any abnormality in either the \$\{object\_1\} or the \$\{object\_2\}? \\
single & Image & 1-image & Given the [time\_filter\_exact1] study of patient \{patient\_id\} [time\_filter\_global1], are there any abnormality in both the \$\{object\_1\} and the \$\{object\_2\}? \\
single & Image & 1-image & Given the [time\_filter\_exact1] study of patient \{patient\_id\} [time\_filter\_global1], list all \$\{category\} in either the \$\{object\_1\} or the \$\{object\_2\}. \\
single & Image & 1-image & Given the [time\_filter\_exact1] study of patient \{patient\_id\} [time\_filter\_global1], list all common \$\{category\} in both the \$\{object\_1\} and the \$\{object\_2\}. \\
single & Image & 1-image & Given the [time\_filter\_exact1] study of patient \{patient\_id\} [time\_filter\_global1], list all \$\{category\} only in the \$\{object\_1\} but not in the \$\{object\_2\}. \\
single & Image & 1-image & Given the [time\_filter\_exact1] study of patient \{patient\_id\} [time\_filter\_global1], list all abnormality in either the \$\{object\_1\} or the \$\{object\_2\}. \\
single & Image & 1-image & Given the [time\_filter\_exact1] study of patient \{patient\_id\} [time\_filter\_global1], list all common abnormality in both the \$\{object\_1\} and the \$\{object\_2\}. \\
single & Image & 1-image & Given the [time\_filter\_exact1] study of patient \{patient\_id\} [time\_filter\_global1], list all abnormality only in the \$\{object\_1\} but not in the \$\{object\_2\}. \\
single & Image & 1-image & Given the [time\_filter\_exact1] study of patient \{patient\_id\} [time\_filter\_global1], are there any \$\{category\}? \\
single & Image & 1-image & Given the [time\_filter\_exact1] study of patient \{patient\_id\} [time\_filter\_global1], are there any abnormality? \\
single & Image & 1-image & Given the [time\_filter\_exact1] study of patient \{patient\_id\} [time\_filter\_global1], are there any \$\{category\_1\} or \$\{category\_2\}? \\
single & Image & 1-image & Given the [time\_filter\_exact1] study of patient \{patient\_id\} [time\_filter\_global1], is there \$\{attribute\}? \\
single & Image & 1-image & Given the [time\_filter\_exact1] study of patient \{patient\_id\} [time\_filter\_global1], are there both \$\{attribute\_1\} and \$\{attribute\_2\}? \\
single & Image & 1-image & Given the [time\_filter\_exact1] study of patient \{patient\_id\} [time\_filter\_global1], is there either \$\{attribute\_1\} or \$\{attribute\_2\}? \\
single & Image & 1-image & Given the [time\_filter\_exact1] study of patient \{patient\_id\} [time\_filter\_global1], list all \$\{category\}. \\
single & Image & 1-image & Given the [time\_filter\_exact1] study of patient \{patient\_id\} [time\_filter\_global1], list all \$\{category\_1\} and \$\{category\_2\}. \\
single & Image & 1-image & Given the [time\_filter\_exact1] study of patient \{patient\_id\} [time\_filter\_global1], list all abnormality. \\
single & Image & 1-image & Given the [time\_filter\_exact1] study of patient \{patient\_id\} [time\_filter\_global1], which \$\{category\} is related, \$\{attribute\_1\} or \$\{attribute\_2\}? \\
single & Image & 1-image & Given the [time\_filter\_exact1] study of patient \{patient\_id\} [time\_filter\_global1], are both the \$\{object\_1\} and the \$\{object\_2\} related to \$\{attribute\}? \\
single & Image & 1-image & Given the [time\_filter\_exact1] study of patient \{patient\_id\} [time\_filter\_global1], is either the \$\{object\_1\} or the \$\{object\_2\} related to \$\{attribute\}? \\
single & Image & 1-image & Given the [time\_filter\_exact1] study of patient \{patient\_id\} [time\_filter\_global1], list all anatomical locations related to \$\{attribute\}. \\
single & Image & 1-image & Given the [time\_filter\_exact1] study of patient \{patient\_id\} [time\_filter\_global1], which anatomical location is related to \$\{attribute\}, the \$\{object\_1\} or the \$\{object\_2\}? \\
single & Image & 1-image & Given the [time\_filter\_exact1] study of patient \{patient\_id\} [time\_filter\_global1], which anatomical location is abnormal, the \$\{object\_1\} or the \$\{object\_2\}? \\
single & Image & 1-image & Given the [time\_filter\_exact1] study of patient \{patient\_id\} [time\_filter\_global1], list all anatomical locations related to either \$\{attribute\_1\} or \$\{attribute\_2\}. \\
single & Image & 1-image & Given the [time\_filter\_exact1] study of patient \{patient\_id\} [time\_filter\_global1], list all common anatomical locations related to both \$\{attribute\_1\} and \$\{attribute\_2\}. \\
single & Image & 1-image & Given the [time\_filter\_exact1] study of patient \{patient\_id\} [time\_filter\_global1], list all anatomical locations related to \$\{attribute\_1\} but not \$\{attribute\_2\}. \\
single & Image & 1-image & Given the [time\_filter\_exact1] study of patient \{patient\_id\} [time\_filter\_global1], are there any \$\{category\} related to the \$\{object\_1\} and the \$\{object\_2\}? \\
single & Image & 1-image & Given the [time\_filter\_exact1] study of patient \{patient\_id\} [time\_filter\_global1], are there any \$\{category\} related to the \$\{object\_1\} or the \$\{object\_2\}? \\
single & Image & 1-image & Given the [time\_filter\_exact1] study of patient \{patient\_id\} [time\_filter\_global1], list all anatomical locations related to any \$\{category\}. \\
single & Image & 1-image & Given the [time\_filter\_exact1] study of patient \{patient\_id\} [time\_filter\_global1], list all anatomical locations related to any \$\{category\_1\} or \$\{category\_2\}. \\
single & Image & 1-image & Given the [time\_filter\_exact1] study of patient \{patient\_id\} [time\_filter\_global1], is the width of the cardiac silhouette wider than 1/2 of the thorax width? \\
single & Image & 1-image & Given the [time\_filter\_exact1] study of patient \{patient\_id\} [time\_filter\_global1], is the width of the upper mediastinum wider than 1/3 of the thorax width? \\
single & Image & 1-image & Given the study \{study\_id\}, are there any \$\{category\} in the \$\{object\}? \\
single & Image & 1-image & Given the study \{study\_id\}, is there \$\{attribute\} in the \$\{object\}? \\
single & Image & 1-image & Given the study \{study\_id\}, is the \$\{object\} abnormal? \\
single & Image & 1-image & Given the study \{study\_id\}, are there any \$\{category\_1\} or \$\{category\_2\} in the \$\{object\}? \\
single & Image & 1-image & Given the study \{study\_id\}, are there both \$\{attribute\_1\} and \$\{attribute\_2\} in the \$\{object\}? \\
single & Image & 1-image & Given the study \{study\_id\}, is there either \$\{attribute\_1\} or \$\{attribute\_2\} in the \$\{object\}? \\
single & Image & 1-image & Given the study \{study\_id\}, list all \$\{category\} in the \$\{object\}. \\
single & Image & 1-image & Given the study \{study\_id\}, list all abnormality in the \$\{object\}. \\
single & Image & 1-image & Given the study \{study\_id\}, list all \$\{category\_1\} and \$\{category\_2\} in the \$\{object\}. \\
single & Image & 1-image & Given the study \{study\_id\}, which \$\{category\} is related to the \$\{object\}, \$\{attribute\_1\} or \$\{attribute\_2\}? \\
single & Image & 1-image & Given the study \{study\_id\}, are there any abnormality in either the \$\{object\_1\} or the \$\{object\_2\}? \\
single & Image & 1-image & Given the study \{study\_id\}, are there any abnormality in both the \$\{object\_1\} and the \$\{object\_2\}? \\
single & Image & 1-image & Given the study \{study\_id\}, list all \$\{category\} in either the \$\{object\_1\} or the \$\{object\_2\}. \\
single & Image & 1-image & Given the study \{study\_id\}, list all common \$\{category\} in both the \$\{object\_1\} and the \$\{object\_2\}. \\
single & Image & 1-image & Given the study \{study\_id\}, list all \$\{category\} only in the \$\{object\_1\} but not in the \$\{object\_2\}. \\
single & Image & 1-image & Given the study \{study\_id\}, list all abnormality in either the \$\{object\_1\} or the \$\{object\_2\}. \\
single & Image & 1-image & Given the study \{study\_id\}, list all common abnormality in both the \$\{object\_1\} and the \$\{object\_2\}. \\
single & Image & 1-image & Given the study \{study\_id\}, list all abnormality only in the \$\{object\_1\} but not in the \$\{object\_2\}. \\
single & Image & 1-image & Given the study \{study\_id\}, are there any \$\{category\}? \\
single & Image & 1-image & Given the study \{study\_id\}, are there any abnormality? \\
single & Image & 1-image & Given the study \{study\_id\}, are there any \$\{category\_1\} or \$\{category\_2\}? \\
single & Image & 1-image & Given the study \{study\_id\}, is there \$\{attribute\}? \\
single & Image & 1-image & Given the study \{study\_id\}, are there both \$\{attribute\_1\} and \$\{attribute\_2\}? \\
single & Image & 1-image & Given the study \{study\_id\}, is there either \$\{attribute\_1\} or \$\{attribute\_2\}? \\
single & Image & 1-image & Given the study \{study\_id\}, list all \$\{category\}. \\
single & Image & 1-image & Given the study \{study\_id\}, list all \$\{category\_1\} and \$\{category\_2\}. \\
single & Image & 1-image & Given the study \{study\_id\}, list all abnormality. \\
single & Image & 1-image & Given the study \{study\_id\}, which \$\{category\} is related, \$\{attribute\_1\} or \$\{attribute\_2\}? \\
single & Image & 1-image & Given the study \{study\_id\}, are both the \$\{object\_1\} and the \$\{object\_2\} related to \$\{attribute\}? \\
single & Image & 1-image & Given the study \{study\_id\}, is either the \$\{object\_1\} or the \$\{object\_2\} related to \$\{attribute\}? \\
single & Image & 1-image & Given the study \{study\_id\}, list all anatomical locations related to \$\{attribute\}. \\
single & Image & 1-image & Given the study \{study\_id\}, which anatomical location is related to \$\{attribute\}, the \$\{object\_1\} or the \$\{object\_2\}? \\
single & Image & 1-image & Given the study \{study\_id\}, which anatomical location is abnormal, the \$\{object\_1\} or the \$\{object\_2\}? \\
single & Image & 1-image & Given the study \{study\_id\}, list all anatomical locations related to either \$\{attribute\_1\} or \$\{attribute\_2\}. \\
single & Image & 1-image & Given the study \{study\_id\}, list all common anatomical locations related to both \$\{attribute\_1\} and \$\{attribute\_2\}. \\
single & Image & 1-image & Given the study \{study\_id\}, list all anatomical locations related to \$\{attribute\_1\} but not \$\{attribute\_2\}. \\
single & Image & 1-image & Given the study \{study\_id\}, are there any \$\{category\} related to the \$\{object\_1\} and the \$\{object\_2\}? \\
single & Image & 1-image & Given the study \{study\_id\}, are there any \$\{category\} related to the \$\{object\_1\} or the \$\{object\_2\}? \\
single & Image & 1-image & Given the study \{study\_id\}, list all anatomical locations related to any \$\{category\}. \\
single & Image & 1-image & Given the study \{study\_id\}, list all anatomical locations related to any \$\{category\_1\} or \$\{category\_2\}. \\
single & Image & 1-image & Given the study \{study\_id\}, is the width of the cardiac silhouette wider than 1/2 of the thorax width? \\
single & Image & 1-image & Given the study \{study\_id\}, is the width of the upper mediastinum wider than 1/3 of the thorax width? \\
single & Image & 2-image & Given the [time\_filter\_exact1] study of patient \{patient\_id\} [time\_filter\_global1], are there any \$\{category\} that are \$\{comparison\} in the \$\{object\} compared to the [time\_filter\_exact2] study of patient \{patient\_id\} [time\_filter\_global2]? \\
single & Image & 2-image & Given the [time\_filter\_exact1] study of patient \{patient\_id\} [time\_filter\_global1], is \$\{attribute\} \$\{comparison\} in the \$\{object\} compared to the [time\_filter\_exact2] study of patient \{patient\_id\} [time\_filter\_global2]? \\
single & Image & 2-image & Given the [time\_filter\_exact1] study of patient \{patient\_id\} [time\_filter\_global1], are there any abnormality that are \$\{comparison\} in the \$\{object\} compared to the [time\_filter\_exact2] study of patient \{patient\_id\} [time\_filter\_global2]? \\
single & Image & 2-image & Given the [time\_filter\_exact1] study of patient \{patient\_id\} [time\_filter\_global1], are there any \$\{category\} that are \$\{comparison\} compared to the [time\_filter\_exact2] study of patient \{patient\_id\} [time\_filter\_global2]? \\
single & Image & 2-image & Given the [time\_filter\_exact1] study of patient \{patient\_id\} [time\_filter\_global1], are there any abnormality that are \$\{comparison\} compared to the [time\_filter\_exact2] study of patient \{patient\_id\} [time\_filter\_global2]? \\
single & Image & 2-image & Given the [time\_filter\_exact1] study of patient \{patient\_id\} [time\_filter\_global1], is \$\{attribute\} \$\{comparison\} compared to the [time\_filter\_exact2] study of patient \{patient\_id\} [time\_filter\_global2]? \\
single & Image & 2-image & Given the [time\_filter\_exact1] study of patient \{patient\_id\} [time\_filter\_global1], list all \$\{category\} that are \$\{comparison\} in the \$\{object\} compared to the [time\_filter\_exact2] study of patient \{patient\_id\} [time\_filter\_global2]? \\
single & Image & 2-image & Given the [time\_filter\_exact1] study of patient \{patient\_id\} [time\_filter\_global1], list all abnormality that are \$\{comparison\} in the \$\{object\} compared to the [time\_filter\_exact2] study of patient \{patient\_id\} [time\_filter\_global2]? \\
single & Image & 2-image & Given the [time\_filter\_exact1] study of patient \{patient\_id\} [time\_filter\_global1], list all \$\{category\} that are \$\{comparison\} compared to the [time\_filter\_exact2] study of patient \{patient\_id\} [time\_filter\_global2]? \\
single & Image & 2-image & Given the [time\_filter\_exact1] study of patient \{patient\_id\} [time\_filter\_global1], list all abnormality that are \$\{comparison\} compared to the [time\_filter\_exact2] study of patient \{patient\_id\} [time\_filter\_global2]? \\
single & Image & 2-image & Given the [time\_filter\_exact1] study of patient \{patient\_id\} [time\_filter\_global1], list all anatomical locations related to \$\{attribute\} that are \$\{comparison\} compared to the [time\_filter\_exact2] study of patient \{patient\_id\} [time\_filter\_global2]? \\
single & Image & 2-image & Given the [time\_filter\_exact1] study of patient \{patient\_id\} [time\_filter\_global1], list all anatomical locations related to any \$\{category\} that are \$\{comparison\} compared to the [time\_filter\_exact2] study of patient \{patient\_id\} [time\_filter\_global2]? \\
single & Image & 2-image & Given the [time\_filter\_exact1] study of patient \{patient\_id\} [time\_filter\_global1], are there any \$\{category\} that are \$\{comparison\} in the \$\{object\} compared to the previous study? \\
single & Image & 2-image & Given the [time\_filter\_exact1] study of patient \{patient\_id\} [time\_filter\_global1], is \$\{attribute\} \$\{comparison\} in the \$\{object\} compared to the previous study? \\
single & Image & 2-image & Given the [time\_filter\_exact1] study of patient \{patient\_id\} [time\_filter\_global1], are there any abnormality that are \$\{comparison\} in the \$\{object\} compared to the previous study? \\
single & Image & 2-image & Given the [time\_filter\_exact1] study of patient \{patient\_id\} [time\_filter\_global1], are there any \$\{category\} that are \$\{comparison\} compared to the previous study? \\
single & Image & 2-image & Given the [time\_filter\_exact1] study of patient \{patient\_id\} [time\_filter\_global1], are there any abnormality that are \$\{comparison\} compared to the previous study? \\
single & Image & 2-image & Given the [time\_filter\_exact1] study of patient \{patient\_id\} [time\_filter\_global1], is \$\{attribute\} \$\{comparison\} compared to the previous study? \\
single & Image & 2-image & Given the [time\_filter\_exact1] study of patient \{patient\_id\} [time\_filter\_global1], list all \$\{category\} that are \$\{comparison\} in the \$\{object\} compared to the previous study? \\
single & Image & 2-image & Given the [time\_filter\_exact1] study of patient \{patient\_id\} [time\_filter\_global1], list all abnormality that are \$\{comparison\} in the \$\{object\} compared to the previous study? \\
single & Image & 2-image & Given the [time\_filter\_exact1] study of patient \{patient\_id\} [time\_filter\_global1], list all \$\{category\} that are \$\{comparison\} compared to the previous study? \\
single & Image & 2-image & Given the [time\_filter\_exact1] study of patient \{patient\_id\} [time\_filter\_global1], list all abnormality that are \$\{comparison\} compared to the previous study? \\
single & Image & 2-image & Given the [time\_filter\_exact1] study of patient \{patient\_id\} [time\_filter\_global1], list all anatomical locations related to \$\{attribute\} that are \$\{comparison\} compared to the previous study? \\
single & Image & 2-image & Given the [time\_filter\_exact1] study of patient \{patient\_id\} [time\_filter\_global1], list all anatomical locations related to any \$\{category\} that are \$\{comparison\} compared to the previous study? \\
single & Image & 2-image & Given the \{study\_id1\} study, are there any \$\{category\} that are \$\{comparison\} in the \$\{object\} compared to the \{study\_id2\} study? \\
single & Image & 2-image & Given the \{study\_id1\} study, is \$\{attribute\} \$\{comparison\} in the \$\{object\} compared to the \{study\_id2\} study? \\
single & Image & 2-image & Given the \{study\_id1\} study, are there any abnormality that are \$\{comparison\} in the \$\{object\} compared to the \{study\_id2\} study? \\
single & Image & 2-image & Given the \{study\_id1\} study, are there any \$\{category\} that are \$\{comparison\} compared to the \{study\_id2\} study? \\
single & Image & 2-image & Given the \{study\_id1\} study, are there any abnormality that are \$\{comparison\} compared to the \{study\_id2\} study? \\
single & Image & 2-image & Given the \{study\_id1\} study, is \$\{attribute\} \$\{comparison\} compared to the \{study\_id2\} study? \\
single & Image & 2-image & Given the \{study\_id1\} study, list all \$\{category\} that are \$\{comparison\} in the \$\{object\} compared to the \{study\_id2\} study? \\
single & Image & 2-image & Given the \{study\_id1\} study, list all abnormality that are \$\{comparison\} in the \$\{object\} compared to the \{study\_id2\} study? \\
single & Image & 2-image & Given the \{study\_id1\} study, list all \$\{category\} that are \$\{comparison\} compared to the \{study\_id2\} study? \\
single & Image & 2-image & Given the \{study\_id1\} study, list all abnormality that are \$\{comparison\} compared to the \{study\_id2\} study? \\
single & Image & 2-image & Given the \{study\_id1\} study, list all anatomical locations related to \$\{attribute\} that are \$\{comparison\} compared to the \{study\_id2\} study? \\
single & Image & 2-image & Given the \{study\_id1\} study, list all anatomical locations related to any \$\{category\} that are \$\{comparison\} compared to the \{study\_id2\} study? \\
single & Image & 2-image & Given the \{study\_id1\} study, are there any \$\{category\} that are \$\{comparison\} in the \$\{object\} compared to the previous study? \\
single & Image & 2-image & Given the \{study\_id1\} study, is \$\{attribute\} \$\{comparison\} in the \$\{object\} compared to the previous study? \\
single & Image & 2-image & Given the \{study\_id1\} study, are there any abnormality that are \$\{comparison\} in the \$\{object\} compared to the previous study? \\
single & Image & 2-image & Given the \{study\_id1\} study, are there any \$\{category\} that are \$\{comparison\} compared to the previous study? \\
single & Image & 2-image & Given the \{study\_id1\} study, are there any abnormality that are \$\{comparison\} compared to the previous study? \\
single & Image & 2-image & Given the \{study\_id1\} study, is \$\{attribute\} \$\{comparison\} compared to the previous study? \\
single & Image & 2-image & Given the \{study\_id1\} study, list all \$\{category\} that are \$\{comparison\} in the \$\{object\} compared to the previous study? \\
single & Image & 2-image & Given the \{study\_id1\} study, list all abnormality that are \$\{comparison\} in the \$\{object\} compared to the previous study? \\
single & Image & 2-image & Given the \{study\_id1\} study, list all \$\{category\} that are \$\{comparison\} compared to the previous study? \\
single & Image & 2-image & Given the \{study\_id1\} study, list all abnormality that are \$\{comparison\} compared to the previous study? \\
single & Image & 2-image & Given the \{study\_id1\} study, list all anatomical locations related to \$\{attribute\} that are \$\{comparison\} compared to the previous study? \\
single & Image & 2-image & Given the \{study\_id1\} study, list all anatomical locations related to any \$\{category\} that are \$\{comparison\} compared to the previous study? \\
single & Image & N-image & How many [unit\_count] have passed since the [time\_filter\_exact1] time patient \{patient\_id\} had a chest X-ray study indicating \$\{attribute\} in the \$\{object\} [time\_filter\_global1]? \\
single & Image & N-image & How many [unit\_count] have passed since the [time\_filter\_exact1] time patient \{patient\_id\} had a chest X-ray study indicating any \$\{category\} in the \$\{object\} [time\_filter\_global1]? \\
single & Image & N-image & How many [unit\_count] have passed since the [time\_filter\_exact1] time patient \{patient\_id\} had a chest X-ray study indicating any abnormality in the \$\{object\} [time\_filter\_global1]? \\
single & Image & N-image & How many [unit\_count] have passed since the [time\_filter\_exact1] time patient \{patient\_id\} had a chest X-ray study indicating \$\{attribute\} [time\_filter\_global1]? \\
single & Image & N-image & How many [unit\_count] have passed since the [time\_filter\_exact1] time patient \{patient\_id\} had a chest X-ray study indicating any \$\{category\} [time\_filter\_global1]? \\
single & Image & N-image & How many [unit\_count] have passed since the [time\_filter\_exact1] time patient \{patient\_id\} had a chest X-ray study indicating any abnormality [time\_filter\_global1]? \\
single & Image & N-image & When was the [time\_filter\_exact1] time that patient \{patient\_id\} had a chest X-ray study indicating \$\{attribute\} in the \$\{object\} [time\_filter\_global1]? \\
single & Image & N-image & When was the [time\_filter\_exact1] time that patient \{patient\_id\} had a chest X-ray study indicating any \$\{category\} in the \$\{object\} [time\_filter\_global1]? \\
single & Image & N-image & When was the [time\_filter\_exact1] time that patient \{patient\_id\} had a chest X-ray study indicating any abnormality in the \$\{object\} [time\_filter\_global1]? \\
single & Image & N-image & When was the [time\_filter\_exact1] time that patient \{patient\_id\} had a chest X-ray study indicating \$\{attribute\} [time\_filter\_global1]? \\
single & Image & N-image & When was the [time\_filter\_exact1] time that patient \{patient\_id\} had a chest X-ray study indicating any \$\{category\} [time\_filter\_global1]? \\
single & Image & N-image & When was the [time\_filter\_exact1] time that patient \{patient\_id\} had a chest X-ray study indicating any abnormality [time\_filter\_global1]? \\
single & Image & N-image & Has \{patient\_id\} had any chest X-ray study indicating \$\{attribute\} in the \$\{object\} [time\_filter\_global1]? \\
single & Image & N-image & Has \{patient\_id\} had any chest X-ray study indicating any \$\{category\} in the \$\{object\} [time\_filter\_global1]? \\
single & Image & N-image & Has \{patient\_id\} had any chest X-ray study indicating any abnormality in the \$\{object\} [time\_filter\_global1]? \\
single & Image & N-image & Has \{patient\_id\} had any chest X-ray study indicating \$\{attribute\} [time\_filter\_global1]? \\
single & Image & N-image & Has \{patient\_id\} had any chest X-ray study indicating any \$\{category\} [time\_filter\_global1]? \\
single & Image & N-image & Has \{patient\_id\} had any chest X-ray study indicating any abnormality [time\_filter\_global1]? \\
single & Image & N-image & Count the number of times that patient \{patient\_id\} had chest X-ray studies indicating \$\{attribute\} in the \$\{object\} [time\_filter\_global1]. \\
single & Image & N-image & Count the number of times that patient \{patient\_id\} had chest X-ray studies indicating any \$\{category\} in the \$\{object\} [time\_filter\_global1]. \\
single & Image & N-image & Count the number of times that patient \{patient\_id\} had chest X-ray studies indicating any abnormality in the \$\{object\} [time\_filter\_global1]. \\
single & Image & N-image & Count the number of times that patient \{patient\_id\} had chest X-ray studies indicating \$\{attribute\} [time\_filter\_global1]. \\
single & Image & N-image & Count the number of times that patient \{patient\_id\} had chest X-ray studies indicating any \$\{category\} [time\_filter\_global1]. \\
single & Image & N-image & Count the number of times that patient \{patient\_id\} had chest X-ray studies indicating any abnormality [time\_filter\_global1]. \\
group & Image & N-image & Count the number of patients who had any chest X-ray study indicating \$\{attribute\} in the \$\{object\} [time\_filter\_global1]. \\
group & Image & N-image & Count the number of patients who had any chest X-ray study indicating any \$\{category\} in the \$\{object\} [time\_filter\_global1]. \\
group & Image & N-image & Count the number of patients who had any chest X-ray study indicating any abnormality in the \$\{object\} [time\_filter\_global1]. \\
group & Image & N-image & Count the number of patients who had any chest X-ray study indicating \$\{attribute\} [time\_filter\_global1]. \\
group & Image & N-image & Count the number of patients who had any chest X-ray study indicating any \$\{category\} [time\_filter\_global1]. \\
group & Image & N-image & Count the number of patients who had any chest X-ray study indicating any abnormality [time\_filter\_global1]. \\
group & Image & N-image & List the IDs of patients who had any chest X-ray study indicating \$\{attribute\} in the \$\{object\} [time\_filter\_global1]. \\
group & Image & N-image & List the IDs of patients who had any chest X-ray study indicating any \$\{category\} in the \$\{object\} [time\_filter\_global1]. \\
group & Image & N-image & List the IDs of patients who had any chest X-ray study indicating any abnormality in the \$\{object\} [time\_filter\_global1]. \\
group & Image & N-image & List the IDs of patients who had any chest X-ray study indicating \$\{attribute\} [time\_filter\_global1]. \\
group & Image & N-image & List the IDs of patients who had any chest X-ray study indicating any \$\{category\} [time\_filter\_global1]. \\
group & Image & N-image & List the IDs of patients who had any chest X-ray study indicating any abnormality [time\_filter\_global1]. \\
\end{longtblr}
\begin{longtblr}
[
  caption = {Full list of 174 \textit{Table}-related question templates in EHRXQA},
  label = {tab:tab_template_full},
]
{
  colspec = {X[1.4,c,m]X[1.4,c,m]X[10,l,m]},
  colsep = 0.3pt,
  rowhead = 1,
  hlines,
  rows={font=\tiny},
  rowsep=0.3pt,
}
\textbf{Patient scope} & \textbf{Modality scope} & {\SetCell[c=1]{c}\textbf{Question Template}} \\
none & Table & What is the intake method of \{drug\_name\}? \\
none & Table & What is the cost of a procedure named \{procedure\_name\}? \\
none & Table & What is the cost of a \{lab\_name\} lab test? \\
none & Table & What is the cost of a drug named \{drug\_name\}? \\
none & Table & What is the cost of diagnosing \{diagnosis\_name\}? \\
none & Table & What does \{abbreviation\} stand for? \\
single & Table & What is the gender of patient \{patient\_id\}? \\
single & Table & What is the date of birth of patient \{patient\_id\}? \\
single & Table & What was the [time\_filter\_exact1] length of hospital stay of patient \{patient\_id\}? \\
single & Table & What is the change in the weight of patient \{patient\_id\} from the [time\_filter\_exact2] value measured [time\_filter\_global2] compared to the [time\_filter\_exact1] value measured [time\_filter\_global1]? \\
single & Table & What is the change in the value of \{lab\_name\} of patient \{patient\_id\} from the [time\_filter\_exact2] value measured [time\_filter\_global2] compared to the [time\_filter\_exact1] value measured [time\_filter\_global1]? \\
single & Table & What is the change in the \{vital\_name\} of patient \{patient\_id\} from the [time\_filter\_exact2] value measured [time\_filter\_global2] compared to the [time\_filter\_exact1] value measured [time\_filter\_global1]? \\
single & Table & Is the value of \{lab\_name\} of patient \{patient\_id\} [time\_filter\_exact2] measured [time\_filter\_global2] [comparison] than the [time\_filter\_exact1] value measured [time\_filter\_global1]? \\
single & Table & Is the \{vital\_name\} of patient \{patient\_id\} [time\_filter\_exact2] measured [time\_filter\_global2] [comparison] than the [time\_filter\_exact1] value measured [time\_filter\_global1]? \\
single & Table & What is\_verb the age of patient \{patient\_id\} [time\_filter\_global1]? \\
single & Table & What is\_verb the name of insurance of patient \{patient\_id\} [time\_filter\_global1]? \\
single & Table & What is\_verb the marital status of patient \{patient\_id\} [time\_filter\_global1]? \\
single & Table & What percentile is the value of \{lab\_value\} in a \{lab\_name\} lab test among patients of the same age as patient \{patient\_id\} [time\_filter\_global1]? \\
single & Table & How many [unit\_count] have passed since patient \{patient\_id\} was admitted to the hospital currently? \\
single & Table & How many [unit\_count] have passed since patient \{patient\_id\} was admitted to the ICU currently? \\
single & Table & How many [unit\_count] have passed since the [time\_filter\_exact1] time patient \{patient\_id\} was transferred to careunit \{careunit\} on the current hospital visit? \\
single & Table & How many [unit\_count] have passed since the [time\_filter\_exact1] time patient \{patient\_id\} was diagnosed with \{diagnosis\_name\} on the current hospital visit? \\
single & Table & How many [unit\_count] have passed since the [time\_filter\_exact1] time patient \{patient\_id\} was prescribed \{drug\_name\} on the current hospital visit? \\
single & Table & How many [unit\_count] have passed since the [time\_filter\_exact1] time patient \{patient\_id\} received a \{lab\_name\} lab test on the current hospital visit? \\
single & Table & How many [unit\_count] have passed since the [time\_filter\_exact1] time patient \{patient\_id\} had a \{intake\_name\} intake on the current ICU visit? \\
single & Table & What was the [time\_filter\_exact1] hospital admission type of patient \{patient\_id\} [time\_filter\_global1]? \\
single & Table & What was the [time\_filter\_exact1] careunit of patient \{patient\_id\} [time\_filter\_global1]? \\
single & Table & What was the [time\_filter\_exact1] measured height of patient \{patient\_id\} [time\_filter\_global1]? \\
single & Table & What was the [time\_filter\_exact1] measured weight of patient \{patient\_id\} [time\_filter\_global1]? \\
single & Table & What was the name of the diagnosis that patient \{patient\_id\} [time\_filter\_exact1] received [time\_filter\_global1]? \\
single & Table & What was the name of the procedure that patient \{patient\_id\} [time\_filter\_exact1] received [time\_filter\_global1]? \\
single & Table & What was the name of the drug that patient \{patient\_id\} was [time\_filter\_exact1] prescribed via \{drug\_route\} route [time\_filter\_global1]? \\
single & Table & What was the name of the drug that patient \{patient\_id\} was [time\_filter\_exact1] prescribed [time\_filter\_global1]? \\
single & Table & What was the name of the drug that patient \{patient\_id\} was prescribed [time\_filter\_within] after having been diagnosed with \{diagnosis\_name\} [time\_filter\_global1]? \\
single & Table & What was the name of the drug that patient \{patient\_id\} was prescribed [time\_filter\_within] after having received a \{procedure\_name\} procedure [time\_filter\_global1]? \\
single & Table & What was the dose of \{drug\_name\} that patient \{patient\_id\} was [time\_filter\_exact1] prescribed [time\_filter\_global1]? \\
single & Table & What was the total amount of dose of \{drug\_name\} that patient \{patient\_id\} were prescribed [time\_filter\_global1]? \\
single & Table & What was the name of the drug that patient \{patient\_id\} were prescribed [n\_times] [time\_filter\_global1]? \\
single & Table & What is the new prescription of patient \{patient\_id\} [time\_filter\_global2] compared to the prescription [time\_filter\_global1]? \\
single & Table & What was the [time\_filter\_exact1] measured value of a \{lab\_name\} lab test of patient \{patient\_id\} [time\_filter\_global1]? \\
single & Table & What was the name of the lab test that patient \{patient\_id\} [time\_filter\_exact1] received [time\_filter\_global1]? \\
single & Table & What was the [agg\_function] \{lab\_name\} value of patient \{patient\_id\} [time\_filter\_global1]? \\
single & Table & What was the organism name found in the [time\_filter\_exact1] \{culture\_name\} microbiology test of patient \{patient\_id\} [time\_filter\_global1]? \\
single & Table & What was the organism name found in the [time\_filter\_exact1] \{test\_name\} test of patient \{patient\_id\} [time\_filter\_global1]? \\
single & Table & What was the name of the specimen that patient \{patient\_id\} was [time\_filter\_exact1] tested [time\_filter\_global1]? \\
single & Table & What was the name of the microbiology test that patient \{patient\_id\} [time\_filter\_exact1] received [time\_filter\_global1]? \\
single & Table & What was the name of the intake that patient \{patient\_id\} [time\_filter\_exact1] had [time\_filter\_global1]? \\
single & Table & What was the total volume of \{intake\_name\} intake that patient \{patient\_id\} received [time\_filter\_global1]? \\
single & Table & What was the total volume of intake that patient \{patient\_id\} received [time\_filter\_global1]? \\
single & Table & What was the name of the output that patient \{patient\_id\} [time\_filter\_exact1] had [time\_filter\_global1]? \\
single & Table & What was the total volume of \{output\_name\} output that patient \{patient\_id\} had [time\_filter\_global1]? \\
single & Table & What was the total volume of output that patient \{patient\_id\} had [time\_filter\_global1]? \\
single & Table & What is the difference between the total volume of intake and output of patient \{patient\_id\} [time\_filter\_global1]? \\
single & Table & What was the [time\_filter\_exact1] measured \{vital\_name\} of patient \{patient\_id\} [time\_filter\_global1]? \\
single & Table & What was the [agg\_function] \{vital\_name\} of patient \{patient\_id\} [time\_filter\_global1]? \\
single & Table & What is\_verb the total hospital cost of patient \{patient\_id\} [time\_filter\_global1]? \\
single & Table & When was the [time\_filter\_exact1] hospital admission time of patient \{patient\_id\} [time\_filter\_global1]? \\
single & Table & When was the [time\_filter\_exact1] hospital admission time that patient \{patient\_id\} was admitted via \{admission\_route\} [time\_filter\_global1]? \\
single & Table & When was the [time\_filter\_exact1] hospital discharge time of patient \{patient\_id\} [time\_filter\_global1]? \\
single & Table & What was the [time\_filter\_exact1] length of ICU stay of patient \{patient\_id\}? \\
single & Table & When was the [time\_filter\_exact1] time that patient \{patient\_id\} was diagnosed with \{diagnosis\_name\} [time\_filter\_global1]? \\
single & Table & When was the [time\_filter\_exact1] procedure time of patient \{patient\_id\} [time\_filter\_global1]? \\
single & Table & When was the [time\_filter\_exact1] time that patient \{patient\_id\} received a \{procedure\_name\} procedure [time\_filter\_global1]? \\
single & Table & When was the [time\_filter\_exact1] prescription time of patient \{patient\_id\} [time\_filter\_global1]? \\
single & Table & When was the [time\_filter\_exact1] time that patient \{patient\_id\} was prescribed \{drug\_name\} [time\_filter\_global1]? \\
single & Table & When was the [time\_filter\_exact1] time that patient \{patient\_id\} was prescribed \{drug\_name1\} and \{drug\_name2\} [time\_filter\_within] [time\_filter\_global1]? \\
single & Table & When was the [time\_filter\_exact1] time that patient \{patient\_id\} was prescribed a medication via \{drug\_route\} route [time\_filter\_global1]? \\
single & Table & When was the [time\_filter\_exact1] lab test of patient \{patient\_id\} [time\_filter\_global1]? \\
single & Table & When was the [time\_filter\_exact1] time that patient \{patient\_id\} received a \{lab\_name\} lab test [time\_filter\_global1]? \\
single & Table & When was the [time\_filter\_exact1] time that patient \{patient\_id\} had the [sort] value of \{lab\_name\} [time\_filter\_global1]? \\
single & Table & When was the [time\_filter\_exact1] microbiology test of patient \{patient\_id\} [time\_filter\_global1]? \\
single & Table & When was patient \{patient\_id\}'s [time\_filter\_exact1] \{culture\_name\} microbiology test [time\_filter\_global1]? \\
single & Table & When was patient \{patient\_id\}'s [time\_filter\_exact1] \{test\_name\} test [time\_filter\_global1]? \\
single & Table & When was the [time\_filter\_exact1] time that patient \{patient\_id\} had a \{intake\_name\} intake [time\_filter\_global1]? \\
single & Table & When was the [time\_filter\_exact1] intake time of patient \{patient\_id\} [time\_filter\_global1]? \\
single & Table & When was the [time\_filter\_exact1] time that patient \{patient\_id\} had a \{output\_name\} output [time\_filter\_global1]? \\
single & Table & When was the [time\_filter\_exact1] time that patient \{patient\_id\} had a \{vital\_name\} measured [time\_filter\_global1]? \\
single & Table & When was the [time\_filter\_exact1] time that the \{vital\_name\} of patient \{patient\_id\} was [comparison] than \{vital\_value\} [time\_filter\_global1]? \\
single & Table & When was the [time\_filter\_exact1] time that patient \{patient\_id\} had the [sort] \{vital\_name\} [time\_filter\_global1]? \\
single & Table & Has\_verb patient \{patient\_id\} been admitted to the hospital [time\_filter\_global1]? \\
single & Table & Has\_verb patient \{patient\_id\} been to an emergency room [time\_filter\_global1]? \\
single & Table & Has\_verb patient \{patient\_id\} received any procedure [time\_filter\_global1]? \\
single & Table & Has\_verb patient \{patient\_id\} received a \{procedure\_name\} procedure [time\_filter\_global1]? \\
single & Table & What was the name of the procedure that patient \{patient\_id\} received [n\_times] [time\_filter\_global1]? \\
single & Table & Has\_verb patient \{patient\_id\} received any diagnosis [time\_filter\_global1]? \\
single & Table & Has\_verb patient \{patient\_id\} been diagnosed with \{diagnosis\_name\} [time\_filter\_global1]? \\
single & Table & Has\_verb patient \{patient\_id\} been prescribed \{drug\_name1\}, \{drug\_name2\}, or \{drug\_name3\} [time\_filter\_global1]? \\
single & Table & Has\_verb patient \{patient\_id\} been prescribed any medication [time\_filter\_global1]? \\
single & Table & Has\_verb patient \{patient\_id\} been prescribed \{drug\_name\} [time\_filter\_global1]? \\
single & Table & Has\_verb patient \{patient\_id\} received any lab test [time\_filter\_global1]? \\
single & Table & Has\_verb patient \{patient\_id\} received a \{lab\_name\} lab test [time\_filter\_global1]? \\
single & Table & Has\_verb patient \{patient\_id\} had any microbiology test result [time\_filter\_global1]? \\
single & Table & Has\_verb patient \{patient\_id\} had any \{culture\_name\} microbiology test result [time\_filter\_global1]? \\
single & Table & Has\_verb patient \{patient\_id\} had any \{test\_name\} test result [time\_filter\_global1]? \\
single & Table & Has\_verb there been any organism found in the [time\_filter\_exact1] \{culture\_name\} microbiology test of patient \{patient\_id\} [time\_filter\_global1]? \\
single & Table & Has\_verb there been any organism found in the [time\_filter\_exact1] \{test\_name\} test of patient \{patient\_id\} [time\_filter\_global1]? \\
single & Table & Has\_verb patient \{patient\_id\} had any \{intake\_name\} intake [time\_filter\_global1]? \\
single & Table & Has\_verb patient \{patient\_id\} had any \{output\_name\} output [time\_filter\_global1]? \\
single & Table & Has\_verb the \{vital\_name\} of patient \{patient\_id\} been ever [comparison] than \{vital\_value\} [time\_filter\_global1]? \\
single & Table & Has\_verb the \{vital\_name\} of patient \{patient\_id\} been normal [time\_filter\_global1]? \\
single & Table & List the hospital admission time of patient \{patient\_id\} [time\_filter\_global1]. \\
single & Table & List the [unit\_average] [agg\_function] \{lab\_name\} lab value of patient \{patient\_id\} [time\_filter\_global1]. \\
single & Table & List the [unit\_average] [agg\_function] weight of patient \{patient\_id\} [time\_filter\_global1]. \\
single & Table & List the [unit\_average] [agg\_function] volume of \{intake\_name\} intake that patient \{patient\_id\} received [time\_filter\_global1]. \\
single & Table & List the [unit\_average] [agg\_function] volume of \{output\_name\} output that patient \{patient\_id\} had [time\_filter\_global1]. \\
single & Table & List the [unit\_average] [agg\_function] \{vital\_name\} of patient \{patient\_id\} [time\_filter\_global1]. \\
single & Table & Count the number of hospital visits of patient \{patient\_id\} [time\_filter\_global1]. \\
single & Table & Count the number of ICU visits of patient \{patient\_id\} [time\_filter\_global1]. \\
single & Table & Count the number of times that patient \{patient\_id\} received a \{procedure\_name\} procedure [time\_filter\_global1]. \\
single & Table & Count the number of drugs patient \{patient\_id\} were prescribed [time\_filter\_global1]. \\
single & Table & Count the number of times that patient \{patient\_id\} were prescribed \{drug\_name\} [time\_filter\_global1]. \\
single & Table & Count the number of times that patient \{patient\_id\} received a \{lab\_name\} lab test [time\_filter\_global1]. \\
single & Table & Count the number of times that patient \{patient\_id\} had a \{intake\_name\} intake [time\_filter\_global1]. \\
single & Table & Count the number of times that patient \{patient\_id\} had a \{output\_name\} output [time\_filter\_global1]. \\
group & Table & Count the number of current patients. \\
group & Table & Count the number of current patients aged [age\_group]. \\
group & Table & What is the [n\_survival\_period] survival rate of patients diagnosed with \{diagnosis\_name\}? \\
group & Table & What is the [n\_survival\_period] survival rate of patients who were prescribed \{drug\_name\} after having been diagnosed with \{diagnosis\_name\}? \\
group & Table & What are the top [n\_rank] diagnoses that have the highest [n\_survival\_period] mortality rate? \\
group & Table & What is\_verb the [agg\_function] total hospital cost that involves a procedure named \{procedure\_name\} [time\_filter\_global1]? \\
group & Table & What is\_verb the [agg\_function] total hospital cost that involves a \{lab\_name\} lab test [time\_filter\_global1]? \\
group & Table & What is\_verb the [agg\_function] total hospital cost that involves a drug named \{drug\_name\} [time\_filter\_global1]? \\
group & Table & What is\_verb the [agg\_function] total hospital cost that involves a diagnosis named \{diagnosis\_name\} [time\_filter\_global1]? \\
group & Table & List the IDs of patients diagnosed with \{diagnosis\_name\} [time\_filter\_global1]. \\
group & Table & What is the [agg\_function] [unit\_average] number of patient records diagnosed with \{diagnosis\_name\} [time\_filter\_global1]? \\
group & Table & Count the number of patients who were dead after having been diagnosed with \{diagnosis\_name\} [time\_filter\_within] [time\_filter\_global1]. \\
group & Table & Count the number of patients who did not come back to the hospital [time\_filter\_within] after diagnosed with \{diagnosis\_name\} [time\_filter\_global1]. \\
group & Table & Count the number of patients who were admitted to the hospital [time\_filter\_global1]. \\
group & Table & Count the number of patients who were discharged from the hospital [time\_filter\_global1]. \\
group & Table & Count the number of patients who stayed in careunit \{careunit\} [time\_filter\_global1]. \\
group & Table & Count the number of patients who were diagnosed with \{diagnosis\_name\} [time\_filter\_within] after having received a \{procedure\_name\} procedure [time\_filter\_global1]. \\
group & Table & Count the number of patients who were diagnosed with \{diagnosis\_name2\} [time\_filter\_within] after having been diagnosed with \{diagnosis\_name1\} [time\_filter\_global1]. \\
group & Table & Count the number of patients who were diagnosed with \{diagnosis\_name\} [time\_filter\_global1]. \\
group & Table & Count the number of patients who received a \{procedure\_name\} procedure [time\_filter\_global1]. \\
group & Table & Count the number of patients who received a \{procedure\_name\} procedure [n\_times] [time\_filter\_global1]. \\
group & Table & Count the number of patients who received a \{procedure\_name2\} procedure [time\_filter\_within] after having received a \{procedure\_name1\} procedure [time\_filter\_global1]. \\
group & Table & Count the number of patients who received a \{procedure\_name\} procedure [time\_filter\_within] after having been diagnosed with \{diagnosis\_name\} [time\_filter\_global1]. \\
group & Table & Count the number of \{procedure\_name\} procedure cases [time\_filter\_global1]. \\
group & Table & Count the number of patients who were prescribed \{drug\_name\} [time\_filter\_global1]. \\
group & Table & Count the number of \{drug\_name\} prescription cases [time\_filter\_global1]. \\
group & Table & Count the number of patients who were prescribed \{drug\_name\} [time\_filter\_within] after having received a \{procedure\_name\} procedure [time\_filter\_global1]. \\
group & Table & Count the number of patients who were prescribed \{drug\_name\} [time\_filter\_within] after having been diagnosed with \{diagnosis\_name\} [time\_filter\_global1]. \\
group & Table & Count the number of patients who received a \{lab\_name\} lab test [time\_filter\_global1]. \\
group & Table & Count the number of patients who received a \{culture\_name\} microbiology test [time\_filter\_global1]. \\
group & Table & Count the number of patients who received a \{test\_name\} test [time\_filter\_global1]. \\
group & Table & Count the number of patients who had a \{intake\_name\} intake [time\_filter\_global1]. \\
group & Table & What are\_verb the top [n\_rank] frequent diagnoses [time\_filter\_global1]? \\
group & Table & What are\_verb the top [n\_rank] frequent diagnoses of patients aged [age\_group] [time\_filter\_global1]? \\
group & Table & What are\_verb the top [n\_rank] frequent diagnoses that patients were diagnosed [time\_filter\_within] after having received a \{procedure\_name\} procedure [time\_filter\_global1]? \\
group & Table & What are\_verb the top [n\_rank] frequent diagnoses that patients were diagnosed [time\_filter\_within] after having been diagnosed with \{diagnosis\_name\} [time\_filter\_global1]? \\
group & Table & What are\_verb the top [n\_rank] frequent procedures [time\_filter\_global1]? \\
group & Table & What are\_verb the top [n\_rank] frequent procedures of patients aged [age\_group] [time\_filter\_global1]? \\
group & Table & What are\_verb the top [n\_rank] frequent procedures that patients received [time\_filter\_within] after having received a \{procedure\_name\} procedure [time\_filter\_global1]? \\
group & Table & What are\_verb the top [n\_rank] frequent procedures that patients received [time\_filter\_within] after having been diagnosed with \{diagnosis\_name\} [time\_filter\_global1]? \\
group & Table & What are\_verb the top [n\_rank] frequently prescribed drugs [time\_filter\_global1]? \\
group & Table & What are\_verb the top [n\_rank] frequently prescribed drugs of patients aged [age\_group] [time\_filter\_global1]? \\
group & Table & What are\_verb the top [n\_rank] frequent prescribed drugs for patients who were also prescribed \{drug\_name\} at the same time [time\_filter\_global1]? \\
group & Table & What are\_verb the top [n\_rank] frequent drugs that patients were prescribed [time\_filter\_within] after having been prescribed with \{drug\_name\} [time\_filter\_global1]? \\
group & Table & What are\_verb the top [n\_rank] frequently prescribed drugs that patients were prescribed [time\_filter\_within] after having received a \{procedure\_name\} procedure [time\_filter\_global1]? \\
group & Table & What are\_verb the top [n\_rank] frequently prescribed drugs that patients were prescribed [time\_filter\_within] after having been diagnosed with \{diagnosis\_name\} [time\_filter\_global1]? \\
group & Table & What are\_verb the top [n\_rank] frequently prescribed drugs that patients aged [age\_group] were prescribed [time\_filter\_within] after having been diagnosed with \{diagnosis\_name\} [time\_filter\_global1]? \\
group & Table & What are\_verb the top [n\_rank] frequently prescribed drugs that \{gender\} patients aged [age\_group] were prescribed [time\_filter\_within] after having been diagnosed with \{diagnosis\_name\} [time\_filter\_global1]? \\
group & Table & What are\_verb the top [n\_rank] frequent lab tests [time\_filter\_global1]? \\
group & Table & What are\_verb the top [n\_rank] frequent lab tests of patients aged [age\_group] [time\_filter\_global1]? \\
group & Table & What are\_verb the top [n\_rank] frequent lab tests that patients had [time\_filter\_within] after having been diagnosed with \{diagnosis\_name\} [time\_filter\_global1]? \\
group & Table & What are\_verb the top [n\_rank] frequent lab tests that patients had [time\_filter\_within] after having received a \{procedure\_name\} procedure [time\_filter\_global1]? \\
group & Table & What are\_verb the top [n\_rank] frequent specimens tested [time\_filter\_global1]? \\
group & Table & What are\_verb the top [n\_rank] frequent microbiology tests [time\_filter\_global1]? \\
group & Table & What are\_verb the top [n\_rank] frequent specimens that patients were tested [time\_filter\_within] after having been diagnosed with \{diagnosis\_name\} [time\_filter\_global1]? \\
group & Table & What are\_verb the top [n\_rank] frequent microbiology tests that patients had [time\_filter\_within] after having been diagnosed with \{diagnosis\_name\} [time\_filter\_global1]? \\
group & Table & What are\_verb the top [n\_rank] frequent specimens that patients were tested [time\_filter\_within] after having received a \{procedure\_name\} procedure [time\_filter\_global1]? \\
group & Table & What are\_verb the top [n\_rank] frequent microbiology tests that patients had [time\_filter\_within] after having received a \{procedure\_name\} procedure [time\_filter\_global1]? \\
group & Table & What are\_verb the top [n\_rank] frequent intake events [time\_filter\_global1]? \\
group & Table & What are\_verb the top [n\_rank] frequent output events [time\_filter\_global1]? \\
\end{longtblr}

\begin{longtblr}
[
  caption = {Full list of 75 \textit{Image+Table}-related question templates in EHRXQA},
  label = {tab:mm_template_full},
]
{
  colspec = {X[1.4,c,m]X[1.4,c,m]X[10,l,m]},
  colsep = 0.3pt,
  rowhead = 1,
  hlines,
  rows={font=\tiny},
  rowsep=0.3pt,
}
\textbf{Patient scope} & \textbf{Modality scope} & {\SetCell[c=1]{c}\textbf{Question Template}} \\
Single & Image + Table  & Has\_verb patient \{patient\_id\} been diagnosed with \{diagnosis\_name\} [time\_filter\_global1] and also had a chest X-ray study indicating \$\{attribute\} in the \$\{object\} within the same period? \\
Single & Image + Table  & Has\_verb patient \{patient\_id\} been diagnosed with \{diagnosis\_name\} [time\_filter\_global1] and also had a chest X-ray study indicating any \$\{category\} in the \$\{object\} within the same period? \\
Single & Image + Table  & Has\_verb patient \{patient\_id\} been diagnosed with \{diagnosis\_name\} [time\_filter\_global1] and also had a chest X-ray study indicating any abnormality in the \$\{object\} within the same period? \\
Single & Image + Table  & Has\_verb patient \{patient\_id\} been diagnosed with \{diagnosis\_name\} [time\_filter\_global1] and also had a chest X-ray study indicating \$\{attribute\} within the same period? \\
Single & Image + Table  & Has\_verb patient \{patient\_id\} been diagnosed with \{diagnosis\_name\} [time\_filter\_global1] and also had a chest X-ray study indicating any \$\{category\} within the same period? \\
Single & Image + Table  & Has\_verb patient \{patient\_id\} been diagnosed with \{diagnosis\_name\} [time\_filter\_global1] and also had a chest X-ray study indicating any abnormality within the same period? \\
Single & Image + Table  & Has\_verb patient \{patient\_id\} received a \{procedure\_name\} procedure [time\_filter\_global1] and also had a chest X-ray study indicating \$\{attribute\} in the \$\{object\} within the same period? \\
Single & Image + Table  & Has\_verb patient \{patient\_id\} received a \{procedure\_name\} procedure [time\_filter\_global1] and also had a chest X-ray study indicating any \$\{category\} in the \$\{object\} within the same period? \\
Single & Image + Table  & Has\_verb patient \{patient\_id\} received a \{procedure\_name\} procedure [time\_filter\_global1] and also had a chest X-ray study indicating any abnormality in the \$\{object\} within the same period? \\
Single & Image + Table  & Has\_verb patient \{patient\_id\} received a \{procedure\_name\} procedure [time\_filter\_global1] and also had a chest X-ray study indicating \$\{attribute\} within the same period? \\
Single & Image + Table  & Has\_verb patient \{patient\_id\} received a \{procedure\_name\} procedure [time\_filter\_global1] and also had a chest X-ray study indicating any \$\{category\} within the same period? \\
Single & Image + Table  & Has\_verb patient \{patient\_id\} received a \{procedure\_name\} procedure [time\_filter\_global1] and also had a chest X-ray study indicating any abnormality within the same period? \\
Single & Image + Table  & Has\_verb patient \{patient\_id\} been prescribed with \{drug\_name\} [time\_filter\_global1] and also had a chest X-ray study indicating \$\{attribute\} in the \$\{object\} within the same period? \\
Single & Image + Table  & Has\_verb patient \{patient\_id\} been prescribed with \{drug\_name\} [time\_filter\_global1] and also had a chest X-ray study indicating any \$\{category\} in the \$\{object\} within the same period? \\
Single & Image + Table  & Has\_verb patient \{patient\_id\} been prescribed with \{drug\_name\} [time\_filter\_global1] and also had a chest X-ray study indicating any abnormality in the \$\{object\} within the same period? \\
Single & Image + Table  & Has\_verb patient \{patient\_id\} been prescribed with \{drug\_name\} [time\_filter\_global1] and also had a chest X-ray study indicating \$\{attribute\} within the same period? \\
Single & Image + Table  & Has\_verb patient \{patient\_id\} been prescribed with \{drug\_name\} [time\_filter\_global1] and also had a chest X-ray study indicating any \$\{category\} within the same period? \\
Single & Image + Table  & Has\_verb patient \{patient\_id\} been prescribed with \{drug\_name\} [time\_filter\_global1] and also had a chest X-ray study indicating any abnormality within the same period? \\
Single & Image + Table  & Has\_verb patient \{patient\_id\} received any diagnosis [time\_filter\_global1] and also had a chest X-ray study indicating \$\{attribute\} in the \$\{object\} within the same period? \\
Single & Image + Table  & Has\_verb patient \{patient\_id\} received any diagnosis [time\_filter\_global1] and also had a chest X-ray study indicating any \$\{category\} in the \$\{object\} within the same period? \\
Single & Image + Table  & Has\_verb patient \{patient\_id\} received any diagnosis [time\_filter\_global1] and also had a chest X-ray study indicating \$\{attribute\} within the same period? \\
Single & Image + Table  & Has\_verb patient \{patient\_id\} received any procedure [time\_filter\_global1] and also had a chest X-ray study indicating \$\{attribute\} in the \$\{object\} within the same period? \\
Single & Image + Table  & Has\_verb patient \{patient\_id\} received any procedure [time\_filter\_global1] and also had a chest X-ray study indicating any \$\{category\} in the \$\{object\} within the same period? \\
Single & Image + Table  & Has\_verb patient \{patient\_id\} received any procedure [time\_filter\_global1] and also had a chest X-ray study indicating \$\{attribute\} within the same period? \\
Single & Image + Table  & Has\_verb patient \{patient\_id\} been prescribed any medication [time\_filter\_global1] and also had a chest X-ray study indicating \$\{attribute\} in the \$\{object\} within the same period? \\
Single & Image + Table  & Has\_verb patient \{patient\_id\} been prescribed any medication [time\_filter\_global1] and also had a chest X-ray study indicating any \$\{category\} in the \$\{object\} within the same period? \\
Single & Image + Table  & Has\_verb patient \{patient\_id\} been prescribed any medication [time\_filter\_global1] and also had a chest X-ray study indicating \$\{attribute\} within the same period? \\
Single & Image + Table  & Has\_verb patient \{patient\_id\} had a chest X-ray study indicating \$\{attribute\} in the \$\{object\} [time\_filter\_within] after having been diagnosed with \{diagnosis\_name\} [time\_filter\_global1]? \\
Single & Image + Table  & Has\_verb patient \{patient\_id\} had a chest X-ray study indicating any \$\{category\} in the \$\{object\} [time\_filter\_within] after having been diagnosed with \{diagnosis\_name\} [time\_filter\_global1]? \\
Single & Image + Table  & Has\_verb patient \{patient\_id\} had a chest X-ray study indicating any abnormality in the \$\{object\} [time\_filter\_within] after having been diagnosed with \{diagnosis\_name\} [time\_filter\_global1]? \\
Single & Image + Table  & Has\_verb patient \{patient\_id\} had a chest X-ray study indicating \$\{attribute\} [time\_filter\_within] after having been diagnosed with \{diagnosis\_name\} [time\_filter\_global1]? \\
Single & Image + Table  & Has\_verb patient \{patient\_id\} had a chest X-ray study indicating any \$\{category\} [time\_filter\_within] after having been diagnosed with \{diagnosis\_name\} [time\_filter\_global1]? \\
Single & Image + Table  & Has\_verb patient \{patient\_id\} had a chest X-ray study indicating any abnormality [time\_filter\_within] after having been diagnosed with \{diagnosis\_name\} [time\_filter\_global1]? \\
Single & Image + Table  & Has\_verb patient \{patient\_id\} had a chest X-ray study indicating \$\{attribute\} in the \$\{object\} [time\_filter\_within] after having received \{procedure\_name\} procedure [time\_filter\_global1] ? \\
Single & Image + Table  & Has\_verb patient \{patient\_id\} had a chest X-ray study indicating any \$\{category\} in the \$\{object\} [time\_filter\_within] after having received \{procedure\_name\} procedure [time\_filter\_global1] ? \\
Single & Image + Table  & Has\_verb patient \{patient\_id\} had a chest X-ray study indicating any abnormality in the \$\{object\} [time\_filter\_within] after having received \{procedure\_name\} procedure [time\_filter\_global1] ? \\
Single & Image + Table  & Has\_verb patient \{patient\_id\} had a chest X-ray study indicating \$\{attribute\} [time\_filter\_within] after having received \{procedure\_name\} procedure [time\_filter\_global1] ? \\
Single & Image + Table  & Has\_verb patient \{patient\_id\} had a chest X-ray study indicating any \$\{category\} [time\_filter\_within] after having received \{procedure\_name\} procedure [time\_filter\_global1] ? \\
Single & Image + Table  & Has\_verb patient \{patient\_id\} had a chest X-ray study indicating any abnormality [time\_filter\_within] after having received \{procedure\_name\} procedure [time\_filter\_global1] ? \\
Single & Image + Table  & Has\_verb patient \{patient\_id\} had a chest X-ray study indicating \$\{attribute\} in the \$\{object\} [time\_filter\_within] after having been prescribed with \{drug\_name\} [time\_filter\_global1] ? \\
Single & Image + Table  & Has\_verb patient \{patient\_id\} had a chest X-ray study indicating any \$\{category\} in the \$\{object\} [time\_filter\_within] after having been prescribed with \{drug\_name\} [time\_filter\_global1] ? \\
Single & Image + Table  & Has\_verb patient \{patient\_id\} had a chest X-ray study indicating any abnormality in the \$\{object\} [time\_filter\_within] after having been prescribed with \{drug\_name\} [time\_filter\_global1] ? \\
Single & Image + Table  & Has\_verb patient \{patient\_id\} had a chest X-ray study indicating \$\{attribute\} [time\_filter\_within] after having been prescribed with \{drug\_name\} [time\_filter\_global1] ? \\
Single & Image + Table  & Has\_verb patient \{patient\_id\} had a chest X-ray study indicating any \$\{category\} [time\_filter\_within] after having been prescribed with \{drug\_name\} [time\_filter\_global1] ? \\
Single & Image + Table  & Has\_verb patient \{patient\_id\} had a chest X-ray study indicating any abnormality [time\_filter\_within] after having been prescribed with \{drug\_name\} [time\_filter\_global1] ? \\
Single & Image + Table  & Has\_verb patient \{patient\_id\} been diagnosed with \{diagnosis\_name\} [time\_filter\_within] after having had a chest X-ray study indicating \$\{attribute\} in the \$\{object\} [time\_filter\_global1] ? \\
Single & Image + Table  & Has\_verb patient \{patient\_id\} been diagnosed with \{diagnosis\_name\} [time\_filter\_within] after having had a chest X-ray study indicating any \$\{category\} in the \$\{object\} [time\_filter\_global1] ? \\
Single & Image + Table  & Has\_verb patient \{patient\_id\} been diagnosed with \{diagnosis\_name\} [time\_filter\_within] after having had a chest X-ray study indicating any abnormality in the \$\{object\} [time\_filter\_global1] ? \\
Single & Image + Table  & Has\_verb patient \{patient\_id\} been diagnosed with \{diagnosis\_name\} [time\_filter\_within] after having had a chest X-ray study indicating \$\{attribute\} [time\_filter\_global1] ? \\
Single & Image + Table  & Has\_verb patient \{patient\_id\} been diagnosed with \{diagnosis\_name\} [time\_filter\_within] after having had a chest X-ray study indicating any \$\{category\} [time\_filter\_global1] ? \\
Single & Image + Table  & Has\_verb patient \{patient\_id\} been diagnosed with \{diagnosis\_name\} [time\_filter\_within] after having had a chest X-ray study indicating any abnormality [time\_filter\_global1] ? \\
Single & Image + Table  & Has\_verb patient \{patient\_id\} received a \{procedure\_name\} procedure [time\_filter\_within] after having had a chest X-ray study indicating \$\{attribute\} in the \$\{object\} [time\_filter\_global1] ? \\
Single & Image + Table  & Has\_verb patient \{patient\_id\} received a \{procedure\_name\} procedure [time\_filter\_within] after having had a chest X-ray study indicating any \$\{category\} in the \$\{object\} [time\_filter\_global1] ? \\
Single & Image + Table  & Has\_verb patient \{patient\_id\} received a \{procedure\_name\} procedure [time\_filter\_within] after having had a chest X-ray study indicating any abnormality in the \$\{object\} [time\_filter\_global1] ? \\
Single & Image + Table  & Has\_verb patient \{patient\_id\} received a \{procedure\_name\} procedure [time\_filter\_within] after having had a chest X-ray study indicating \$\{attribute\} [time\_filter\_global1] ? \\
Single & Image + Table  & Has\_verb patient \{patient\_id\} received a \{procedure\_name\} procedure [time\_filter\_within] after having had a chest X-ray study indicating any \$\{category\} [time\_filter\_global1] ? \\
Single & Image + Table  & Has\_verb patient \{patient\_id\} received a \{procedure\_name\} procedure [time\_filter\_within] after having had a chest X-ray study indicating any abnormality [time\_filter\_global1] ? \\
Single & Image + Table  & Has\_verb patient \{patient\_id\} been prescribed with \{drug\_name\} [time\_filter\_within] after having had a chest X-ray study indicating \$\{attribute\} in the \$\{object\} [time\_filter\_global1] ? \\
Single & Image + Table  & Has\_verb patient \{patient\_id\} been prescribed with \{drug\_name\} [time\_filter\_within] after having had a chest X-ray study indicating any \$\{category\} in the \$\{object\} [time\_filter\_global1] ? \\
Single & Image + Table  & Has\_verb patient \{patient\_id\} been prescribed with \{drug\_name\} [time\_filter\_within] after having had a chest X-ray study indicating any abnormality in the \$\{object\} [time\_filter\_global1] ? \\
Single & Image + Table  & Has\_verb patient \{patient\_id\} been prescribed with \{drug\_name\} [time\_filter\_within] after having had a chest X-ray study indicating \$\{attribute\} [time\_filter\_global1] ? \\
Single & Image + Table  & Has\_verb patient \{patient\_id\} been prescribed with \{drug\_name\} [time\_filter\_within] after having had a chest X-ray study indicating any \$\{category\} [time\_filter\_global1] ? \\
Single & Image + Table  & Has\_verb patient \{patient\_id\} been prescribed with \{drug\_name\} [time\_filter\_within] after having had a chest X-ray study indicating any abnormality [time\_filter\_global1] ? \\
Group & Image + Table  & Count the number of patients aged [age\_group] who had a chest X-ray study during hospital visit indicating \$\{attribute\} in the \$\{object\} [time\_filter\_global1]. \\
Group & Image + Table  & Count the number of patients aged [age\_group] who had a chest X-ray study during hospital visit indicating any \$\{category\} in the \$\{object\} [time\_filter\_global1]. \\
Group & Image + Table  & Count the number of patients aged [age\_group] who had a chest X-ray study during hospital visit indicating \$\{attribute\} [time\_filter\_global1]. \\
Group & Image + Table  & List the IDs of patients aged [age\_group] who had a chest X-ray study during hospital visit indicating \$\{attribute\} in the \$\{object\} [time\_filter\_global1]. \\
Group & Image + Table  & List the IDs of patients aged [age\_group] who had a chest X-ray study during hospital visit indicating any \$\{category\} in the \$\{object\} [time\_filter\_global1]. \\
Group & Image + Table  & List the IDs of patients aged [age\_group] who had a chest X-ray study during hospital visit indicating \$\{attribute\} [time\_filter\_global1]. \\
Group & Image + Table  & Count the number of \{gender\} patients aged [age\_group] who had a chest X-ray study during hospital visit indicating \$\{attribute\} in the \$\{object\} [time\_filter\_global1]. \\
Group & Image + Table  & Count the number of \{gender\} patients aged [age\_group] who had a chest X-ray study during hospital visit indicating any \$\{category\} in the \$\{object\} [time\_filter\_global1]. \\
Group & Image + Table  & Count the number of \{gender\} patients aged [age\_group] who had a chest X-ray study during hospital visit indicating \$\{attribute\} [time\_filter\_global1]. \\
Group & Image + Table  & List the IDs of \{gender\} patients aged [age\_group] who had a chest X-ray study during hospital visit indicating \$\{attribute\} in the \$\{object\} [time\_filter\_global1]. \\
Group & Image + Table  & List the IDs of \{gender\} patients aged [age\_group] who had a chest X-ray study during hospital visit indicating any \$\{category\} in the \$\{object\} [time\_filter\_global1]. \\
Group & Image + Table  & List the IDs of \{gender\} patients aged [age\_group] who had a chest X-ray study during hospital visit indicating \$\{attribute\} [time\_filter\_global1]. \\
\end{longtblr}

\newpage
\subsection{QA dataset generation} \label{supp_EHRXQA_generation}
\subsubsection{SQL/NeuralSQL annotation and QA pairs sampling}
During the construction of our EHRXQA dataset, we sample \textit{table}-related QA pairs by drawing from (Question, SQL) pairs and executing the SQL query to retrieve the answer. 
However, the process for \textit{image}-related or \textit{image+table}-related QA pairs is more complex, as we cannot sample QA pairs from (Question, SQL) without the label information for images.
To overcome this, we create a new temporary table, \texttt{TB\_CXR\_PLUS}, which stores the label information for CXR images. 
The \texttt{TB\_CXR\_PLUS} table includes all the columns of the \texttt{TB\_CXR} table, as well as additional columns that represent the 563 relationships between objects and attributes, as pre-processed in the MIMIC-CXR-VQA dataset.
This table aids in annotating SQL queries to retrieve image information, effectively serving as an `answer sheet' for QA dataset generation process. 
It is important to note that this temporary table is only used during data construction.

\begin{figure}[H]
    \centering
    \includegraphics[width=0.7\columnwidth]{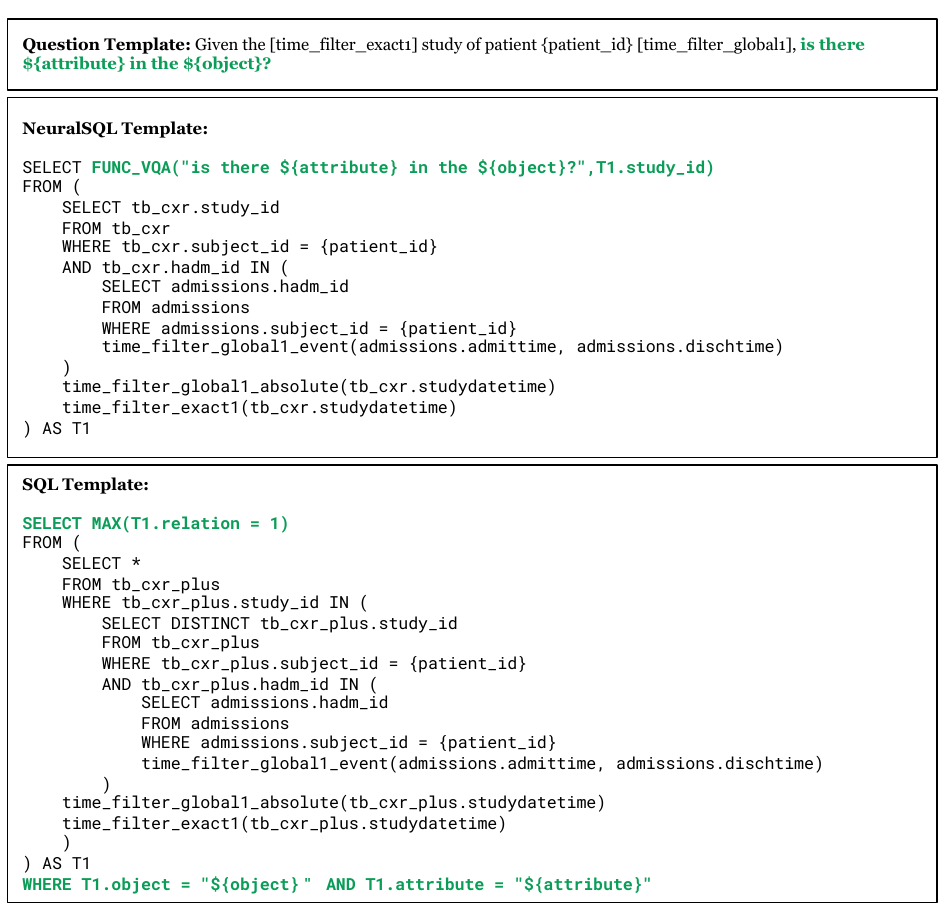}
    \caption{Comparison of NeuralSQL and SQL templates given a \textit{image}-related question template.
    Components highlighted in green indicate the same semantic meaning across the question, SQL, and NeuralSQL templates.
    }
    \label{fig:appendix_example_nsqlsql}
\end{figure}

In keeping with our goal of retrieving rich information directly from the images themselves, we employ our new approach, NeuralSQL. 
As part of this, we use \texttt{TB\_CXR\_PLUS} to annotate SQL queries and \texttt{TB\_CXR} to annotate NeuralSQL queries for all question templates related to \textit{image} and \textit{image+table}.
An example of two queries can be seen in \cref{fig:appendix_example_nsqlsql}.
To sum up,
\begin{itemize}[itemsep=1pt, topsep=-2pt, leftmargin=7mm]
    \item 
    For \textit{table}-related question templates, we annotate the corresponding SQL query. 
    During the dataset sampling process, we use these SQL queries to derive the answers. 
    The final format of this part of the dataset is (Question, SQL, Answer).
    \item 
    For \textit{image}-related or \textit{image+table}-related question templates, we annotate the corresponding SQL query using \texttt{TB\_CXR\_PLUS} and the NeuralSQL query using \texttt{TB\_CXR}. 
    During the dataset sampling process, we use the \texttt{TB\_CXR\_PLUS} query as an `answer sheet' to dervie answers, and the NeuralSQL query to formulate questions directly over the image. 
    The final format of this part of our dataset is (Question, NeuralSQL, Answer).
\end{itemize}

\newpage
\subsubsection{NeuralSQL annotation details}
\begin{itemize}[itemsep=1pt, topsep=-2pt, leftmargin=7mm]
    \item
    Four individuals, experienced in SQL and familiar with the EHRSQL dataset and its schema, participated in the SQL and NeuralSQL annotations. They were organized into teams to review each other's work.
    \item
    For NeuralSQL, if the sentence to be included in \texttt{FUNC\_VQA} involved logical or set operations, we assigned the responsibility for these operations to SQL.
    \item
    In NeuralSQL, we aimed to maintain the natural language style of the original query in the VQA sentence to be included in \texttt{FUNC\_VQA}.
\end{itemize}

\subsubsection{QA dataset split} \label{supp_ehrxqa_split}
We derive our train QA (\ie, from silver database) and testing QA (\ie, from gold database) sets from separate databases, with up to 80 samples for training and 10 for testing per template. 
The train QA set is divided further into a training QA set and a validation QA set, following a 7:1 ratio. 
This partitioning results in a validation set of a size comparable to that of the testing set.

\subsubsection{Paraphrasing}
To generate paraphrases, we leveraged the OpenAI UI, applying \cref{prompt:ehrxqa_prompt_para} to create 15 paraphrases per template using the GPT-4 model (version May 24, 2023).
Following this, human reviewers pruned any paraphrases that strayed from the initial template's meaning. 
For the \textit{Image}-related and \textit{Image+Table}-related question templates, we used paraphrases crafted with GPT-4. 
In contrast, for the \textit{Table}-related templates, we adopted machine paraphrases provided by EHRSQL. 
On average, each \textit{Table}-related template contained 47.7 paraphrases, while each \textit{Image} or \textit{Image+Table}-related template contained 10.4 paraphrases.
We randomly selected from these paraphrase pools and incorporated them into our datasets.

\newpage
\begin{figure}[H]
\centering
\scalebox{0.90}{
\input{supp_prompts/ehrxqa_prompt_para}
}
\caption{Prompt Template for Paraphrasing Question Templates for EHRXQA. Elements enclosed within double braces \{\{\}\} are substituted with values specific to each template.}
\label{prompt:ehrxqa_prompt_para}
\end{figure}

\newpage
\subsection{Data statistics} \label{supp_data_analysis}
\cref{tab:statistic_ehrxqa_details} presents detailed statistics of the EHRXQA dataset, providing a comprehensive breakdown of sample counts across various modality and patient categories in the training, validation, and test sets.
\begin{table}[H]
\caption{Detailed Statistics of EHRXQA}
\label{tab:statistic_ehrxqa_details}
\centering
\renewcommand{\arraystretch}{1.0}
\begin{adjustbox}{width=0.6\columnwidth,center}
\begin{tabular}{cccccc}
\toprule
\multirow{2.5}{*}{modality-based} &  \multicolumn{2}{c}{\multirow{2.5}{*}{patient-based}} & \multicolumn{3}{c}{\# of samples} \\ \cmidrule(lr){4-6}
                                &                            &        & train set                   & valid set                   & test set \\ \midrule
\multirow{5.5}{*}{Image}        & \multirow{3.5}{*}{single} & 1-image & 6,615 (18.3\%)              & \enskip 945 (18.3\%)        & \enskip 840 (17.5\%)        \\ \cmidrule(lr){3-6} 
                               &                            & 2-image & 3,410 \enskip(9.4\%)        & \enskip 488 \enskip (9.4\%) & \enskip 468 \enskip (9.7\%) \\ \cmidrule(lr){3-6} 
                               &                            & N-image & 1,890 \enskip(5.2\%)        & \enskip 279 \enskip (5.2\%) & \enskip 240 \enskip (5.0\%) \\ \cmidrule(lr){2-6} 
                               & \multicolumn{2}{c}{group}            & \enskip 945 \enskip(2.6\%)  & \enskip 135 \enskip (2.6\%) & \enskip 120 \enskip (2.5\%) \\ \cmidrule(lr){1-6} 
\multirow{4}{*}{Table}         & \multicolumn{2}{c}{none}             & \enskip 396 \enskip(1.1\%)  & \enskip 54 \enskip (1.0\%)  & \enskip 50 \enskip (1.0\%)  \\ \cmidrule(lr){2-6} 
                               & \multicolumn{2}{c}{single}           & 8,219 (22.7\%)              & 1,151 (22.3\%)              & 1,080 (22.5\%)              \\ \cmidrule(lr){2-6} 
                               & \multicolumn{2}{c}{group}            & 4,346 (12.0\%)              & \enskip 647 (12.5\%)        & \enskip 586 (12.2\%)        \\ \cmidrule(lr){1-6} 
\multirow{2.5}{*}{Image + Table} & \multicolumn{2}{c}{single}         & 9,517 (26.3\%)              & 1,362 (26.3\%)              & 1,210 (25.2\%)              \\ \cmidrule(lr){2-6} 
                               & \multicolumn{2}{c}{group}            & \enskip 836 \enskip(2.3\%) & \enskip 118 \enskip (2.3\%)  & \enskip 214 \enskip (4.5\%) \\ \bottomrule 
\end{tabular}
\end{adjustbox}
\end{table}

\section{NeuralSQL with visual question answering}\label{supp_nsql}
To ensure compatibility with the original SQL grammar, we extend the production rules of the SQL query language for our API call \texttt{FUNC\_VQA}. 
We achieve this by using the sqlgot\footnote{\url{https://github.com/tobymao/sqlglot}} parser in the NeuralSQL interpreter. The sqlgot parser is designed to handle a wide range of SQL inputs and generate syntactically correct SQL in the targeted dialects.
In our implementation, we perform batch-wise inference for the external VQA API to handle questions that involve multiple CXR images. 
This allows us to effectively process queries such as ``Count the number of patients who had a chest X-ray study indicating...''.

\newpage
\section{Experiments}\label{supp_exp}
\subsection{MIMIC-CXR-VQA}\label{supp_exp_vqa}
\subsubsection{MIMIC-CXR-VQA: Experimental settings}
\paragraph{\CheckedBox\;\;Standardization and Pre-processing of Pre-training Corpus}\label{sec:vqapretraining}
To ensure fair comparisons, we standardize the pre-training corpus across all VLP models. Our strategy mitigates biases from varied data sources and ensures performance differences are attributed to the models' architectural characteristics and fine-tuning efforts rather than discrepancies in pre-training data. 
We adopt MedViLL's pre-processing strategy for both image and text data from MIMIC-CXR-JPG. For X-ray images, we remove the marginal space, adjust the resolution to fit the model's input size, and maintain the aspect ratio, discarding outliers not within 0.8 and 1.2. In processing the text data, we extract~\footnote{{\url{https://github.com/MIT-LCP/mimic-cxr/tree/master/txt}}} the `\textit{findings}' and `\textit{impressions}' sections from the reports and then concatenate these sections for our models. If the token count, as tokenized by BERT, exceeds 253, we opt for the longer of the two sections. By adhering to the original MIMIC-CXR splits, we compile a corpus comprising 156,260 image-text pairs for the training set and 1,276 pairs for validation. We present results for both the original and controlled models, with the latter—pretrained on a carefully curated MIMIC-CXR corpus—denoted by an asterisk ($\ast$).

\paragraph{\CheckedBox\;\;Implementation details of VQA baselines}
\begin{itemize}[itemsep=1pt, topsep=-2pt, leftmargin=7mm]
    \item Prior (Most): 
    This model is a prior model that outputs the most popular answer in the training and validation set, which is ``yes''.
    \item Prior (Question): 
    This model is an advanced prior model that outputs the most popular answer in the training and validation set for each question (template).
    \item PubMedCLIP: We follow the original implementation code~\footnote{\url{https://github.com/sarahESL/PubMedCLIP}}.
    \item MedViLL: We follow the original implementation code~\footnote{\url{https://github.com/SuperSupermoon/MedViLL}}.
    \item M$^3$AE: 
    We follow the original implementation code~\footnote{\url{https://github.com/zhjohnchan/M3AE}}.
\end{itemize}

\begin{table}[H]
    \caption{
    Training, model configurations for VQA baselines along with resource information. 
    Some model configurations are not reported if not applicable. 
    For any other configurations that are not reported here, we followed the original paper.
    }
    \label{tab:sup_qa_configs}
    \centering
    \resizebox{\textwidth}{!}{
    \begin{tabular}{cccccc}
        \toprule
        Name & M$^3$AE$^\ast$ & M$^3$AE & MedViLL$^\ast$ & PubMedCLIP$^\ast$ & PubMedCLIP \\
        \midrule
        \multicolumn{6}{l}{\textbf{\textit{Model configurations}}} \\
        Visual encoder & ViT-B/16 & ViT-B/16 & RN50 & RN50 & RN50 \\
        Text encoder & RoBERTa$_{base}$ & RoBERTa$_{base}$ & BERT$_{base}$ & Transformer / GRU & Transformer / GRU \\
        \midrule
        \multicolumn{6}{l}{\textbf{\textit{Pre-training configurations}}} \\
        Training epoch & $50$ & $50$ & $50$ & $100$ & $100$ \\
        Batch size & $256$ & $256$ & $128$ & $64$ & $64$ \\
        Learning rate & $5$e-$5$ & $5$e-$5$ & $1$e-$5$ & $1$e-$5$ & $1$e-$5$ \\
        \midrule
        \multicolumn{6}{l}{\textbf{\textit{Finetuninig configurations}}} \\
        Training epoch & $50$ & $50$ & $20$ & $20$ & $20$ \\
        Batch size & $64$ & $64$ & $32$ & $16$ & $16$ \\
        Learning rate & $5$e-$6$ & $5$e-$6$ & $3$e-$5$ & $1$e-$3$ & $1$e-$3$ \\
        \midrule
        \multicolumn{6}{l}{\textbf{\textit{Resources} (pretraining / finetuning)} } \\
        GPU device & A6000 $\times$ $4$ / $1$ & A6000 $\times$ $4$ / $1$ & A6000 $\times$ $4$ / $1$ & A6000 $\times$ $1$ / $1$ & A6000 $\times$ $1$ / $1$ \\
        Training time & $35$ / $16$ hours & $69$ / $16$ hours & $32$ / $66$ hours & $226$ / $62$ hours &  $226$ / $62$ hours \\
        \midrule
    \end{tabular}
    }
\end{table}

\newpage
\subsubsection{MIMIC-CXR-VQA: Experimental results}\;\;
\begin{table}[H]
    \setlength{\tabcolsep}{3pt}
    \noindent
    \centering
        \captionsetup{font={small}}
        \caption{Comparison of performance of models on \textbf{MIMIC-CXR-VQA}.
        }
        \resizebox{0.4\textwidth}{!}{%
            \begin{tabular}{@{}cccccc@{}}
            \toprule
            \multirow{2}{*}{Model} & \multicolumn{2}{c}{Valid} & \multicolumn{2}{c}{Test} \\ \cmidrule(l){2-5}
                                   & Acc & F1 (micro) & Acc & F1 (micro)
                                   \\ \midrule
            Prior (Most)   
            & 26.8 & 0.27 & 25.4 & 0.25 \\
            Prior (Question)
            & 34.3 & 0.34 & 32.4 & 0.32 \\
            PubMedCLIP
            & $55.1_{\text{ ± } 1.7}$ & $0.56_{\text{ ± } 0.02}$ & $54.9_{\text{ ± } 1.3}$ & $0.54_{\text{ ± } 0.02}$ \\
            PubMedCLIP$^{\ast}$
            & $56.6_{\text{ ± } 1.9}$ & $0.58_{\text{ ± } 0.02}$ & $56.5_{\text{ ± } 2.1}$ & $0.56_{\text{ ± } 0.02}$ \\
            MedViLL$^{\ast}$
            & $64.7_{\text{ ± } 0.2}$ & $0.69_{\text{ ± } 0.00}$ & $63.6_{\text{ ± } 0.1}$ & $0.67_{\text{ ± } 0.00}$ \\
            M$^{3}$AE
            & $68.9_{\text{ ± } 0.2}$ & $0.73_{\text{ ± } 0.00}$ & $68.9_{\text{ ± } 0.3}$ & $0.72_{\text{ ± } 0.00}$ \\
            M$^{3}$AE$^{\ast}$
            & $70.2_{\text{ ± } 0.1}$ & $0.74_{\text{ ± } 0.00}$ & $69.2_{\text{ ± } 0.4}$ & $0.73_{\text{ ± } 0.00}$ \\
            \bottomrule
            \end{tabular}%
             }
  \end{table}

\begin{table}[H]
\setlength{\tabcolsep}{3pt}
    \noindent
    \centering
    \caption{ Comparison of performance (Acc) of models across content types on \textbf{MIMIC-CXR-VQA}. 
    }
    \resizebox{0.75\textwidth}{!}{%
        \begin{tabular}{cccccccc}
        \toprule
        \multicolumn{8}{c}{Valid} \\
        \midrule
        Model & Plane & Gender & Size & Abnormality & Anatomy & Attribute & Presence \\
        \midrule
        Prior (Most) 
        & $16.7 $ 
        & $16.7 $ 
        & $50.0 $ 
        & $24.8 $ 
        & $0.0 $ 
        & $0.0 $ 
        & $50.1 $ \\
        Prior (Question) 
        & $50.0 $ 
        & $50.0 $ 
        & $50.0 $ 
        & $29.5 $ 
        & $12.8 $ 
        & $15.7 $ 
        & $50.1 $ \\
        PubMedCLIP 
        & $84.5_{\text{ ± } 1.0}$ 
        & $44.3_{\text{ ± } 6.6}$ 
        & $75.2_{\text{ ± } 1.2}$ 
        & $49.7_{\text{ ± } 2.0}$ 
        & $34.6_{\text{ ± } 1.8}$ 
        & $40.8_{\text{ ± } 3.0}$ 
        & $69.1_{\text{ ± } 0.9}$ \\
        PubMedCLIP$^\ast$ 
        & $89.8_{\text{ ± } 4.2}$ 
        & $53.5_{\text{ ± } 13.3}$ 
        & $74.7_{\text{ ± } 1.9}$ 
        & $50.9_{\text{ ± } 1.9}$ 
        & $35.7_{\text{ ± } 2.5}$ 
        & $43.0_{\text{ ± } 2.5}$ 
        & $70.2_{\text{ ± } 0.9}$ \\
        MedViLL$^\ast$ 
        & $89.4_{\text{ ± } 5.2}$
        & $65.0_{\text{ ± } 5.2}$
        & $77.7_{\text{ ± } 0.2}$
        & $60.3_{\text{ ± } 0.4}$
        & $43.8_{\text{ ± } 0.6}$
        & $55.0_{\text{ ± } 0.3}$
        & $76.6_{\text{ ± } 0.3}$ \\
        M$^{3}$AE 
        & $98.3_{\text{ ± } 1.2}$
        & $92.9_{\text{ ± } 0.4}$
        & $81.7_{\text{ ± } 0.3}$
        & $64.4_{\text{ ± } 0.5}$
        & $48.1_{\text{ ± } 0.6}$
        & $59.4_{\text{ ± } 0.2}$
        & $78.6_{\text{ ± } 0.1}$ \\
        M$^{3}$AE$^\ast$
        & $97.7_{\text{ ± } 1.6}$
        & $91.2_{\text{ ± } 1.1}$
        & $82.5_{\text{ ± } 0.4}$
        & $65.0_{\text{ ± } 0.5}$
        & $49.1_{\text{ ± } 0.3}$
        & $60.4_{\text{ ± } 0.3}$
        & $81.0_{\text{ ± } 0.1}$ \\
        \toprule
        \multicolumn{8}{c}{Test} \\
        \midrule
        Model & Plane & Gender & Size & Abnormality & Anatomy & Attribute & Presence \\
        \midrule
        Prior (Most)
        & $17.1 $ & $16.7 $ & $43.3 $ & $23.9 $ & $0.0 $ & $0.0 $ & $50.4 $ \\
        Prior (Question) 
        & $48.7 $ 
        & $50.0 $
        & $43.3 $ 
        & $29.0 $ 
        & $11.7 $ 
        & $12.4 $ 
        & $50.4 $ \\
        PubMedCLIP
        & $80.7_{\text{ ± } 2.0}$
        & $44.3_{\text{ ± } 8.4}$
        & $73.1_{\text{ ± } 0.1}$
        & $49.6_{\text{ ± } 1.6}$
        & $37.8_{\text{ ± } 1.0}$
        & $42.3_{\text{ ± } 2.3}$
        & $69.1_{\text{ ± } 0.8}$ \\
        PubMedCLIP$^\ast$ 
        & $87.5_{\text{ ± } 5.4}$
        & $51.3_{\text{ ± } 11.0}$
        & $74.1_{\text{ ± } 0.9}$
        & $50.3_{\text{ ± } 1.8}$
        & $38.5_{\text{ ± } 2.4}$
        & $45.2_{\text{ ± } 2.4}$
        & $70.1_{\text{ ± } 1.2}$ \\
        MedViLL$^\ast$ 
        & $90.2_{\text{ ± } 5.2}$
        & $67.2_{\text{ ± } 5.0}$
        & $75.1_{\text{ ± } 0.1}$
        & $59.2_{\text{ ± } 0.6}$
        & $45.3_{\text{ ± } 2.0}$
        & $53.1_{\text{ ± } 0.3}$
        & $76.0_{\text{ ± } 0.5}$ \\
        M$^{3}$AE 
        & $98.0_{\text{ ± } 1.0}$
        & $94.1_{\text{ ± } 0.8}$
        & $79.1_{\text{ ± } 0.5}$
        & $64.6_{\text{ ± } 0.5}$
        & $51.6_{\text{ ± } 0.4}$
        & $59.7_{\text{ ± } 0.6}$
        & $78.3_{\text{ ± } 1.3}$ \\
        M$^{3}$AE$^\ast$ 
        & $98.6_{\text{ ± } 1.0}$
        & $90.9_{\text{ ± } 1.0}$
        & $80.2_{\text{ ± } 1.1}$
        & $63.9_{\text{ ± } 0.2}$
        & $51.9_{\text{ ± } 1.7}$
        & $60.2_{\text{ ± } 0.4}$
        & $79.5_{\text{ ± } 0.7}$ \\
        \bottomrule
    \end{tabular}
    }
\end{table}

\newpage
\subsubsection{MIMIC-CXR-VQA: Relative metric for VQA grounding}
\paragraph{\CheckedBox\;\;Overview}
To estimate the achievable perception accuracy for single-image verification questions in MIMIC-CXR-VQA, we designed the reference model as a classification model. 
This model is capable of addressing our basic verification questions, which follow the template: ``\textit{Is there \$\{attribute\} in the \$\{object\}?}''.
The reference model is trained using a decomposed version of the MIMIC-CXR-VQA dataset. 
The details of the dataset construction and the reference model implementation are provided below.

\paragraph{\CheckedBox\;\;Construction of train/valid/test (\textit{verify})}
The dataset is constructed with a focus on the basic verification template (\ie, ``Is there \$\{attribute\} in \$\{object\}?'').
Note that all questions within the MIMIC-CXR-VQA dataset can be restructured as combinations of this basic verification template. 
For instance, a question such as ``\textit{Are there both \$\{attribute\_1\} and \$\{attribute\_2\} in the \$\{object\}?}'' can be divided into ``\textit{Is there \$\{attribute\_1\} in the \$\{object\}?}'' and ``\textit{Is there \$\{attribute\_2\} in the \$\{object\}?}''. 
Using this approach, we build the \textit{verify} dataset, which consists of train (\textit{verify}), valid (\textit{verify}), and test (\textit{verify}) sets. 
We decomposed the questions in each MIMIC-CXR-VQA dataset split to construct these subsets. 
We excluded category-related questions to avoid potential label imbalance. 

\paragraph{\CheckedBox\;\;Reference model structure}
The reference model focuses on local regions within the overall CXR image, guided by the bounding box data from Chest ImaGenome. 
Instead of considering the entire image as input, the model concentrates on these local regions in combination with attribute-specific headers to facilitate output for a grounding task. 
These headers identify the presence or absence of each attribute based on the local information provided. 
Through this strategy, the model can classify the relationship between objects and attributes, offering a distinct contrast to VQA models that rely on natural language questions to infer such relationships.
The backbone of the reference model is the Vision Transformer (ViT), which is pre-trained using the Self-Distillation with No Labels (DINO) method. 
DINO is a self-supervised learning strategy that utilizes a momentum encoder and multi-crop training. 
This approach empowers the self-supervised ViT features to effectively capture explicit semantic segmentation information from an image.
Considering that our VQA task requires attention to the anatomical locations specified in each question (\ie, ``\textit{... in the \$\{object\}}?'') and classifying its attribute relationship (\ie, ``\textit{Is there \$\{attribute\} ...}), the features derived from the DINO method offer significant advantages. 
Consequently, our reference model incorporates the DINO pre-trained ViT model as its backbone and adds a 3-layer MLP head for each attribute.

\paragraph{\CheckedBox\;\;Experimental details}
We train our reference model following the setting of the previous work. We first pre-train the DINO model using our pre-training corpus (\cref{sec:vqapretraining}). This pre-trained model is then fine-tuned using both the Train (\textit{verify}) and Valid set (\textit{verify}). The backbone of our model is ViT-S/16, and we use a 2D convolution layer with ReLU activation in each MLP head. During the pre-training phase, the model is trained using the AdamW optimizer with a batch size of 512, distributed over 8 GPUs. The learning rate increases linearly for the initial 10 epochs to a base value determined by the following linear scaling rule: $\textit{lr} = 0.0005 \times \text{batchsize}/256$. Following this warm-up period, the learning rate decays according to a cosine schedule. For fine-tuning, we train the model for 100 epochs with a batch size of 1024. Here, we use the Adam optimizer with an initial learning rate of 1e-3.

\newpage
\paragraph{Experimental results with relative metric}\;\;
The performance of our VQA models is evaluated by utilizing AUROC (Area Under the Receiver Operating Characteristic Curve) and relative AUROC metrics. Note that we only used object-attribute pairs with 10 or more instances where the corresponding object-attribute relationship was identified as positive (1) in the test set, to enhance the reliability of the evaluation. Thus, these metrics are computed across 82 specific (object, attribute) pairs within the MIMIC-CXR-VQA Test set (\textit{verify}). These metrics deliver an all-encompassing view on the model's predictive precision and its ability to accurately identify attributes within objects.
\begin{table}[H]
\caption{Comparison of performance of models across 82 (object, attribute) pairs on \textbf{MIMIC-CXR-VQA} Test set (\textit{Verify}). 
}
\resizebox{\textwidth}{!}{%
\begin{tabular}{llr|c|ccc|ccc}
\toprule
\multicolumn{1}{c}{\multirow{2.5}{*}{object}} &
  \multicolumn{1}{c}{\multirow{2.5}{*}{attribute}} &
  \multicolumn{1}{c}{\multirow{2.5}{*}{support}} &
  \multicolumn{4}{c}{AUROC} &
  \multicolumn{3}{c}{AUROC$_{rel}$} \\ \cmidrule(lr){4-7} \cmidrule(lr){8-10} 
   &
   &
   &
  \textit{\textbf{ref.}} model&
  M$^3$AE$^\ast$  &
  MedViLL &
  PubMedCLIP$^\ast$ &
  M$^3$AE$^\ast$  &
  MedViLL &
  PubMedCLIP$^\ast$  \\ \midrule
    aortic arch             & tortuous aorta & 69 & 0.719 & 0.880 & 0.852 & 0.645 & 1.225 & 1.185 & 0.898 \\
    aortic arch             & vascular calcification & 61 & 0.773 & 0.862 & 0.779 & 0.721 & 1.115 & 1.009 & 0.933 \\
    cardiac silhouette      & cardiac pacer and wires & 55 & 1.000 & 0.997 & 0.992 & 0.551 & 0.997 & 0.992 & 0.551 \\
    cardiac silhouette      & enlarged cardiac silhouette & 117 & 0.909 & 0.910 & 0.903 & 0.802 & 1.001 & 0.993 & 0.882 \\
    cardiac silhouette      & fluid overload/heart failure & 58 & 0.861 & 0.856 & 0.874 & 0.760 & 0.995 & 1.016 & 0.883 \\
    carina                  & endotracheal tube & 64 & 0.671 & 0.916 & 0.883 & 0.754 & 1.369 & 1.319 & 1.126 \\
    left costophrenic angle & costophrenic angle blunting & 62 & 0.681 & 0.873 & 0.762 & 0.689 & 1.286 & 1.123 & 1.016 \\
    left costophrenic angle & lung opacity & 151 & 0.694 & 0.864 & 0.799 & 0.682 & 1.246 & 1.152 & 0.983 \\
    left costophrenic angle & pleural effusion & 105 & 0.803 & 0.893 & 0.817 & 0.664 & 1.112 & 1.018 & 0.828 \\
    left hemidiaphragm      & hernia & 63 & 0.801 & 0.888 & 0.794 & 0.722 & 1.110 & 0.993 & 0.903 \\
    left hilar structures   & enlarged hilum & 60 & 0.671 & 0.778 & 0.709 & 0.634 & 1.161 & 1.059 & 0.945 \\
    left hilar structures   & lung opacity & 145 & 0.744 & 0.817 & 0.820 & 0.694 & 1.099 & 1.102 & 0.933 \\
    left hilar structures   & pulmonary edema/hazy opacity & 102 & 0.862 & 0.921 & 0.904 & 0.702 & 1.068 & 1.049 & 0.814 \\
    left hilar structures   & vascular congestion & 106 & 0.833 & 0.869 & 0.905 & 0.793 & 1.044 & 1.087 & 0.953 \\
    left lower lung zone    & aspiration & 63 & 0.897 & 0.898 & 0.820 & 0.746 & 1.001 & 0.914 & 0.831 \\
    left lower lung zone    & atelectasis & 113 & 0.833 & 0.841 & 0.735 & 0.646 & 1.010 & 0.883 & 0.776 \\
    left lower lung zone    & linear/patchy atelectasis & 69 & 0.661 & 0.766 & 0.643 & 0.643 & 1.159 & 0.972 & 0.973 \\
    left lower lung zone    & low lung volumes & 102 & 0.742 & 0.805 & 0.690 & 0.693 & 1.088 & 0.932 & 0.938 \\
    left lower lung zone    & lung opacity & 151 & 0.770 & 0.788 & 0.741 & 0.561 & 1.023 & 0.962 & 0.728 \\
    left lower lung zone    & pneumonia & 92 & 0.813 & 0.830 & 0.748 & 0.719 & 1.021 & 0.919 & 0.885 \\
    left lung               & aspiration & 75 & 0.901 & 0.871 & 0.839 & 0.707 & 0.967 & 0.931 & 0.785 \\
    left lung               & atelectasis & 123 & 0.871 & 0.874 & 0.779 & 0.684 & 1.004 & 0.895 & 0.786 \\
    left lung               & copd/emphysema & 52 & 0.904 & 0.904 & 0.866 & 0.678 & 1.000 & 0.958 & 0.751 \\
    left lung               & costophrenic angle blunting & 70 & 0.762 & 0.883 & 0.760 & 0.664 & 1.159 & 0.998 & 0.872 \\
    left lung               & enlarged hilum & 60 & 0.772 & 0.765 & 0.695 & 0.585 & 0.991 & 0.901 & 0.758 \\
    left lung               & fluid overload/heart failure & 60 & 0.827 & 0.864 & 0.893 & 0.753 & 1.045 & 1.080 & 0.910 \\
    left lung               & hyperaeration & 81 & 0.924 & 0.943 & 0.924 & 0.741 & 1.020 & 1.000 & 0.802 \\
    left lung               & linear/patchy atelectasis & 67 & 0.636 & 0.784 & 0.681 & 0.616 & 1.234 & 1.072 & 0.969 \\
    left lung               & low lung volumes & 107 & 0.904 & 0.913 & 0.865 & 0.714 & 1.010 & 0.957 & 0.790 \\
    left lung               & lung lesion & 85 & 0.685 & 0.685 & 0.654 & 0.577 & 1.004 & 0.958 & 0.845 \\
    left lung               & lung opacity & 158 & 0.802 & 0.785 & 0.739 & 0.662 & 0.979 & 0.921 & 0.825 \\
    left lung               & mass/nodule (not otherwise specified) & 87 & 0.661 & 0.763 & 0.588 & 0.548 & 1.159 & 0.893 & 0.832 \\
    left lung               & pleural effusion & 120 & 0.865 & 0.893 & 0.809 & 0.674 & 1.032 & 0.935 & 0.779 \\
    left lung               & pleural/parenchymal scarring & 86 & 0.782 & 0.824 & 0.768 & 0.602 & 1.056 & 0.984 & 0.772 \\
    left lung               & pneumonia & 99 & 0.820 & 0.766 & 0.711 & 0.668 & 0.936 & 0.868 & 0.815 \\
    left lung               & pulmonary edema/hazy opacity & 109 & 0.901 & 0.912 & 0.890 & 0.711 & 1.011 & 0.988 & 0.789 \\
    left lung               & vascular congestion & 103 & 0.869 & 0.871 & 0.885 & 0.786 & 1.001 & 1.018 & 0.904 \\
    left mid lung zone      & lung opacity & 148 & 0.685 & 0.821 & 0.712 & 0.523 & 1.200 & 1.041 & 0.765 \\
    mediastinum             & cardiac pacer and wires & 54 & 0.999 & 0.997 & 0.992 & 0.543 & 0.998 & 0.993 & 0.544 \\
    mediastinum             & enlarged cardiac silhouette & 122 & 0.905 & 0.899 & 0.875 & 0.793 & 0.994 & 0.968 & 0.876 \\
    mediastinum             & enteric tube & 66 & 0.890 & 0.998 & 0.978 & 0.777 & 1.122 & 1.099 & 0.873 \\
    mediastinum             & fluid overload/heart failure & 57 & 0.825 & 0.851 & 0.865 & 0.749 & 1.033 & 1.049 & 0.908 \\
    mediastinum             & hernia & 76 & 0.894 & 0.915 & 0.755 & 0.647 & 1.024 & 0.844 & 0.724 \\
    mediastinum             & ij line & 47 & 0.951 & 0.990 & 0.976 & 0.784 & 1.041 & 1.026 & 0.824 \\
    mediastinum             & superior mediastinal mass/enlargement & 68 & 0.752 & 0.804 & 0.689 & 0.625 & 1.070 & 0.916 & 0.831 \\
    mediastinum             & tortuous aorta & 74 & 0.904 & 0.867 & 0.837 & 0.619 & 0.959 & 0.926 & 0.685 \\
    mediastinum             & vascular calcification & 63 & 0.866 & 0.870 & 0.803 & 0.724 & 1.005 & 0.928 & 0.837 \\
    right costophrenic angle& lung opacity & 149 & 0.828 & 0.888 & 0.827 & 0.699 & 1.074 & 1.000 & 0.845 \\
    right costophrenic angle& pleural effusion & 102 & 0.733 & 0.844 & 0.804 & 0.618 & 1.151 & 1.096 & 0.843 \\
    right hemidiaphragm     & elevated hemidiaphragm & 58 & 0.912 & 0.968 & 0.775 & 0.547 & 1.062 & 0.850 & 0.600 \\
    right hilar structures  & enlarged hilum & 58 & 0.739 & 0.838 & 0.751 & 0.620 & 1.134 & 1.017 & 0.839 \\
    right hilar structures  & lung opacity & 152 & 0.790 & 0.855 & 0.831 & 0.689 & 1.082 & 1.051 & 0.872 \\
    right hilar structures  & pulmonary edema/hazy opacity & 96 & 0.867 & 0.908 & 0.896 & 0.693 & 1.047 & 1.034 & 0.799 \\
    right hilar structures  & vascular congestion & 100 & 0.860 & 0.857 & 0.884 & 0.798 & 0.997 & 1.028 & 0.929 \\
    right lower lung zone   & aspiration & 63 & 0.883 & 0.939 & 0.856 & 0.783 & 1.064 & 0.969 & 0.888 \\
    right lower lung zone   & atelectasis & 113 & 0.790 & 0.814 & 0.753 & 0.642 & 1.031 & 0.953 & 0.813 \\
    right lower lung zone   & linear/patchy atelectasis & 64 & 0.743 & 0.800 & 0.680 & 0.680 & 1.077 & 0.915 & 0.916 \\
    right lower lung zone   & low lung volumes & 106 & 0.827 & 0.833 & 0.713 & 0.708 & 1.010 & 0.864 & 0.858 \\
    right lower lung zone   & lung opacity & 154 & 0.786 & 0.769 & 0.741 & 0.600 & 0.979 & 0.943 & 0.764 \\
    right lower lung zone   & pneumonia & 91 & 0.835 & 0.789 & 0.768 & 0.612 & 0.945 & 0.920 & 0.733 \\
    right lung              & airspace opacity & 62 & 0.825 & 0.839 & 0.786 & 0.562 & 1.018 & 0.954 & 0.682 \\
    right lung              & aspiration & 65 & 0.936 & 0.953 & 0.864 & 0.740 & 1.019 & 0.924 & 0.791 \\
    right lung              & atelectasis & 129 & 0.809 & 0.833 & 0.758 & 0.685 & 1.031 & 0.938 & 0.848 \\
    right lung              & copd/emphysema & 51 & 0.912 & 0.916 & 0.877 & 0.702 & 1.004 & 0.962 & 0.770 \\
    right lung              & enlarged hilum & 62 & 0.768 & 0.830 & 0.715 & 0.586 & 1.082 & 0.931 & 0.763 \\
    right lung              & fluid overload/heart failure & 67 & 0.859 & 0.861 & 0.886 & 0.763 & 1.002 & 1.032 & 0.888 \\
    right lung              & hyperaeration & 78 & 0.961 & 0.941 & 0.940 & 0.766 & 0.979 & 0.978 & 0.797 \\
    right lung              & linear/patchy atelectasis & 65 & 0.804 & 0.806 & 0.751 & 0.728 & 1.002 & 0.935 & 0.905 \\
    right lung              & low lung volumes & 105 & 0.922 & 0.915 & 0.861 & 0.715 & 0.993 & 0.934 & 0.775 \\
    right lung              & lung lesion & 78 & 0.767 & 0.832 & 0.698 & 0.557 & 1.085 & 0.910 & 0.726 \\
    right lung              & lung opacity & 153 & 0.803 & 0.779 & 0.722 & 0.622 & 0.971 & 0.899 & 0.775 \\
    right lung              & mass/nodule (not otherwise specified) & 83 & 0.823 & 0.810 & 0.775 & 0.639 & 0.985 & 0.943 & 0.777 \\
    right lung              & pleural effusion & 113 & 0.841 & 0.851 & 0.803 & 0.632 & 1.012 & 0.955 & 0.751 \\
    right lung              & pleural/parenchymal scarring & 98 & 0.741 & 0.845 & 0.706 & 0.602 & 1.140 & 0.953 & 0.813 \\
    right lung              & pneumonia & 92 & 0.867 & 0.816 & 0.785 & 0.668 & 0.941 & 0.905 & 0.770 \\
    right lung              & pulmonary edema/hazy opacity & 111 & 0.928 & 0.924 & 0.895 & 0.727 & 0.995 & 0.964 & 0.783 \\
    right lung              & vascular congestion & 107 & 0.878 & 0.849 & 0.865 & 0.798 & 0.967 & 0.985 & 0.909 \\
    right mid lung zone     & lung opacity & 145 & 0.674 & 0.866 & 0.730 & 0.635 & 1.285 & 1.082 & 0.941 \\
    trachea                 & endotracheal tube & 61 & 0.973 & 0.975 & 0.947 & 0.795 & 1.003 & 0.974 & 0.817 \\
    upper mediastinum       & superior mediastinal mass/enlargement & 65 & 0.774 & 0.821 & 0.712 & 0.644 & 1.062 & 0.920 & 0.832 \\
    upper mediastinum       & tortuous aorta & 67 & 0.843 & 0.881 & 0.845 & 0.621 & 1.045 & 1.003 & 0.737 \\
    upper mediastinum       & vascular calcification & 63 & 0.871 & 0.859 & 0.777 & 0.734 & 0.986 & 0.892 & 0.843 \\
\bottomrule
\end{tabular}%
}
\end{table}

\subsection{MIMIC-CXR-VQA: Exploring data redundancy and the impact of paraphrasing}
Given the brittleness of templates in prior EHR QA work, where the large sample size of emrQA (medication=220K, relation=900K) led to the redundancy issue, we decided to investigate the MIMIC-CXR-VQA dataset specifically, as it has a larger dataset size (377K).

First, we explore the degree to which additional templates actually improve the model to determine whether our dataset simply contains repetitive templates without any novelty. To this end, we conducted an ablation experiment with the MIMIC-CXR-VQA dataset. We evaluate the test set performance by randomly sampling training data at various template usage proportions (\ie, how many unique templates are used for training), such as 5\%, 10\%, 20\%, 50\%, and 100\%. 

As shown in \cref{tab:mimic-cxr-vqa_sampleratio}, our experiment with MIMIC-CXR-VQA demonstrated that using a higher number of questions (\ie, training size increases) generated from templates questions generated from more diverse templates positively impacts the model's performance (\ie, test Acc/F1) across all models.  This observation suggests that MIMIC-CXR-VQA does not have the redundancy of questions, and question diversity generated from all templates contributes to the performance improvement.

\begin{table}[H]
    \setlength{\tabcolsep}{3pt}
    \noindent
    \centering
        \caption{Results of the ablation experiment on the MIMIC-CXR-VQA test set. Comparison of performance across different training data proportions based on unique template usage: 5\%, 10\%, 20\%, 50\%, and 100\%. We ran the experiments with one seed.
        }
    \label{tab:mimic-cxr-vqa_sampleratio}
        \resizebox{0.5\textwidth}{!}{%
            \begin{tabular}{@{}cccc@{}}
            \toprule
                               & \multicolumn{1}{c}{PubMedCLIP$^{\ast}$} & \multicolumn{1}{c}{MedViLL} & \multicolumn{1}{c}{M$^{3}$AE$^{\ast}$} \\ \cmidrule{2-4}
            template usage (\%) & test (Acc/F1)                            & test (Acc/F1)                         & test (Acc/F1)                      \\ \midrule
            5\%   & 49.1 / 0.39 & 44.8 / 0.44 & 56.9 / 0.56 \\
            10\%  & 48.4 / 0.50 & 54.4 / 0.55 & 61.3 / 0.64 \\
            20\%  & 51.2 / 0.50 & 60.0 / 0.62 & 65.0 / 0.67 \\
            50\%  & 55.8 / 0.56 & 62.6 / 0.65 & 68.9 / 0.72 \\
            100\% & 56.5 / 0.56 & 63.6 / 0.67 & 69.2 / 0.73 \\
            \bottomrule
            \end{tabular}%
            }
  \end{table}

Next, we explored the relationship between the variety of paraphrases produced using GPT-4 and model performance. To investigate whether increasing the diversity of paraphrases (thus reducing the redundancy of the question) improves the model's ability, we keep the dataset size constant and vary the number of paraphrases per template for the training dataset. We started with 48 seed templates as the training dataset (denoted as ``seed template'') and created two distinct training dataset variants named ``low-diversity'' and ``high-diversity''. These variants contained 20\% and 100\% paraphrased templates of the original template in the MIMIC-CXR-VQA training set, respectively. Note that each model, trained with three different datasets, was evaluated against the same original MIMIC-CXR-VQA test dataset. We conducted the experiments with three different seeds.

Our findings indicate that using paraphrased templates via GPT-4 improves model performance. This observation is evident when comparing the results of the ``seed template'' (without paraphrasing) with models trained in the variants ``low-diversity'' or ``high-diversity''. Further, as the diversity of paraphrases increases, the model performance might also improve. The effect of paraphrasing can differ based on the text encoder used. Models that incorporate a BERT-based language architecture, such as MedViLL and M$^3$AE, appear to benefit more from increased diversity. This suggests that even templates crafted through automatic paraphrasing can substantially boost performance, especially when there is adequate diversity.

\begin{table}[H]
    \setlength{\tabcolsep}{3pt}
    \noindent
    \centering
        \caption{Results of the ablation experiment on the MIMIC-CXR-VQA test set. Comparison of performance across different degrees of paraphrasing diversity. We ran the experiments with three different seeds (mean ± std).
        }
    \label{tab:mimic-cxr-vqa_diversity}
        \resizebox{0.8\textwidth}{!}{%
            \begin{tabular}{@{}lccc@{}}
            \toprule
                               & \multicolumn{1}{c}{PubMedCLIP$^{\ast}$} & \multicolumn{1}{c}{MedViLL} & \multicolumn{1}{c}{M$^{3}$AE$^{\ast}$} \\ \cmidrule{2-4}
            \multicolumn{1}{c}{Training Dataset}   & test  (Acc/F1)          & test  (Acc/F1)          & test  (Acc/F1)          \\ \midrule
            MIMIC-CXR-VQA (seed template) & 39.6 ± 0.4 / 0.36 ± 0.0 & 46.3 ± 0.3 / 0.48 ± 0.0 & 65.6 ± 0.5 / 0.68 ± 0.0 \\
            MIMIC-CXR-VQA (low-diversity)    & 56.5 ± 0.3 / 0.57 ± 0.0 & 62.5 ± 0.3 / 0.65 ± 0.0 & 69.2 ± 0.4 / 0.72 ± 0.0 \\
            MIMIC-CXR-VQA (high-diversity)    & 56.5 ± 2.1 / 0.56 ± 0.0 & 63.6 ± 0.1 / 0.67 ± 0.0 & 69.2 ± 0.4 / 0.73 ± 0.0 \\
            \bottomrule
            \end{tabular}%
            }
  \end{table}

\newpage
\subsection{EHRXQA}\label{supp_exp_xqa}
\subsubsection{Implementation details of baselines}

Our NeuralSQL-based approach integrates a large language model (LLM) as a parser with an external VQA API module. 
We use ChatGPT~\cite{openai2022chatgpt} (\texttt{gpt-3.5-turbo-0613})\footnote{Since the Codex API is no longer supported, we conducted all our experiments using ChatGPT (\texttt{gpt-3.5-turbo-0613}) instead.} as our parser and utilize the \MAE~model, pre-trained on the MIMIC-CXR-VQA training set, as the VQA API module (frozen). 
We conduct in-context learning with few-shot samples, leveraging the capabilities of large language models, specifically in a 10-shot setting. 

We have defined two prompting strategies: 1) Fixed: which uses 10-shot (Question, NeuralSQL) pairs; 2) BM25 (train): which retrieves 10 relevant pairs via BM25 from the training set for any given question. 
The fixed examples are randomly selected from the training set, but we ensured that at least one pair was sampled from each modality to provide the minimum required information (3 for \textit{Table}-related, 4 for \textit{Image}-related, and 3 for \textit{Image+Table}-related). 
When executing NeuralSQL parsed by LLM, the batch size of the external VQA module is set to 16.

\begin{tcolorbox}[
    colback=white,
    colframe=black,
    fonttitle=\bfseries,
    coltitle=white,
    title={Prompt for Fixed strategy.},
]
Generate NeuralSQL (\ie, extended SQL with the following conditions) given the question to answer the question correctly. If the question can only be answered by examining a chest X-ray image and requires a VQA model, use the new syntax FUNC\_VQA() to create a query. 
When the VQA sentence in FUNC\_VQA() syntax contains logical operations such as union, difference, intersection, disjunction, or conjunction, decomposes the VQA statement into minimal semantic units and uses the SQL syntax to generate NeuralSQL.
For example, decompose the original sentence "Are there any technical assessment or tubes/lines?" into "Are there any technical assessment?" and "Are there any tubes/lines?" by separating the logical disjunction (or) and creating two separate questions.
\newline
\newline
Q: \texttt{\{\{1st question\}\}}
\newline
NeuralSQL: \texttt{\{\{1st query\}\}}
\newline
\newline
\texttt{...}
\newline
\newline
Q: \texttt{\{\{10th question\}\}}
\newline
NeuralSQL: \texttt{\{\{10th query\}\}}
\newline
\newline
Parse the question into NeuralSQL.
\newline
Q: \texttt{\{\{target question\}\}}
\newline
NeuralSQL:
\end{tcolorbox}

\newpage
\begin{tcolorbox}[
    colback=white,
    colframe=black,
    fonttitle=\bfseries,
    coltitle=white,
    title={Prompt for BM25 strategy.},
]
Generate NeuralSQL (\ie, extended SQL with the following conditions) given the question to answer the question correctly. If the question can only be answered by examining a chest X-ray image and requires a VQA model, use the new syntax FUNC\_VQA() to create a query. 
When the VQA sentence in FUNC\_VQA() syntax contains logical operations such as union, difference, intersection, disjunction, or conjunction, decomposes the VQA statement into minimal semantic units and uses the SQL syntax to generate NeuralSQL.
For example, decompose the original sentence "Are there any technical assessment or tubes/lines?" into "Are there any technical assessment?" and "Are there any tubes/lines?" by separating the logical disjunction (or) and creating two separate questions.
Q: \texttt{\{\{1st question retrieved from training QA set\}\}}
\newline
NeuralSQL: \texttt{\{\{1st query\}\}}
\newline
\newline
\texttt{...}
\newline
\newline
Q: \texttt{\{\{10th question retrieved from training QA set\}\}}
\newline
NeuralSQL: \texttt{\{\{10th query\}\}}
\newline
\newline
Parse the question into NeuralSQL.
\newline
Q: \texttt{\{\{target question\}\}}
\newline
NeuralSQL:
\end{tcolorbox}

\newpage
\subsubsection{EHRXQA: Experimental results}\label{supp_exp_xqa_detail_result}
\begin{table}[H]
\caption{Performance of ChatGPT + M$^{3}$AE (BM25 (train)) on the \textbf{EHRXQA} dataset, categorized by modality-based and patient-based scope.}
\label{tab:performance_detail_bm25}
\vspace{3mm}
\centering
\renewcommand{\arraystretch}{1.0}
\begin{adjustbox}{width=0.6\columnwidth,center}
\begin{tabular}{cccccc}
\toprule
Modality-based & \multicolumn{2}{c}{Patient-based} & $Acc_{LF}$   & $Acc_{EX|gt}$   & $Acc_{EX|pred}$  \\ \midrule
\multirow{5.5}{*}{Image}        & \multirow{3.5}{*}{single} & 1-image & 94.4  & 57.6 &  56.3 \\ \cmidrule(lr){3-6} 
                               &                            & 2-image & 73.3  & 51.1 &  50.0 \\ \cmidrule(lr){3-6} 
                               &                            & N-image & 90.8  & 39.6 &  39.6 \\ \cmidrule(lr){2-6} 
                               & \multicolumn{2}{c}{group}            & 85.0  & \enskip 5.0  &  \enskip 1.7  \\ \cmidrule(lr){1-6} 
\multirow{4}{*}{Table}         & \multicolumn{2}{c}{none}             & 98.0  & 100.0 &  98.0 \\ \cmidrule(lr){2-6} 
                               & \multicolumn{2}{c}{single}           & 87.1  & 100.0 &  95.6 \\ \cmidrule(lr){2-6} 
                               & \multicolumn{2}{c}{group}            & 44.7  & 100.0 &  87.5 \\ \cmidrule(lr){1-6} 
\multirow{2.5}{*}{Image + Table} & \multicolumn{2}{c}{single}         & 70.5  & 78.3  &  75.2 \\ \cmidrule(lr){2-6} 
                               & \multicolumn{2}{c}{group}            & 83.6  & 15.0  &  13.1  \\ \bottomrule 
\end{tabular}
\end{adjustbox}
\end{table}

\begin{table}[H]
\caption{Performance of ChatGPT + M$^{3}$AE (Fixed) on the \textbf{EHRXQA} dataset, categorized by modality-based and patient-based scope.}
\label{tab:performance_detail_fixed}
\vspace{3mm}
\centering
\renewcommand{\arraystretch}{1.0}
\begin{adjustbox}{width=0.6\columnwidth,center}
\begin{tabular}{cccccc}
\toprule
Modality-based & \multicolumn{2}{c}{Patient-based} & $Acc_{LF}$   & $Acc_{EX|gt}$   & $Acc_{EX|pred}$  \\ \midrule
\multirow{5.5}{*}{Image}        & \multirow{3.5}{*}{single} & 1-image & \enskip 1.4  & \enskip 57.6 &  27.7 \\ \cmidrule(lr){3-6} 
                               &                            & 2-image & \enskip 0.2  & \enskip 51.1 &  10.3 \\ \cmidrule(lr){3-6} 
                               &                            & N-image & \enskip 0.0  & \enskip 39.6 &  \enskip 3.8 \\ \cmidrule(lr){2-6} 
                               & \multicolumn{2}{c}{group}            & \enskip 5.0  & \enskip 5.0  &  \enskip 0.8  \\ \cmidrule(lr){1-6} 
\multirow{4}{*}{Table}         & \multicolumn{2}{c}{none}             & \enskip 0.0  & 100.0 &  \enskip 6.0 \\ \cmidrule(lr){2-6} 
                               & \multicolumn{2}{c}{single}           & \enskip 5.9  & 100.0 &  33.6 \\ \cmidrule(lr){2-6} 
                               & \multicolumn{2}{c}{group}            & \enskip 3.4  & 100.0 &  25.4 \\ \cmidrule(lr){1-6} 
\multirow{2.5}{*}{Image + Table} & \multicolumn{2}{c}{single}         & \enskip 3.3  & \enskip 78.3  &  40.8 \\ \cmidrule(lr){2-6} 
                               & \multicolumn{2}{c}{group}            & 13.1 & \enskip 15.0  &  \enskip 6.5  \\ \bottomrule 
\end{tabular}
\end{adjustbox}
\end{table}

\newpage
\section{Author statement}
The authors of this paper bear all responsibility in case of violation of rights, etc. associated with the MIMIC-CXR-VQA and EHRXQA dataset.

\end{document}